\documentclass[10pt,twocolumn,letterpaper]{article}

\usepackage{iccv}
\usepackage{times}
\usepackage{epsfig}
\usepackage{graphicx}
\usepackage{amsmath}
\usepackage{amssymb}
\usepackage{subfigure}
\usepackage{mathtools}
\usepackage{booktabs}

\usepackage[pagebackref=true,breaklinks=true,letterpaper=true,colorlinks,bookmarks=false]{hyperref}

\definecolor{mariocolor}{rgb}{1,0,1}

\newcommand{\ning}[1]{#1}

\iccvfinalcopy 

\def\iccvPaperID{2010} 

\ificcvfinal\fi
\begin{document}

\title{
Attributing Fake Images to GANs: Learning and Analyzing GAN Fingerprints}

\author{
Ning Yu\hspace{0.01in}\textsuperscript{1,2} \hspace{2cm} Larry Davis\hspace{0.01in}\textsuperscript{1} \hspace{2cm} Mario Fritz\hspace{0.01in}\textsuperscript{3}\\
\textsuperscript{1}\hspace{0.01in}University of Maryland, College Park \\
\textsuperscript{2}\hspace{0.01in}Max Planck Institute for Informatics\\Saarland Informatics Campus, Germany\\
\textsuperscript{3}\hspace{0.01in}CISPA Helmholtz Center for Information Security\\Saarland Informatics Campus, Germany\\
{\tt\small ningyu@mpi-inf.mpg.de}
\hspace{3ex}{\tt\small lsd@cs.umd.edu}
\hspace{3ex}{\tt\small fritz@cispa.saarland}\\
}

\maketitle
\thispagestyle{empty}

\begin{abstract}
Recent advances in Generative Adversarial Networks (GANs) have shown increasing success in generating photorealistic images. But they also raise challenges to visual forensics and model attribution. We present the first study of learning GAN fingerprints towards image attribution and using them to classify an image as real or GAN-generated. For GAN-generated images, we further identify their sources. Our experiments show that (1) GANs carry distinct model fingerprints and leave stable fingerprints in their generated images, which support image attribution; (2) even minor differences in GAN training can result in different fingerprints, which enables fine-grained model authentication; (3) fingerprints persist across different image frequencies and patches and are not biased by GAN artifacts; (4) fingerprint finetuning is effective in immunizing against five types of adversarial image perturbations; and (5) comparisons also show our learned fingerprints consistently outperform several baselines in a variety of setups~\footnote{Code, data, models, and supplementary material are available at \href{https://github.com/ningyu1991/GANFingerprints.git}{GitHub}.}.
\end{abstract}

\section{Introduction}

In the last two decades, photorealistic image generation and manipulation techniques have rapidly evolved. Visual contents can now be easily created and edited without leaving obvious perceptual traces~\cite{zollhofer2018state}. Recent breakthroughs in generative adversarial networks (GANs)~\cite{goodfellow2014generative,salimans2016improved,arjovsky2017wasserstein,gulrajani2017improved,karras2018progressive,brock2018large} have further improved the quality and photorealism of generated images. The adversarial framework of GANs can also be used in conditional scenarios for image translation~\cite{isola2017image,zhu2017unpaired,zhu2017toward} or manipulation in a given context~\cite{thies2015real,thies2016face2face,suwajanakorn2017synthesizing,bau2018visualizing,yu2019texture}, which diversifies media synthesis.  

At the same time, however, the success of GANs has raised two challenges to the vision community: visual forensics and intellectual property protection.

\begin{figure}[t]
\centering
\includegraphics[width=\linewidth]{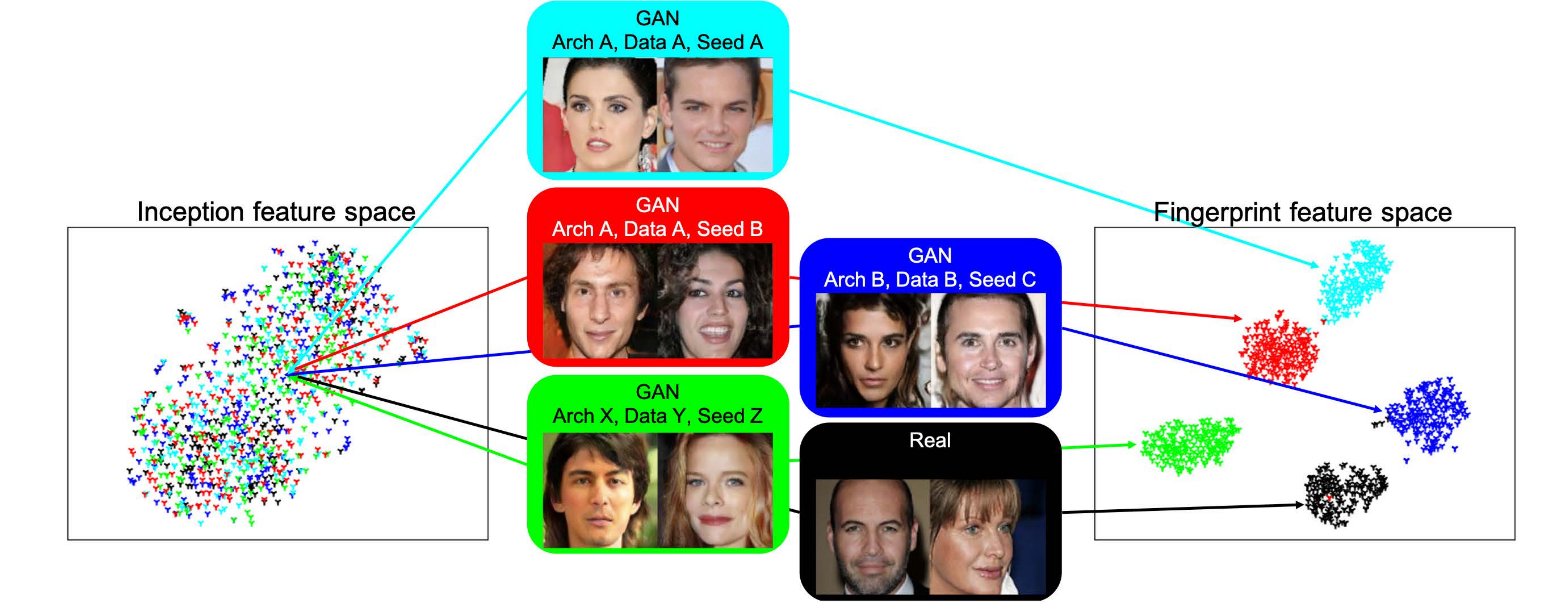}
\caption{A t-SNE~\cite{maaten2008visualizing} visual comparison between our fingerprint features (right) and the baseline inception features~\cite{salimans2016improved} (left) for image attribution. Inception features are highly entangled, indicating the challenge to differentiate high-quality GAN-generated images from real ones. However, our result shows any single difference in GAN architectures, training sets, or even initialization seeds can result in distinct fingerprint features for effective attribution.}
\label{fig:teaser}
\vspace{-12pt}
\end{figure}

\vspace{-16pt}
\paragraph{GAN challenges to visual forensics.} There is a widespread concern about the impact of this technology when used maliciously. This issue has also received increasing public attention, in terms of disruptive consequences to visual security, laws, politics, and society in general~\cite{guardian,nbc,newyorker}. Therefore, it is critical to look into effective visual forensics against threats from GANs.

While recent state-of-the-art visual forensics techniques demonstrate impressive results for detecting fake visual media~\cite{bestagini2013local,sencar2013digital,farid2016photo,bayar2016deep,cozzolino2017recasting,bappy2017exploiting,huh2018fighting,zhou2017two,zhou2018learning,d2019patchmatch}, they have only focused on semantic, physical, or statistical inconsistency of specific forgery scenarios, e.g., copy-move manipulations\cite{bestagini2013local,d2019patchmatch} or face swapping~\cite{zhou2017two}.  Forensics on GAN-generated images~\cite{marra2018detection,mo2018fake,tariq2018detecting} shows good accuracy, but each method operates on only one GAN architecture by identifying its unique artifacts and results deteriorate when the GAN architecture is changed. It is still an open question of whether GANs leave stable marks that are commonly shared by their generated images. That motivates us to investigate an effective feature representation that differentiates GAN-generated images from real ones.

\vspace{-16pt}
\paragraph{GAN challenges to intellectual property protection.} Similar to other successful applications of deep learning technology to image recognition~\cite{he2016deep} or natural language processing~\cite{goldberg2016primer}, building a product based on GANs is non-trivial~\cite{jia2014caffe,modelGallery,stolenData}. It requires a large amount of training data, powerful computing resources, significant machine learning expertise, and numerous trial-and-error iterations for identifying optimal model architectures and their model hyper-parameters. As GAN services become widely deployed with commercial potential, they will become increasingly vulnerable to pirates. Such copyright plagiarism may jeopardize the intellectual property of model owners and take future market share from them. Therefore, methods for attributing GAN-generated image origins are highly desirable for protecting intellectual property. 

\ning{Given the level of realism that GAN techniques already achieve today, attribution by human inspection is no longer feasible (see the mixture of Figure~\ref{fig:data}).} The state-of-the-art digital identification techniques can be separated into two categories: digital watermarking and digital fingerprint detection. Neither of them is applicable to GAN attribution. Previous work on watermarking deep neural networks~\cite{zhang2018protecting,uchida2017embedding} depends on an embedded security scheme during ``white-box" model training, requires control of the input, and is impractical when only GAN-generated images are accessible in a ``black-box" scenario. Previous work on digital fingerprints is limited to device fingerprints~\cite{lukas2006digital,chen2008determining} or in-camera post-processing fingerprints~\cite{cozzolino2018noiseprint}, which cannot be easily adapted to GAN-generated images. That motivates us to investigate GAN fingerprints that attribute different GAN-generated images to their sources.

We present the first study addressing the two GAN challenges simultaneously by learning GAN fingerprints for image attribution: \ning{We introduce GAN fingerprints and use them to classify an image as real or GAN-generated.} For GAN-generated images, we further identify their sources. \ning{We approach this by training a neural network classifier and predicting the source of an image. Our experiments show that GANs carry distinct model fingerprints and leave stable fingerprints in their generated images, which support image attribution.}

\ning{We summarize our \textbf{contributions} as demonstrating the existence, uniqueness, persistence, immunizability, and visualization of GAN fingerprints. We address the following questions:}

\vspace{-16pt}
\paragraph{\ning{Existence and uniqueness:} Which GAN parameters differentiate image attribution?} We present experiments on GAN parameters including architecture, training data, as well as random initialization seed. \ning{We find that a difference in any one of these parameters results in a unique GAN fingerprint for image attribution.} See Figure~\ref{fig:teaser}, Section~\ref{sec:attribution} and \ref{sec:attribution_exp}.

\vspace{-16pt}
\paragraph{\ning{Persistence:} Which image components contain fingerprints for attribution?} We investigate image components in different frequency bands and in different patch sizes. In order to eliminate possible bias from GAN artifact components, we apply a perceptual similarity metric to distill an artifact-free subset for attribution evaluation. \ning{We find that GAN fingerprints are persistent across different frequencies and patch sizes, and are not dominated by artifacts.} See Section~\ref{sec:pooling} and \ref{sec:pooling_exp}.

\vspace{-16pt}
\paragraph{\ning{Immunizability:} How robust is attribution to image perturbation attacks and how effective are the defenses?} We investigate common attacks that aim at destroying image fingerprints. They include noise, blur, cropping, JPEG compression, relighting, and random combinations of them. We also defend against such attacks by finetuning our attribution classifier. See Section~\ref{sec:attack_defense}.

\vspace{-16pt}
\paragraph{\ning{Visualization:} How to expose GAN fingerprints?} We propose an alternative classifier variant to explicitly visualize GAN fingerprints in the image domain, \ning{so as to better interpret the effectiveness of attribution}. See Section~\ref{sec:fingerprint} and \ref{sec:fingerprint_exp}.

\vspace{-16pt}
\paragraph{Comparison to baselines.} In terms of attribution accuracy, our method consistently outperforms three baseline methods (including a very recent one~\cite{marra2019gans}) on two datasets under a variety of experimental conditions. In terms of feature representation, our fingerprints show superior distinguishability across image sources compared to inception features~\cite{salimans2016improved}.

\section{Related work}
\paragraph{Generative Adversarial Networks (GANs).} GANs~\cite{goodfellow2014generative,salimans2016improved,arjovsky2017wasserstein,gulrajani2017improved,karras2018progressive,brock2018large} have shown improved photorealism in image synthesis~\cite{li2016precomputed, bergmann2017learning,zhou2018nonstationary}, translation~\cite{isola2017image,zhu2017unpaired,zhu2017toward}, or manipulation~\cite{antipov2017face,thies2015real,thies2016face2face}. We focus on unconditional GANs as the subject of our study. We choose the following four GAN models as representative candidates of the current state of the art: ProGAN~\cite{karras2018progressive}, SNGAN~\cite{miyato2018spectral}, CramerGAN~\cite{bellemare2017cramer}, and MMDGAN~\cite{binkowski2018demystifying}, considering their outstanding performances on face generation.

\vspace{-16pt}
\paragraph{Visual forensics.} Visual forensics targets detecting statistical or physics-based artifacts and then recognizing the authenticity of visual media without evidence from an embedded security mechanism~\cite{fridrich2012digital,farid2016photo}. An example is a steganalysis-based method~\cite{fridrich2012rich}, which uses hand-crafted features plus a linear Support Vector Machine to detect forgeries. Recent CNN-based methods~\cite{bayar2016deep,cozzolino2017recasting,bondi2017tampering,bappy2017exploiting,huh2018fighting,zhou2017two,zhou2018learning,afchar2018mesonet,cozzolino2018forensictransfer,d2019patchmatch} learn deep features and further improve tampering detection performance on images or videos. R\"{o}ssler \etal~\cite{rossler2018faceforensics,rossler2019faceforensics++} introduced a large-scale face manipulation dataset to benchmark forensics classification and segmentation tasks, and demonstrated superior performance when using additional domain-specific knowledge. For forensics on GAN-generated images, several existing works~\cite{marra2018detection,mo2018fake,tariq2018detecting} show good accuracy. However, each method considers only one GAN architecture and results do not generalize across architectures.

\vspace{-16pt}
\paragraph{Digital fingerprints.} Prior digital fingerprint techniques focus on detecting hand-crafted features for either device fingerprints or postprocessing fingerprints. The device fingerprints rely on the fact that individual devices, due to manufacturing imperfections, leave a unique and stable mark on each acquired image, i.e., the photo-response non-uniformity (PRNU) pattern~\cite{lukas2006digital,chen2008determining}. Likewise, postprocessing fingerprints come from the specific in-camera postprocessing suite (demosaicking, compression, etc.) during each image acquisition procedure~\cite{cozzolino2018noiseprint}. \ning{Recently, Marra \etal~\cite{marra2019gans} visualize GAN fingerprints based on PRNU, and show their application to GAN source identification. We replace their hand-crafted fingerprint formulation with a learning-based one, decoupling model fingerprint from image fingerprint, and show superior performances in a variety of experimental conditions.} 

\vspace{-16pt}
\paragraph{Digital watermarking.} Digital watermarking is a complementary forensics technique for image authentication~\cite{swanson1998multimedia,langelaar2000watermarking,Saini2014ASO}. It involves embedding artificial watermarks in images. It can be used to reveal image source and ownership so as to protect their copyright. It has been shown that neural networks can also be actively watermarked during training~\cite{zhang2018protecting,uchida2017embedding}. In such models, a characteristic pattern can be built into the learned representation but with a trade-off between watermarking accuracy and the original performance. However, such watermarking has not been studied for GANs. In contrast, we utilize inherent fingerprints for image attribution without any extra embedding burden or quality deterioration.

\section{Fingerprint learning for image attribution}

Inspired by the prior works on digital fingerprints ~\cite{lukas2006digital,cozzolino2018noiseprint}, we introduce the concepts of GAN model fingerprint and image fingerprint. Both are simultaneously learned from an image attribution task.

\vspace{-16pt}
\paragraph{Model fingerprint.} Each GAN model is characterized by many parameters: training dataset distribution, network architecture, loss design, optimization strategy, and hyper-parameter settings. Because of the non-convexity of the objective function and the instability of adversarial equilibrium between the generator and discriminator in GANs, the values of model weights are sensitive to their random initializations and do not converge to the same values during each training. This indicates that even though two well-trained GAN models may perform equivalently, they generate high-quality images differently. This suggests the existence and uniqueness of GAN fingerprints. We define the model fingerprint per GAN instance as a reference vector, such that it consistently interacts with all its generated images. In a specifically designed case, the model fingerprint can be an RGB image the same size as its generated images. See Section~\ref{sec:fingerprint}.

\vspace{-16pt}
\paragraph{Image fingerprint.} GAN-generated images are the outcomes of a large number of fixed filtering and non-linear processes, which generate common and stable patterns within the same GAN instances but are distinct across different GAN instances. That suggests the existence of image fingerprints and attributability towards their GAN sources. We introduce the fingerprint per image as a feature vector encoded from that image. In a specifically designed case, an image fingerprint can be an RGB image the same size as the original image. See Section~\ref{sec:fingerprint}.

\subsection{Attribution network}\label{sec:attribution}
Similar to the authorship attribution task in natural language processing~\cite{stamatatos2009survey,afroz2014doppelganger}, we train an attribution classifier that can predict the source of an image: real or from a GAN model. 

\begin{figure}
\centering
\subfigure[]{\includegraphics[width=0.10\textwidth]{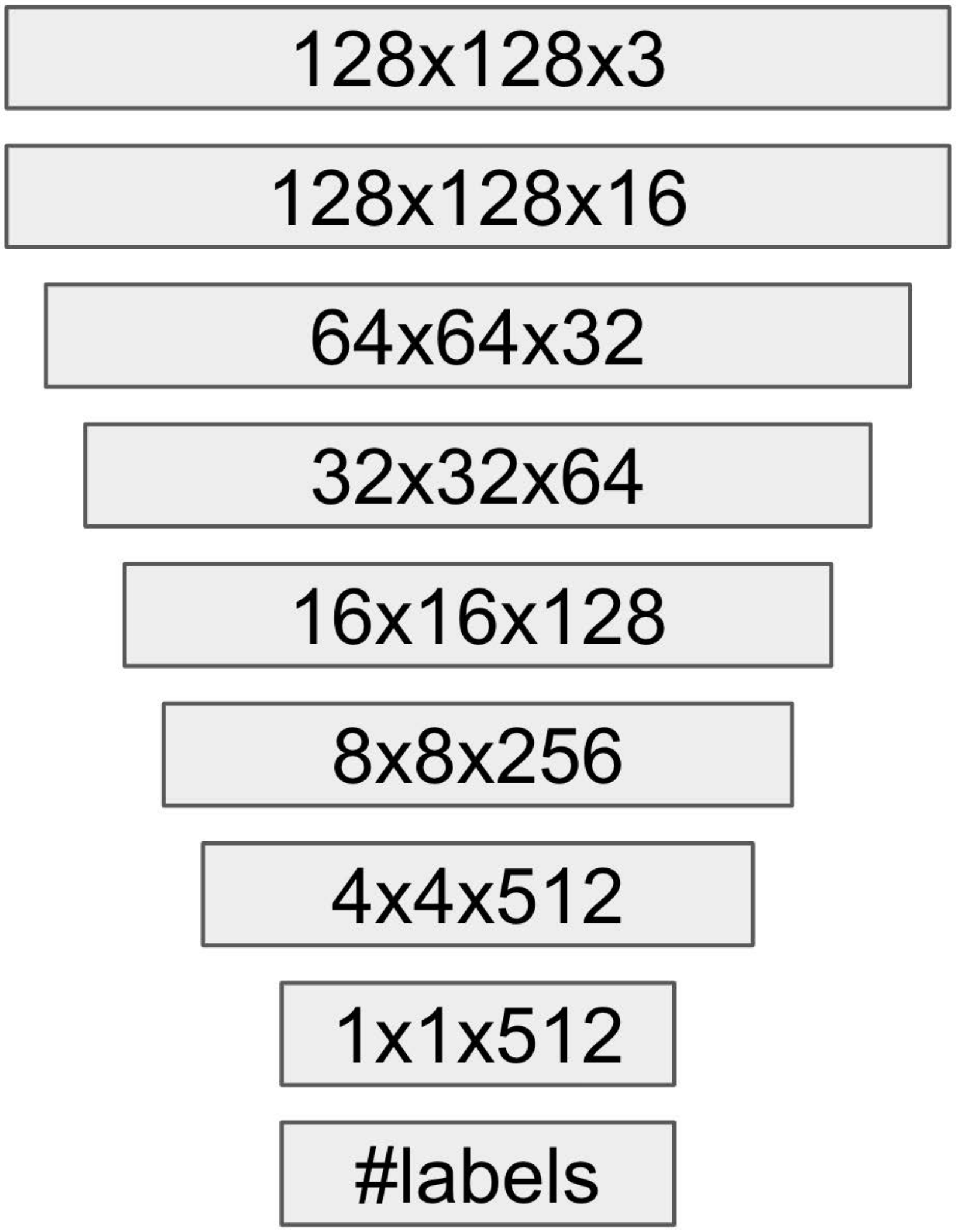}\label{fig:architecture}}
\hspace{0.1cm}
\subfigure[]{\includegraphics[width=0.112\textwidth]{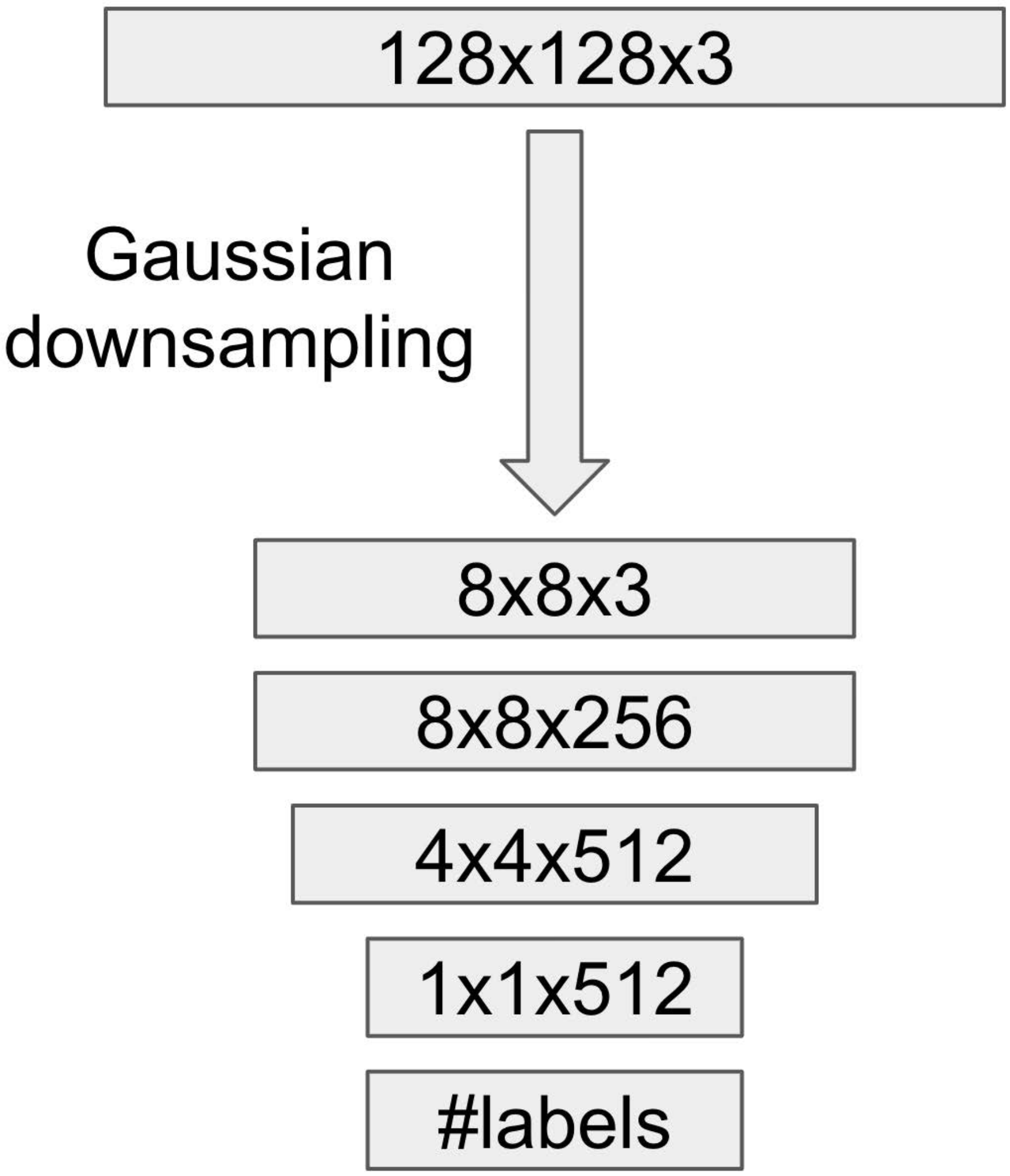}\label{fig:predownsampling}}
\hspace{0.1cm}
\subfigure[]{\includegraphics[width=0.112\textwidth]{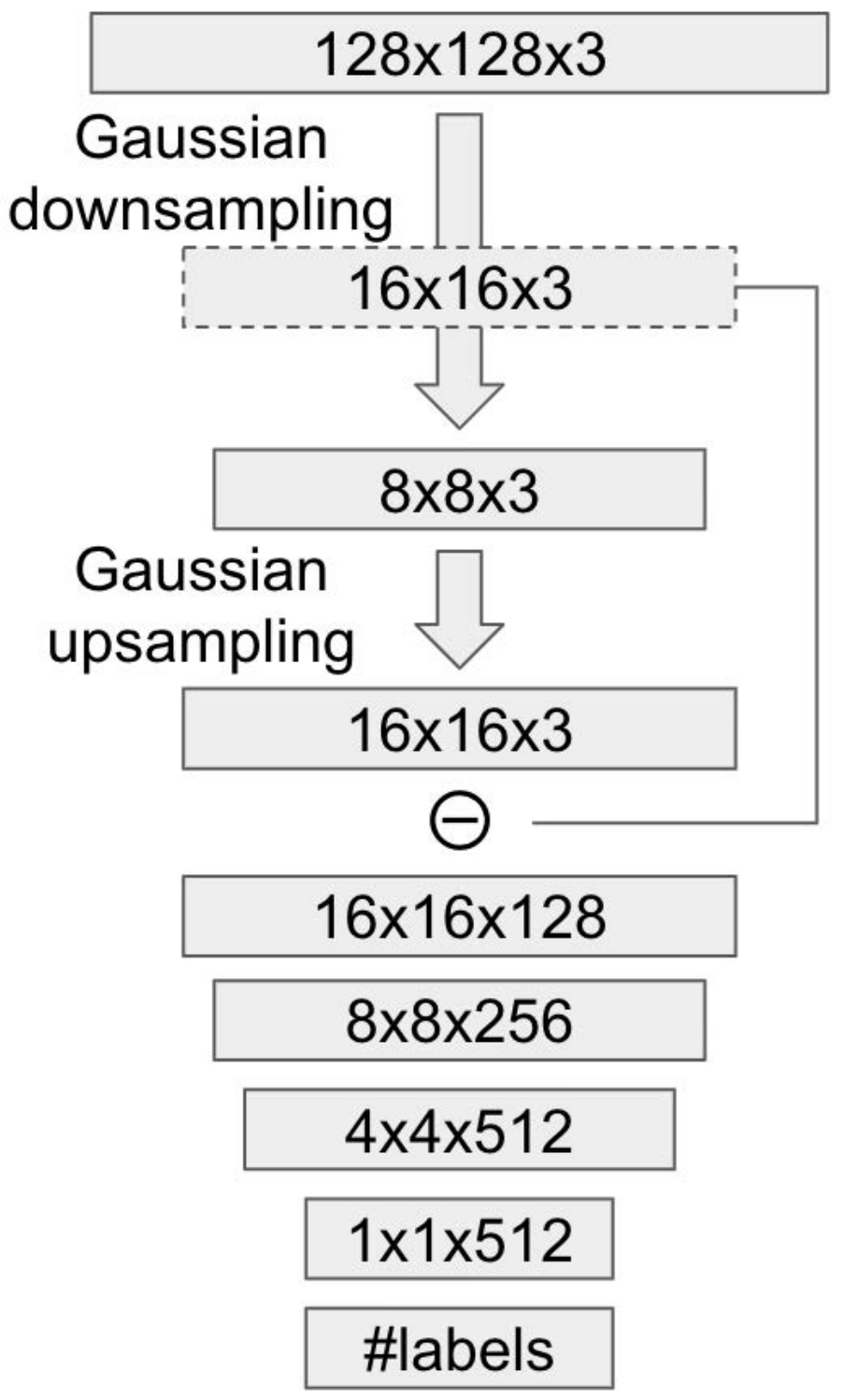}\label{fig:predownsampling_residual}}
\hspace{0.1cm}
\subfigure[]{\includegraphics[width=0.10\textwidth]{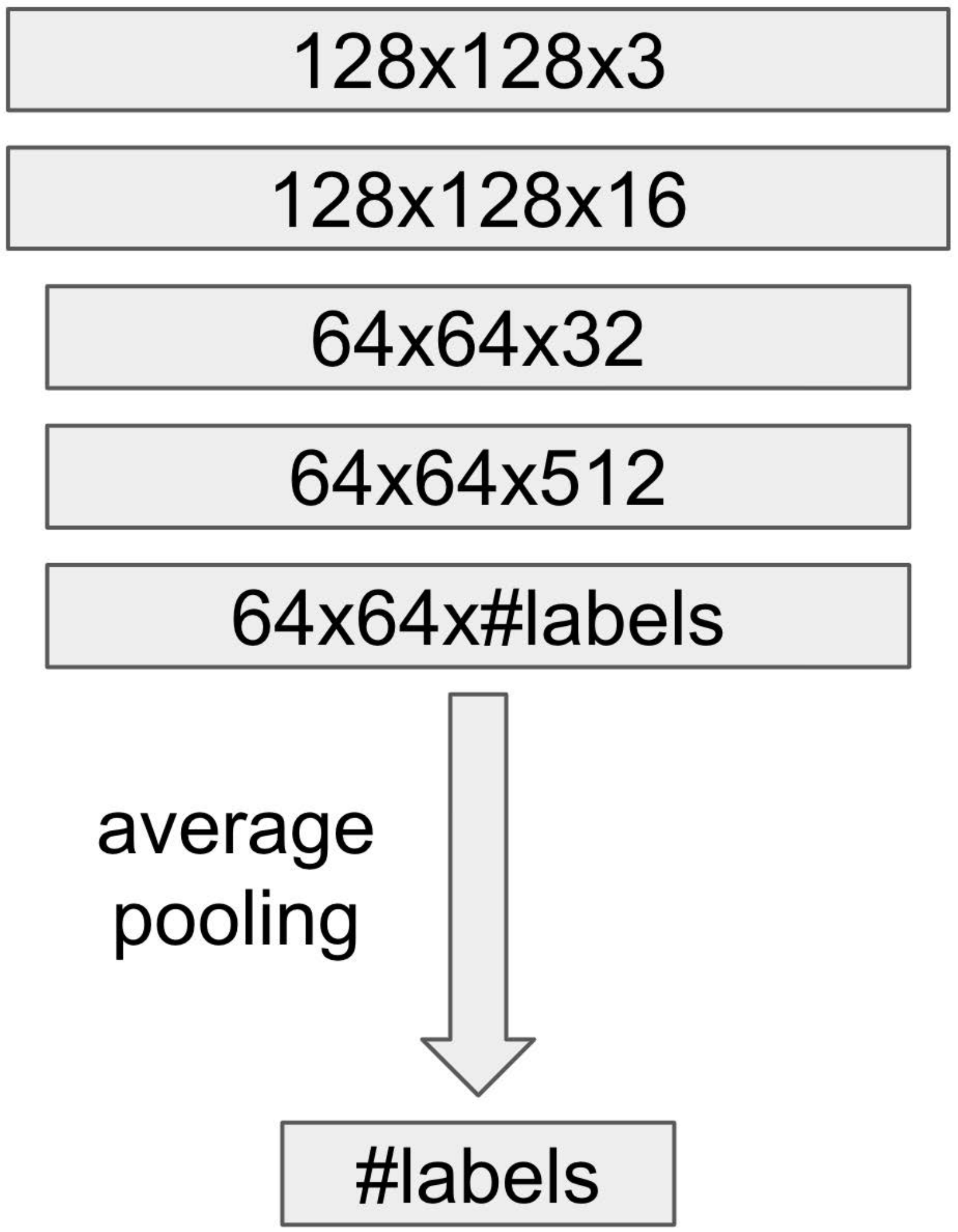}\label{fig:postpooling}}
\caption{Different attribution network architectures. Tensor representation is specified by two spatial dimensions followed by the number of channels. The network is trained to minimize cross-entropy classification loss. (a) Attribution network. (b) Pre-downsampling network example that downsamples input image to $8\times8$ before convolution. (c) Pre-downsampling residual network example that extracts the residual component between $16\times16$ and $8\times8$ resolutions. (d) Post-pooling network example that starts average pooling at $64\times64$ resolution.}
\vspace{-12pt}
\end{figure}

We approach this using a deep convolutional neural network supervised by image-source pairs $\{(I, y)\}$ where $I\sim \mathbb{I}$ is sampled from an image set and $y\in\mathbb{Y}$ is the source ground truth belonging to a finite set. That set is composed of pre-trained GAN instances plus the real world. Figure~\ref{fig:architecture} depicts an overview of our attribution network. 

We implicitly represent image fingerprints as the final classifier features (the $1\times1\times512$ tensor before the final fully connected layer) and represent GAN model fingerprints as the corresponding classifier parameters (the $1\times1\times512$ weight tensor of the final fully connected layer). 

\ning{Why is it necessary to use such an external classifier when GAN training already provides a discriminator? The discriminator learns a hyperplane in its own embedding space to distinguish generated images from real ones. Different embedding spaces are not aligned. In contrast, the proposed classifier necessarily learns a unified embedding space to distinguish generated images from different GAN instances or from real images.}

Note that our motivation to investigate ``white-box" GANs subject to known parameters is to validate the attributability along different GAN parameter dimensions. In practice, our method also applies to ``black-box" GAN API services. The only required supervision is the source label of an image. We can simply query different services, collect their generated images, and label them by service indices. Our classifier would test image authenticity by predicting if an image is sampled from the desired service. We also test service authenticity by checking if most of their generated images have the desired source prediction.

\subsection{Component analysis networks}\label{sec:pooling}
In order to analyze which image components contain fingerprints, we propose three variants of the network.

\vspace{-16pt}
\paragraph{Pre-downsampling network.} We propose to test whether fingerprints and attribution can be derived from different frequency bands. We investigate attribution performance w.r.t. downsampling factor. Figure~\ref{fig:predownsampling} shows an architecture example that extracts low-frequency bands. We replace the trainable convolution layers with our Gaussian downsampling layers from the input end and systematically control at which resolution we stop such replacement. 

\vspace{-16pt}
\paragraph{Pre-downsampling residual network.} Complementary to extracting low-frequency bands, Figure~\ref{fig:predownsampling_residual} shows an architecture example that extracts a residual high-frequency band between one resolution and its factor-2 downsampled resolution. \ning{It is reminiscent of a Laplacian Pyramid~\cite{burt1983laplacian}.} We systematically vary the resolution at which we extract such residual.

\vspace{-16pt}
\paragraph{Post-pooling network.} We propose to test whether fingerprints and attribution can be derived locally based on patch statistics. We investigate attribution performance w.r.t. patch size. Figure~\ref{fig:postpooling} shows an architecture example. Inspired by PatchGAN~\cite{isola2017image}, we regard a ``pixel" in a neural tensor as the feature representation of a local image patch covered by the receptive field of that ``pixel". Therefore, post-pooling operations count for patch-based neural statistics. Earlier post-pooling corresponds to a smaller patch size. We systematically vary at which tensor resolution we start this pooling in order to switch between more local and more global patch statistics.

\subsection{Fingerprint visualization}\label{sec:fingerprint}
\begin{figure}[t]
\centering
\includegraphics[width=\linewidth]{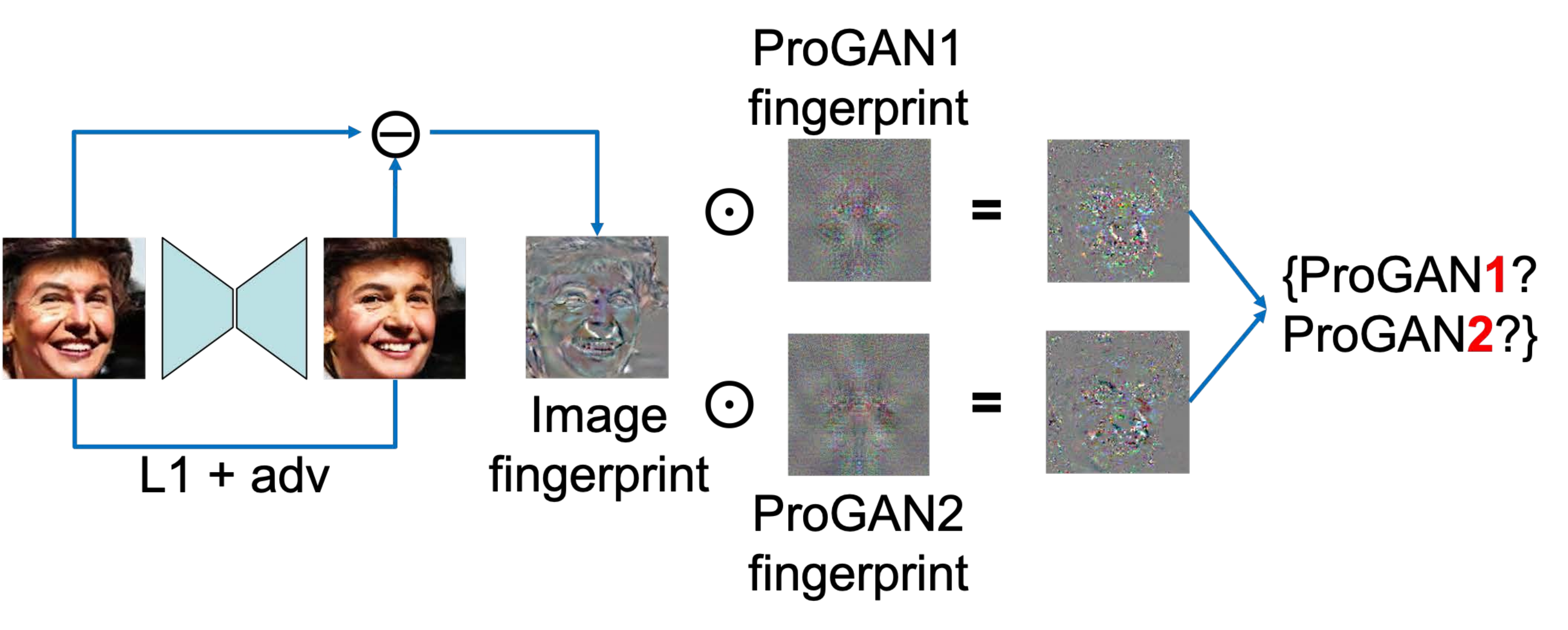}
\caption{Fingerprint visualization diagram. We train an AutoEncoder and GAN fingerprints end-to-end. $\odot$ indicates pixel-wise multiplication of two normalized images.}
\label{fig:fingerprint_diagram}
\vspace{-12pt}
\end{figure}

Alternatively to our attribution network in Section~\ref{sec:attribution} where fingerprints are implicitly represented in the feature domain, we describe a model similar in spirit to Marra \etal~\cite{marra2019gans} to explicitly represent them in the image domain. \ning{But in contrast to their hand-crafted PRNU-based representation, we modify our attribution network architecture and learn fingerprint images from image-source pairs $(\{I, y\})$. We also decouple the representation of model fingerprints from image fingerprints.} Figure~\ref{fig:fingerprint_diagram} depicts the fingerprint visualization model.

Abstractly, we learn to map from input image to its fingerprint image. But without fingerprint supervision, we choose to ground the mapping based on a reconstruction task with an AutoEncoder. We then define the reconstruction residual as the image fingerprint. We simultaneously learn a model fingerprint for each source (each GAN instance plus the real world), such that the correlation index between one image fingerprint and each model fingerprint serves as softmax logit for classification.

\begin{figure*}[t]
\centering
\subfigure[CelebA real data]{\includegraphics[height=3cm]{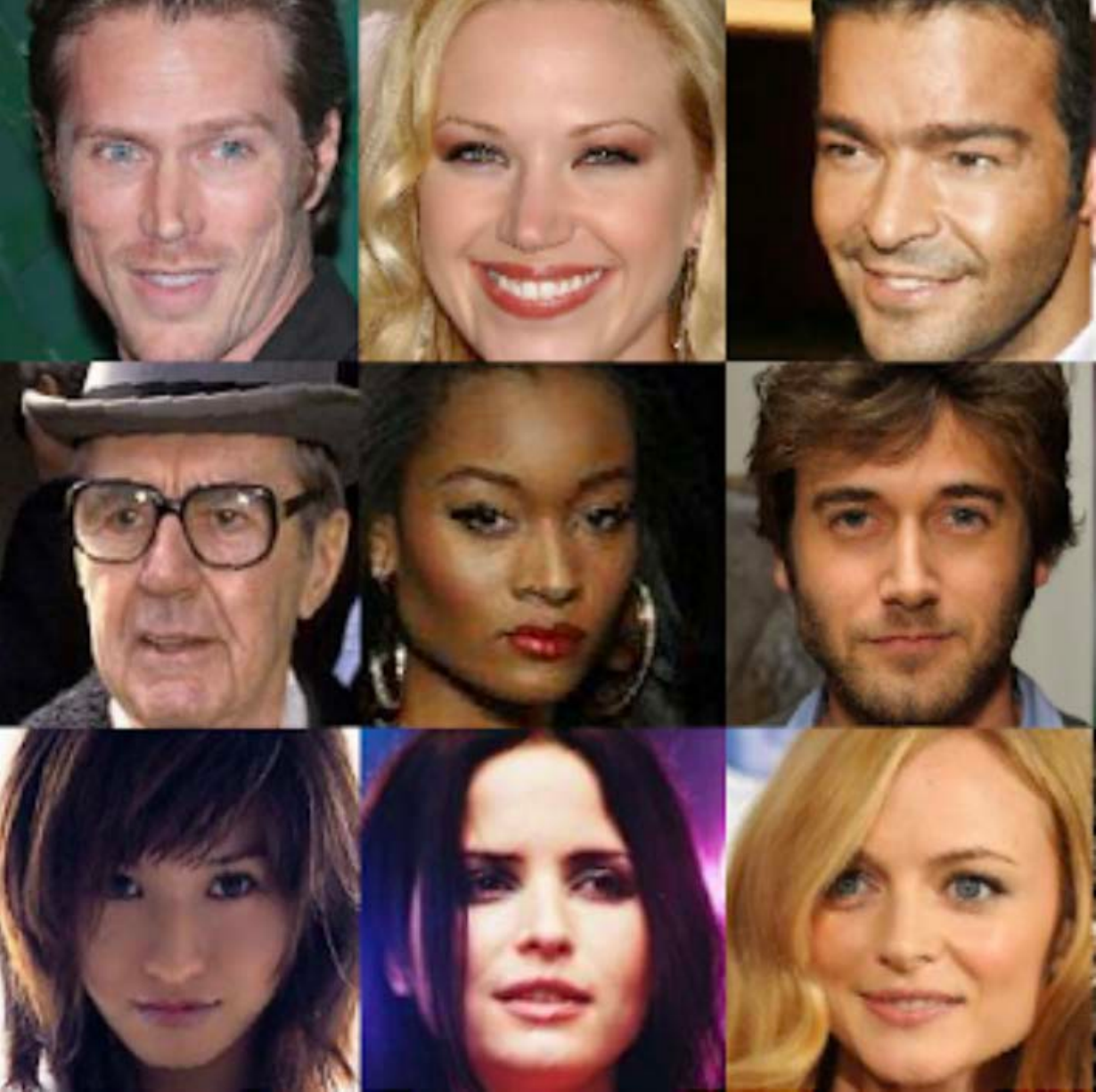}}
\hspace{0.1cm}
\subfigure[ProGAN]{\includegraphics[height=3cm]{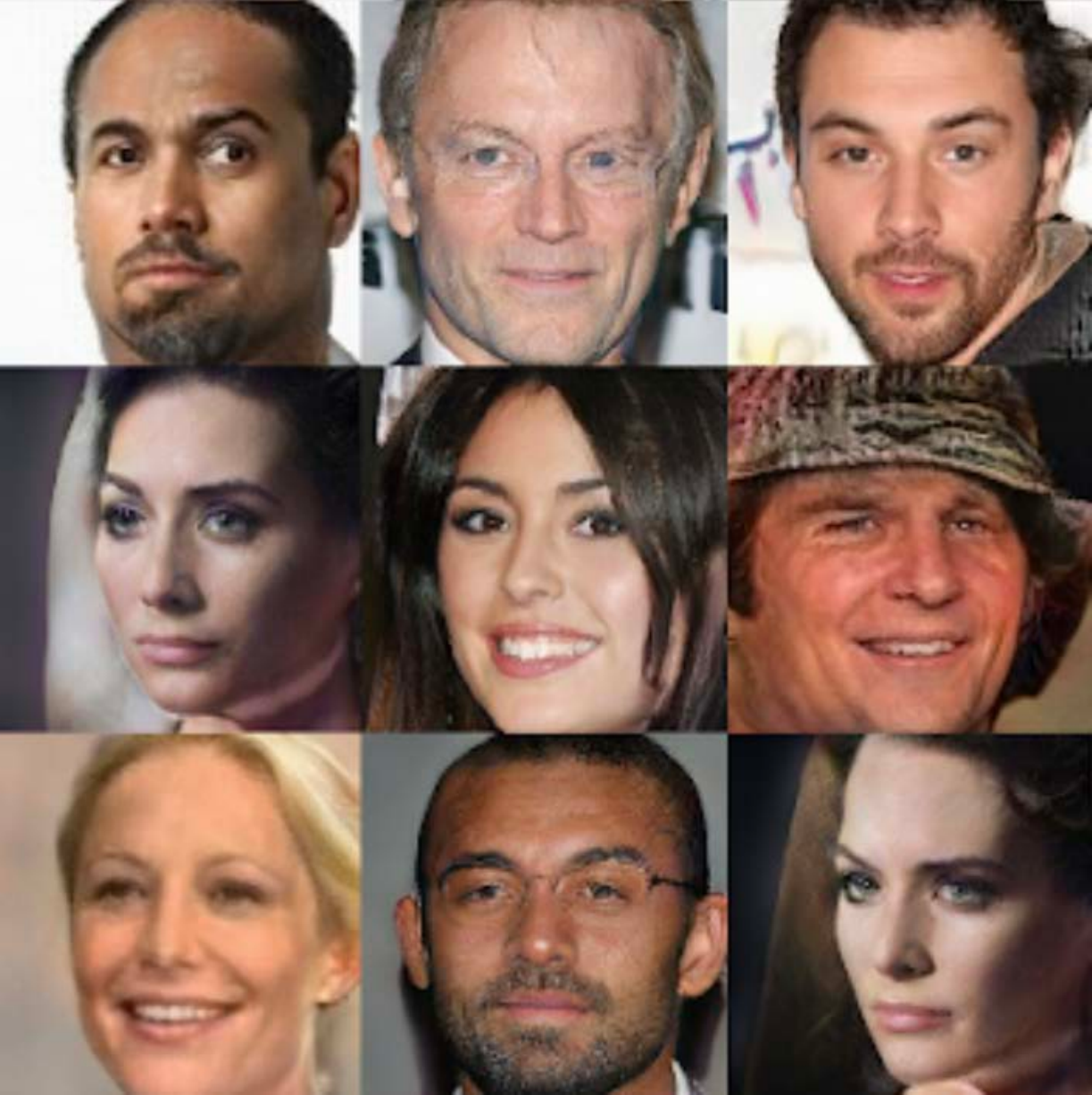}}
\hspace{0.1cm}
\subfigure[SNGAN]{\includegraphics[height=3cm]{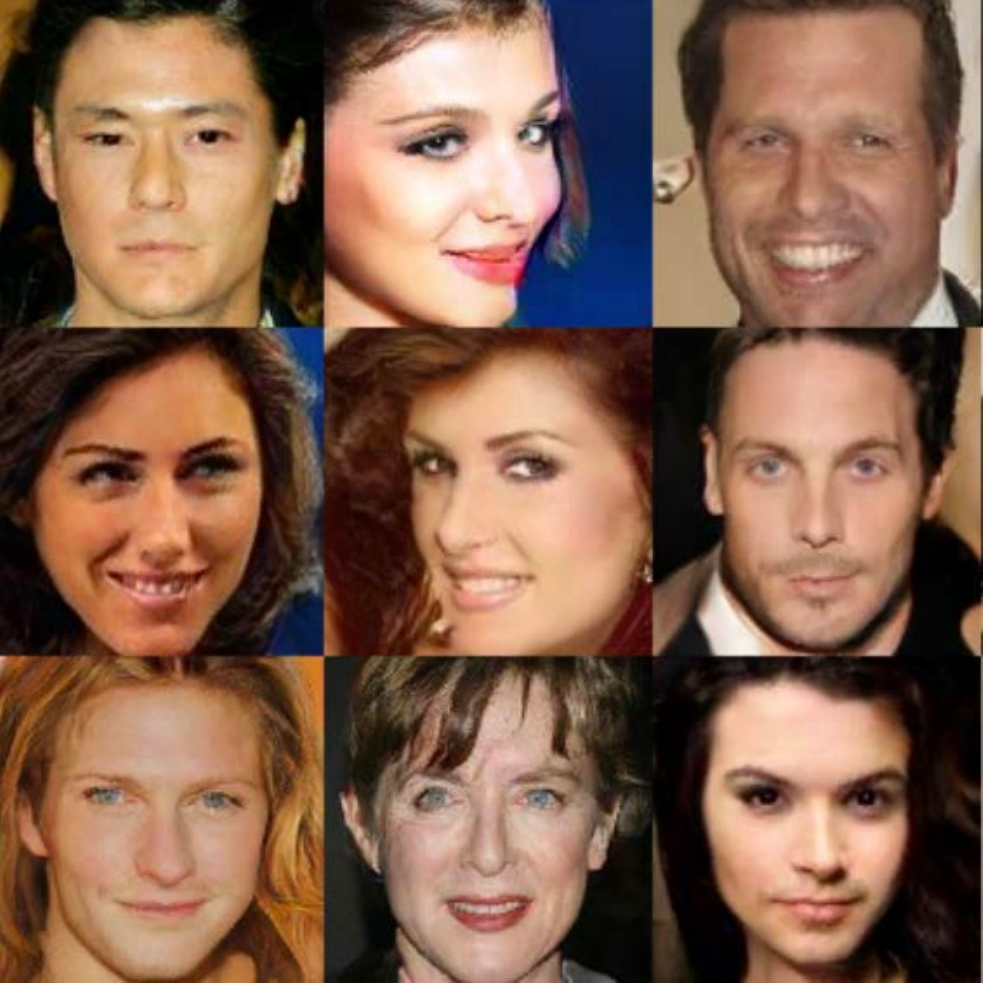}}
\hspace{0.1cm}
\subfigure[CramerGAN]{\includegraphics[height=3cm]{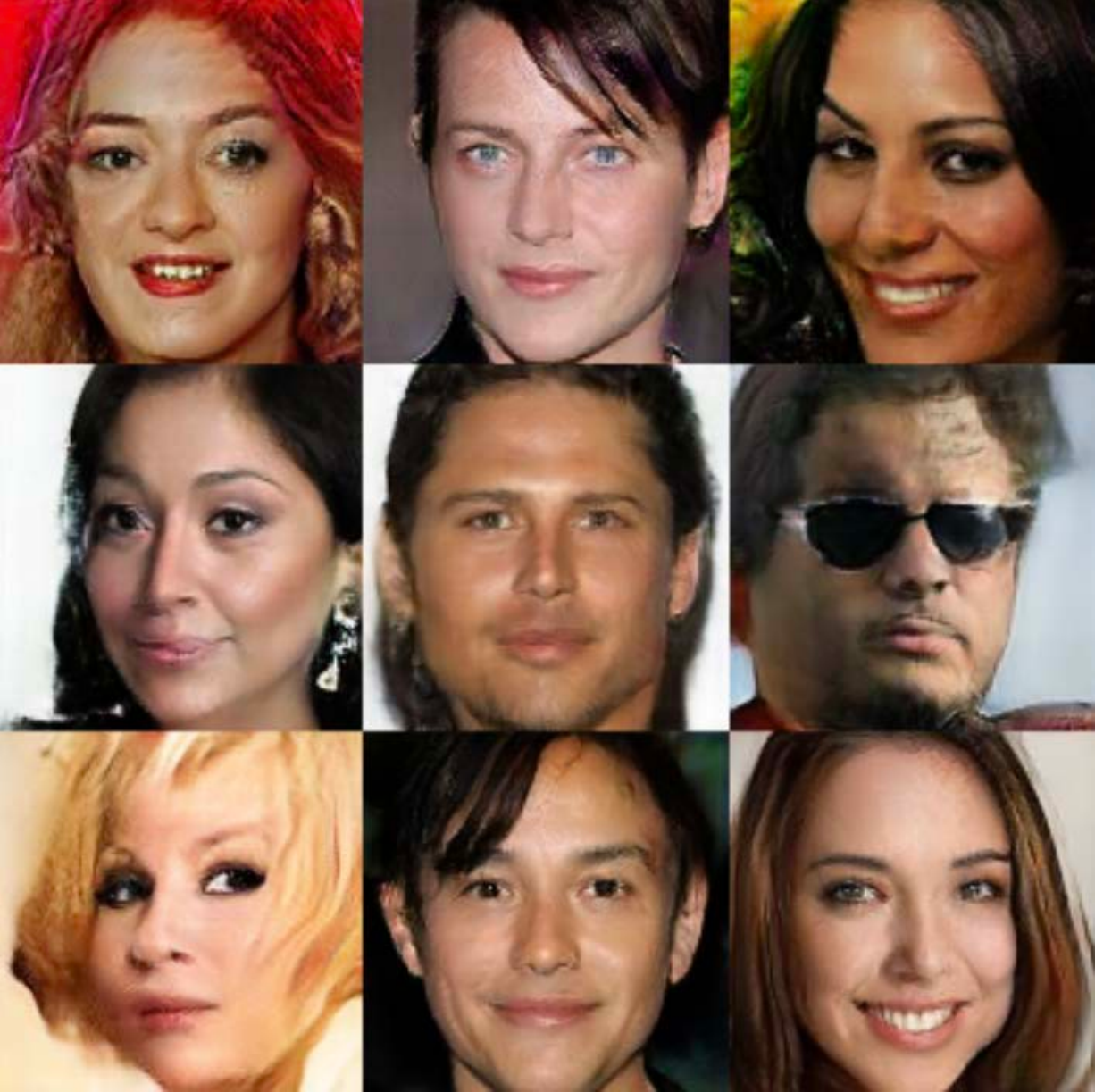}}
\hspace{0.1cm}
\subfigure[MMDGAN]{\includegraphics[height=3cm]{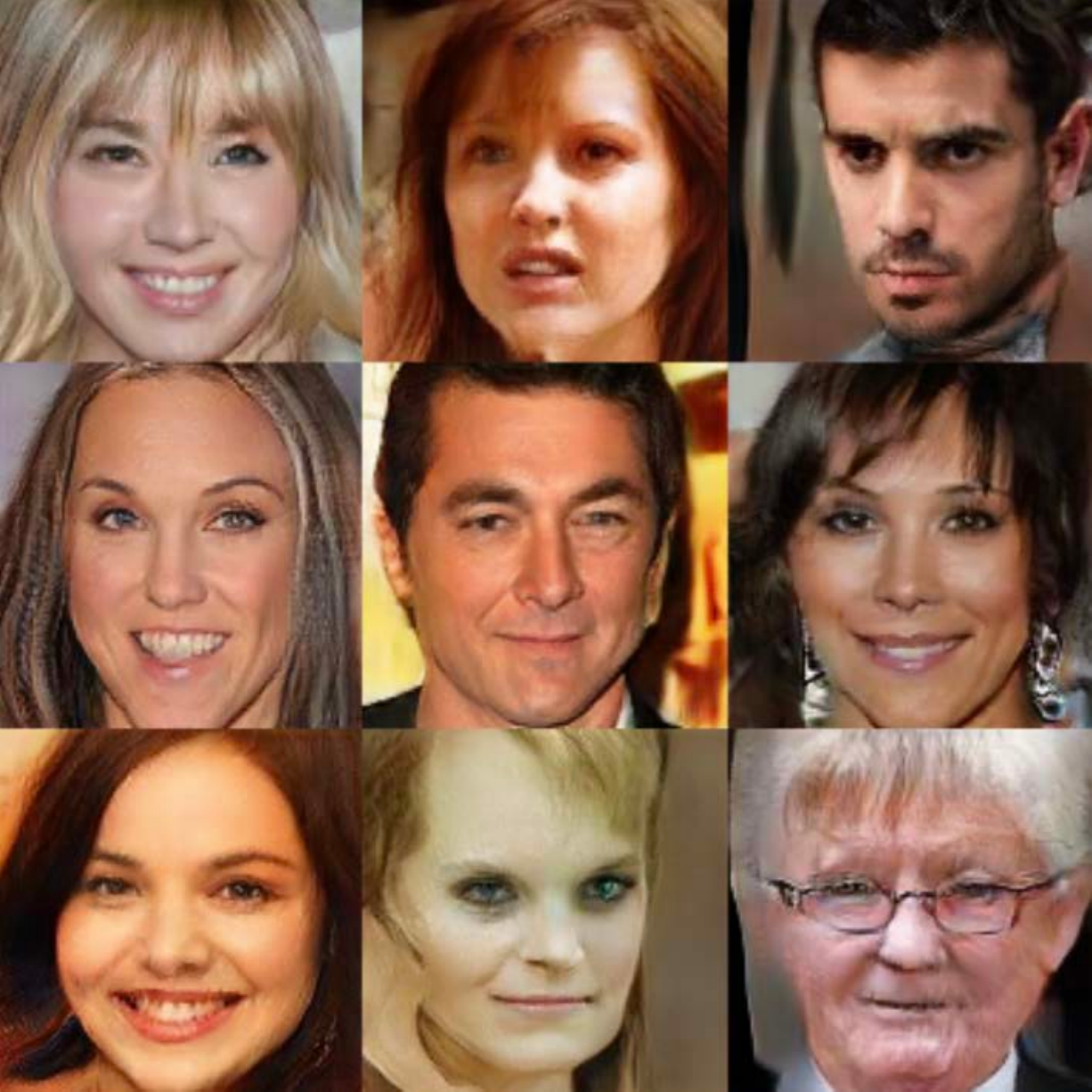}}
\hspace{0.1cm}
\caption{Face samples from difference sources.}
\label{fig:data}
\vspace{-12pt}
\end{figure*}

Mathematically, given an image-source pair $(I, y)$ where $y\in\mathbb{Y}$ belongs to the finite set $\mathbb{Y}$ of GAN instances plus the real world, we formulate a reconstruction mapping $R$ from $I$ to $R(I)$. We ground our reconstruction based on pixel-wise $L_1$ loss plus adversarial loss:
\begin{equation}
L_{\textit{pix}}(I) = ||R(I) - I||_1
\end{equation}
\begin{equation}
L_{\textit{adv}}(I) = D_{\textit{rec}}\big(R(I)\big) - D_{\textit{rec}}\big(I\big) + \textit{GP}\big(R(I), I | D_{\textit{rec}}\big)
\end{equation}
where $D_{\textit{rec}}$ is an adversarially trained discriminator, and $\textit{GP}(\cdot)$ is the gradient penalty regularization term defined in~\cite{gulrajani2017improved}. 

We then explicitly define image fingerprint $F_{\textit{im}}^I$ as the reconstruction residual $F_{\textit{im}}^I = R(I) - I$.

We further explicitly define model fingerprint $F_{\textit{mod}}^y$ as freely trainable parameters with the same size as $F_{\textit{im}}^I$, such that $\textit{corr}(F_{\textit{im}}^I, F_{\textit{mod}}^y)$, the correlation index between $F_{\textit{im}}^I$ and $F_{\textit{mod}}^y$, is maximized over $\mathbb{Y}$. This can be formulated as the softmax logit for the cross-entropy classification loss supervised by the source ground truth:
\begin{equation}
L_{\textit{cls}}(I, y) = -\log\frac{\textit{corr}(F_{\textit{im}}^I, F_{\textit{mod}}^{y})}{\sum_{\hat{y}\in\mathbb{Y}}\textit{corr}(F_{\textit{im}}^I, F_{\textit{mod}}^{\hat{y}})}
\end{equation}
where $\textit{corr}(A, B) = \hat{A}\odot\hat{B}$, $\hat{A}$ and $\hat{B}$ are the zero-mean, unit-norm, and vectorized version of images $A$ and $B$, and $\odot$ is the inner product operation.

Our final training objective is
\begin{equation}
\min_{R, \{F_{\textit{mod}}^{\tilde{y}} | \tilde{y}\in\mathbb{Y}\}}\;\max_{D_{\textit{rec}}}\;\underset{\{(I, y)\}}{\mathbb{E}}\;(\lambda_1 L_{\textit{pix}} + \lambda_2 L_{\textit{adv}} + \lambda_3 L_{\textit{cls}})
\end{equation}
where $\lambda_1 = 20.0$, $\lambda_2 = 0.1$, and $\lambda_3 = 1.0$ are used to balance the order of magnitude of each loss term, which are not sensitive to dataset \ning{and are fixed}.

Note that this network variant is used to better visualize and interpret the effectiveness of image attribution. However, it introduces extra training complexity and thus is not used if we only focus on attribution. 

\section{Experiments}\label{sec:exp}
We discuss the experimental setup in Section~\ref{sec:setup}. From Section~\ref{sec:attribution_exp} to \ref{sec:fingerprint_exp}, we explore the four research questions discussed in the Introduction.

\subsection{Setup} \label{sec:setup}
\paragraph{Datasets}. We employ CelebA human face dataset~\cite{liu2015faceattributes} and LSUN bedroom scene dataset~\cite{yu2015lsun}, both containing $20,000$ real-world RGB images.


\vspace{-16pt}
\paragraph{GAN models.} We consider four recent state-of-the-art GAN architectures: ProGAN~\cite{karras2018progressive}, SNGAN~\cite{miyato2018spectral}, CramerGAN~\cite{bellemare2017cramer}, and MMDGAN~\cite{binkowski2018demystifying}. Each model is trained from scratch with their default settings except we fix the number of training epochs to $240$ and fix the output size of a generator to $128 \times 128 \times 3$.

\vspace{-16pt}
\paragraph{Baseline methods.} \ning{Given real-world datasets and four pre-trained GAN models,} we compare with three baseline classification methods: k-nearest-neighbor (kNN) on raw pixels, Eigenface~\cite{sirovich1987low}, and the very recent PRNU-based fingerprint method from Marra \etal~\cite{marra2019gans}.

\vspace{-16pt}
\paragraph{Evaluation.} We use classification accuracy to evaluate image attribution performance.

In addition, we use the ratio of inter-class and intra-class Fr\'{e}chet Distance ~\cite{dowson1982frechet}, denoted as FD ratio, to evaluate the distinguishability of a feature representation across classes. The larger the ratio, the more distinguishable the feature representation across sources. \ning{See supplementary material for more detail.} We compare our fingerprint features to image inception features~\cite{salimans2016improved}. \ning{The FD of inception features is also known as FID for GAN evaluation~\cite{heusel2017gans}. Therefore, the FD ratio of inception features can serve as a reference to show how challenging it is to attribute high-quality GAN-generated images manually or without fingerprint learning.} 



\subsection{\ning{Existence and uniqueness:} which GAN parameters differentiate image attribution?} \label{sec:attribution_exp}
We consider GAN architecture, training set, and initialization seed respectively by varying one type of parameter and keeping the other two fixed.

\vspace{-16pt}
\paragraph{Different architectures.} First, we leverage all the real images to train ProGAN, SNGAN, CramerGAN, and MMDGAN separately. For the classification task, we configure training and testing sets with $5$ classes: $\{$\textit{real}, \textit{ProGAN}, \textit{SNGAN}, \textit{CramerGAN}, \textit{MMDGAN}$\}$. We randomly collect $100,000$ images from each source for classification training and another $10,000$ images from each source for testing. We show face samples from each source in Figure~\ref{fig:data} and bedroom samples in the supplementary material. Table~\ref{table:eval_arch} shows that we can effectively differentiate GAN-generated images from real ones and attribute generated images to their sources, just using a regular CNN classifier. There do exist unique fingerprints in images that differentiate GAN architectures, even though it is far more challenging to attribute those images manually or through inception features~\cite{salimans2016improved}.

\vspace{-16pt}
\paragraph{Different GAN training sets.} We further narrow down the investigation to GAN training sets. From now we only focus on ProGAN plus real dataset. We first randomly select a base real subset containing $100,000$ images, denoted as \textit{real\_subset\_diff\_0}. We then randomly select $10$ other real subsets also containing $100,000$ images, denoted as \textit{real\_subset\_diff\_\#i}, where $i$ $\in$ $\{1$, $10$, $100$, $1000$, $10000$, $20000$, $40000$, $60000$, $80000$, $100000\}$ indicates the number of images that are not from the base subset. We collect such sets of datasets to explore the relationship between attribution performance and GAN training set overlaps.

For each \textit{real\_subset\_diff\_\#i}, we separately train a ProGAN model and query $100,000$ images for classifier training and another $10,000$ images for testing, labeled as \textit{ProGAN\_subset\_diff\_\#i}. In this setup of $\{$\textit{real}, \textit{ProGAN\_subset\_diff\_\#i}$\}$, we show the performance evaluation in Table~\ref{table:subset}. Surprisingly, we find that attribution performance remains equally high regardless of the amount of GAN training set overlap. Even GAN training sets that differ in just one image can lead to distinct GAN instances. That indicates that one-image mismatch during GAN training results in a different optimization step in one iteration and finally results in distinct fingerprints. That motivates us to investigate the attribution performance among GAN instances that were trained with identical architecture and dataset but with different random initialization seeds.

\begin{table}[!t]
\center
\small
\caption{Evaluation on $\{$\textit{real}, \textit{ProGAN}, \textit{SNGAN}, \textit{CramerGAN}, \textit{MMDGAN}$\}$. The best performance is highlighted in \textbf{bold}.}
\begin{tabular}{clcc}
\toprule
& & CelebA & LSUN \tabularnewline
\midrule
\multicolumn{1}{c}{ } & kNN & 28.00 & 36.30 \tabularnewline
\multicolumn{1}{c}{Accuracy} & Eigenface~\cite{sirovich1987low} & 53.28 & - \tabularnewline
\multicolumn{1}{c}{(\%)} & PRNU~\cite{marra2019gans} & 86.61 & 67.84 \tabularnewline
\multicolumn{1}{c}{ } & Ours & \textbf{99.43} & \textbf{98.58} \tabularnewline
\midrule
\multicolumn{1}{c}{FD ratio} & Inception~\cite{salimans2016improved} & 2.36 & 5.27 \tabularnewline
\multicolumn{1}{c}{} & Our fingerprint & \textbf{454.76} & \textbf{226.59} \tabularnewline
\bottomrule
\end{tabular}
\label{table:eval_arch}
\vspace{-6pt}
\end{table}

\begin{table}[!t]
\center
\small
\caption{Evaluation on $\{$\textit{real}, \textit{ProGAN\_subset\_diff\_\#i}$\}$. The best performance is highlighted in \textbf{bold}.}
\begin{tabular}{clcc}
\toprule
& & CelebA & LSUN \tabularnewline
\midrule
\multicolumn{1}{c}{ } & kNN & 11.46 & 10.72 \tabularnewline
\multicolumn{1}{c}{Accuracy} & Eigenface~\cite{sirovich1987low} & 27.98 & - \tabularnewline
\multicolumn{1}{c}{(\%)} & PRNU~\cite{marra2019gans} & 92.28 & 70.55 \tabularnewline
\multicolumn{1}{c}{ } & Ours & \textbf{99.50} & \textbf{97.66} \tabularnewline
\midrule
\multicolumn{1}{c}{FD ratio} & Inception~\cite{salimans2016improved} & 1.08 & 1.64 \tabularnewline
\multicolumn{1}{c}{} & Our fingerprint & \textbf{111.41} & \textbf{39.96} \tabularnewline
\bottomrule
\end{tabular}
\label{table:subset}
\vspace{-6pt}
\end{table}

\begin{table}[!t]
\center
\small
\caption{Evaluation on $\{$\textit{real}, \textit{ProGAN\_seed\_v\#i}$\}$. The best performance is highlighted in \textbf{bold}. ``Our visNet" row indicates our fingerprint visualization network described in Section~\ref{sec:fingerprint} and evaluated in Section~\ref{sec:fingerprint_exp}.}
\begin{tabular}{clcc}
\toprule
& & CelebA & LSUN \tabularnewline
\midrule
\multicolumn{1}{c}{ } & kNN & 10.88 & 10.58 \tabularnewline
\multicolumn{1}{c}{Accuracy} & Eigenface~\cite{sirovich1987low} & 23.12 & - \tabularnewline
\multicolumn{1}{c}{(\%)} & PRNU~\cite{marra2019gans} & 89.40 & 69.73 \tabularnewline
\multicolumn{1}{c}{ } & Ours & \textbf{99.14} & \textbf{97.04} \tabularnewline
\multicolumn{1}{c}{ } & Our visNet & 97.07 & 96.58 \tabularnewline
\midrule
\multicolumn{1}{c}{FD ratio} & Inception~\cite{salimans2016improved} & 1.10 & 1.29 \tabularnewline
\multicolumn{1}{c}{} & Our fingerprint & \textbf{80.28} & \textbf{36.48} \tabularnewline
\bottomrule
\end{tabular}
\label{table:seed}
\vspace{-6pt}
\end{table}

\vspace{-16pt}
\paragraph{Different initialization seeds.} We next investigate the impact of GAN training initialization on image attributability. We train $10$ ProGAN instances with the entire real dataset and with different initialization seeds. We sample $100,000$ images for classifier training and another $10,000$ images for testing. In this setup of $\{$\textit{real}, \textit{ProGAN\_seed\_v\#i}$\}$ where $i$ $\in$ $\{1$, ..., $10\}$, we show the performance evaluation in Table~\ref{table:seed}. We conclude that it is the difference in optimization (e.g., caused by different randomness) that leads to attributable fingerprints. In order to verify our experimental setup, we ran sanity checks. For example, two identical ProGAN instances trained with the same seed remain indistinguishable and result in random-chance attribution performance.

\subsection{\ning{Persistence:} which image components contain fingerprints for attribution?}\label{sec:pooling_exp}
We systematically explore attribution performance w.r.t. image components in different frequency bands or with different patch sizes. We also investigate possible performance bias from GAN artifacts.

\vspace{-16pt}
\paragraph{Different frequencies.}
We investigate if band-limited images carry effective fingerprints for attribution. We separately apply the proposed pre-downsampling network and pre-downsampling residual network for image attribution. Given the setup of $\{$\textit{real}, \textit{ProGAN\_seed\_v\#i}$\}$, Table~\ref{table:predownsampling} shows the classification accuracy w.r.t. downsampling factors. We conclude that (1) a wider frequency band carries more fingerprint information for image attribution, (2) the low-frequency and high-frequency components (even at the resolution of $8 \times 8$) individually carry effective fingerprints and result in attribution performance better than random, \ning{and (3) at the same resolution, high-frequency components carry more fingerprint information than low-frequency components.}

\begin{table}[!t]
\center
\small
\caption{Classification accuracy (\%) of our network w.r.t. downsampling factor on low-frequency or high-frequency components of $\{$\textit{real}, \textit{ProGAN\_seed\_v\#i}$\}$. ``L-f" column indicates the low-frequency components and represents the performances from the pre-downsampling network. ``H-f" column indicates the high-frequency components and represents the performances from the pre-downsampling residual network.}
\begin{tabular}{cccccc}
\toprule
Downsample & Res- & \multicolumn{2}{c}{CelebA} & \multicolumn{2}{c}{LSUN} \tabularnewline
factor & olution & L-f & H-f & L-f & H-f \tabularnewline
\midrule
1 & $128^2$ & 99.14 & 99.14 & 97.04 & 97.04 \tabularnewline
2 & $64^2$ & 98.74 & 98.64 & 96.78 & 96.84 \tabularnewline
4 & $32^2$ & 95.50 & 98.52 & 91.08 & 96.04 \tabularnewline
8 & $16^2$ & 87.20 & 92.90 & 83.02 & 91.58 \tabularnewline
16 & $8^2$ & 67.44 & 78.74 & 63.80 & 80.58 \tabularnewline
32 & $4^2$ & 26.58 & 48.42 & 28.24 & 54.50 \tabularnewline
\bottomrule
\end{tabular}
\label{table:predownsampling}
\vspace{-6pt}
\end{table}

\begin{table}[!t]
\center
\small
\caption{Classification accuracy (\%) of our network w.r.t. patch size on $\{$\textit{real}, \textit{ProGAN\_seed\_v\#i}$\}$.}
\begin{tabular}{cccc}
\toprule
Pooling starts at & Patch size & CelebA & LSUN \tabularnewline
\midrule
$4^2$ & $128^2$ & 99.34 & 97.44 \tabularnewline
$8^2$ & $108^2$ & 99.32 & 96.30 \tabularnewline
$16^2$ & $52^2$ & 99.30 & 95.94 \tabularnewline
$32^2$ & $24^2$ & 99.24 & 88.36 \tabularnewline
$64^2$ & $10^2$ & 89.60 & 18.26 \tabularnewline
$128^2$ & $3^2$ & 13.42 & 17.10 \tabularnewline
\bottomrule
\end{tabular}
\label{table:postpooling}
\vspace{-6pt}
\end{table}

\vspace{-16pt}
\paragraph{Different local patch sizes.} We also investigate if local image patches carry effective fingerprints for attribution. We apply the post-pooling network for image attribution. Given the setup of $\{$\textit{real}, \textit{ProGAN\_seed\_v\#i}$\}$, Table~\ref{table:postpooling} shows the classification accuracy w.r.t. patch sizes. We conclude that for CelebA face dataset a patch of size $24 \times 24$ or larger carries sufficient fingerprint information for image attribution without deterioration; for LSUN, a patch of size $52 \times 52$ or larger carries a sufficient fingerprint.

\begin{figure}
\centering
\subfigure[Non-selected samples]{\includegraphics[width=0.45\linewidth]{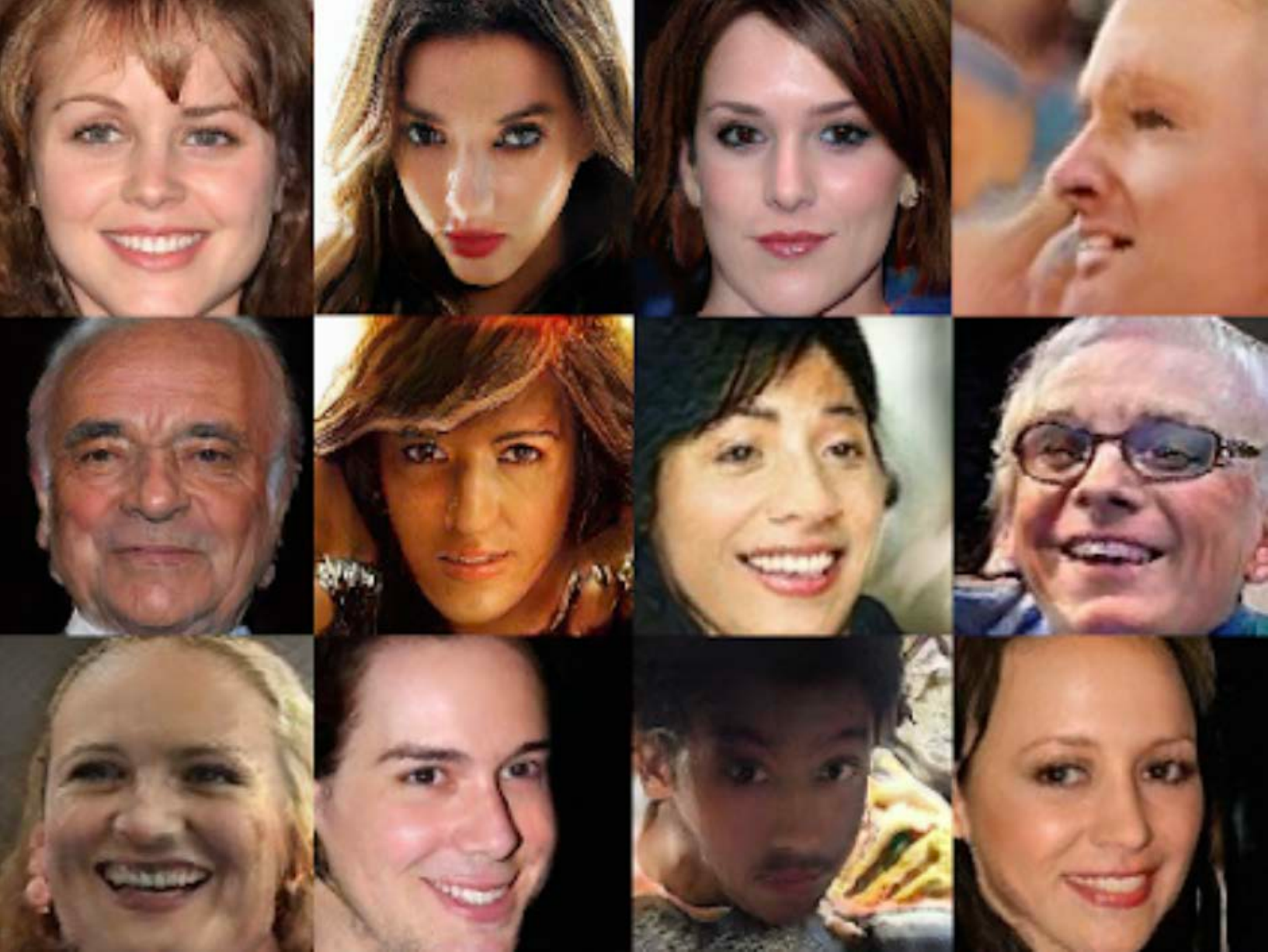}\label{fig:random_samples}}
\hspace{0.25cm}
\subfigure[Selected samples]{\includegraphics[width=0.45\linewidth]{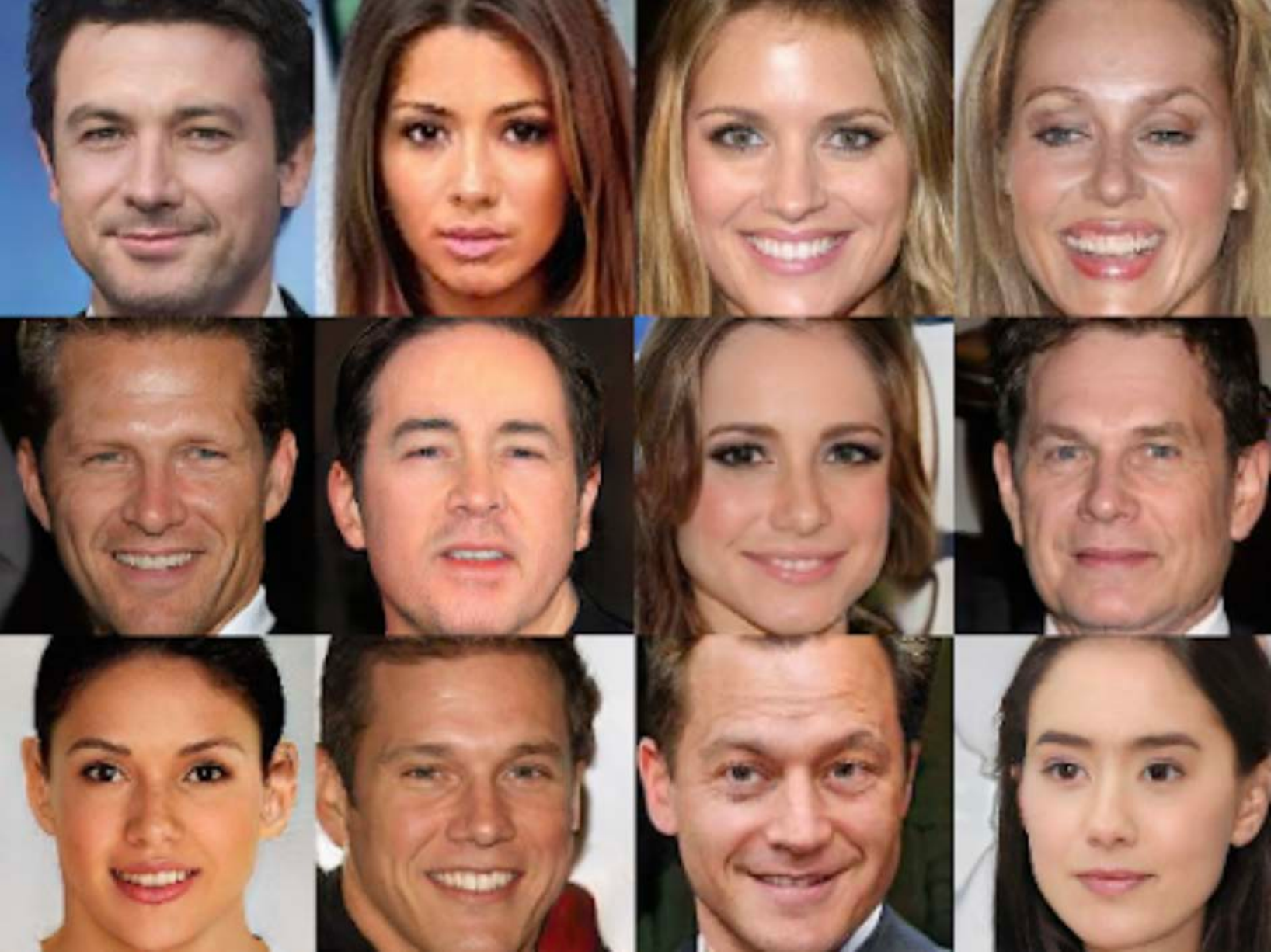}\label{fig:filtered_samples}}
\caption{Visual comparisons between (a) arbitrary face samples and (b) selected samples with top 10\% Perceptual Similarity~\cite{zhang2018perceptual} to CelebA real dataset. We notice the selected samples have higher quality and fewer artifacts. They are also more similar to each other, which challenge more on attribution.}
\label{fig:compare_artifact_free}
\vspace{-12pt}
\end{figure}

\begin{table}[!t]
\center
\small
\caption{Evaluation on the 10\% selected images of $\{$\textit{real}, \textit{ProGAN\_seed\_v\#i}$\}$. The best performance is highlighted in \textbf{bold}.}
\begin{tabular}{clcc}
\toprule
& & CelebA & LSUN \tabularnewline
\midrule
\multicolumn{1}{c}{ } & kNN & 11.99 & 10.35 \tabularnewline
\multicolumn{1}{c}{Accuracy} & Eigenface~\cite{sirovich1987low} & 26.69 & - \tabularnewline
\multicolumn{1}{c}{(\%)} & PRNU~\cite{marra2019gans} & 93.50 & 74.49 \tabularnewline
\multicolumn{1}{c}{ } & Ours & \textbf{99.93} & \textbf{98.16} \tabularnewline
\midrule
\multicolumn{1}{c}{FD ratio} & Inception~\cite{salimans2016improved} & 1.04 & 1.22 \tabularnewline
\multicolumn{1}{c}{} & Our fingerprint & \textbf{15.63} & \textbf{6.27} \tabularnewline
\bottomrule
\end{tabular}
\label{table:no_artifact}
\vspace{-6pt}
\end{table}

\vspace{-16pt}
\paragraph{Artifact-free subset.} Throughout our experiments, the state-of-the-art GAN approaches are capable of generating high-quality images -- but are also generating obvious artifacts in some cases. There is a concern that attribution might be biased by such artifacts. In order to eliminate this concern, we use Perceptual Similariy~\cite{zhang2018perceptual} to measure the 1-nearest-neighbor similarity between each testing generated image and the real-world dataset, and then select the 10\% with the highest similarity for attribution. We compare face samples between non-selected and selected sets in Figure~\ref{fig:compare_artifact_free} and compare bedroom samples in the supplementary material. We notice this metric is visually effective in selecting samples of higher quality and with fewer artifacts.

Given the setup of 10\% selected $\{$\textit{real}, \textit{ProGAN\_seed\_v\#i}$\}$, we show the performance evaluation in Table~\ref{table:no_artifact}. All the FD ratio measures consistently decreased compared to Table~\ref{table:seed}. This indicates our selection also moves the image distributions from different GAN instances closer to the real dataset and consequently closer to each other. This makes the attribution task more challenging. Encouragingly, our classifier, pre-trained on non-selected images, can perform equally well on the selected high-quality images and is hence not biased by artifacts.

\begin{figure*}[t]
\centering
\subfigure[No attack]{\includegraphics[height=2.2cm]{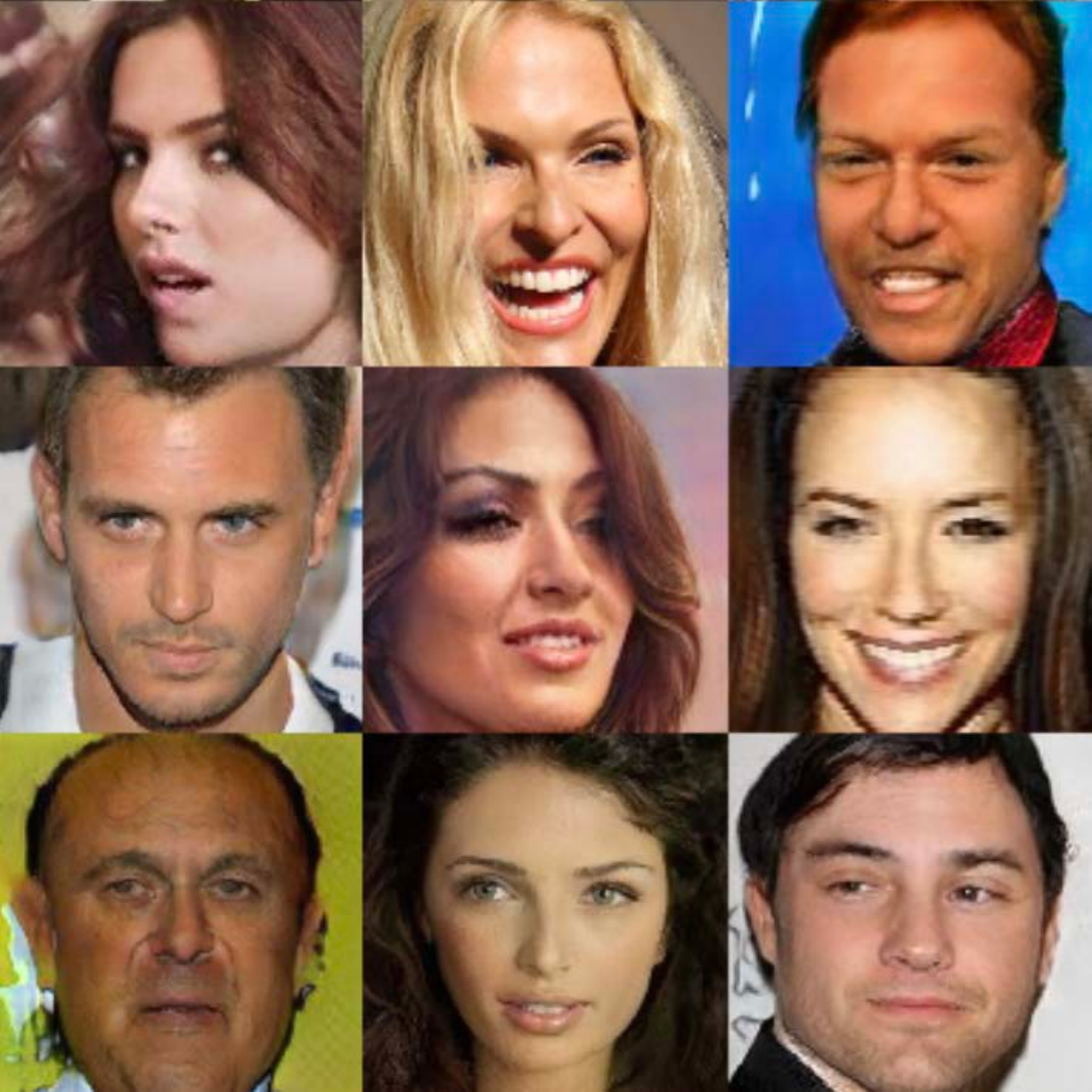}}
\hspace{0.1cm}
\subfigure[Noise]{\includegraphics[height=2.2cm]{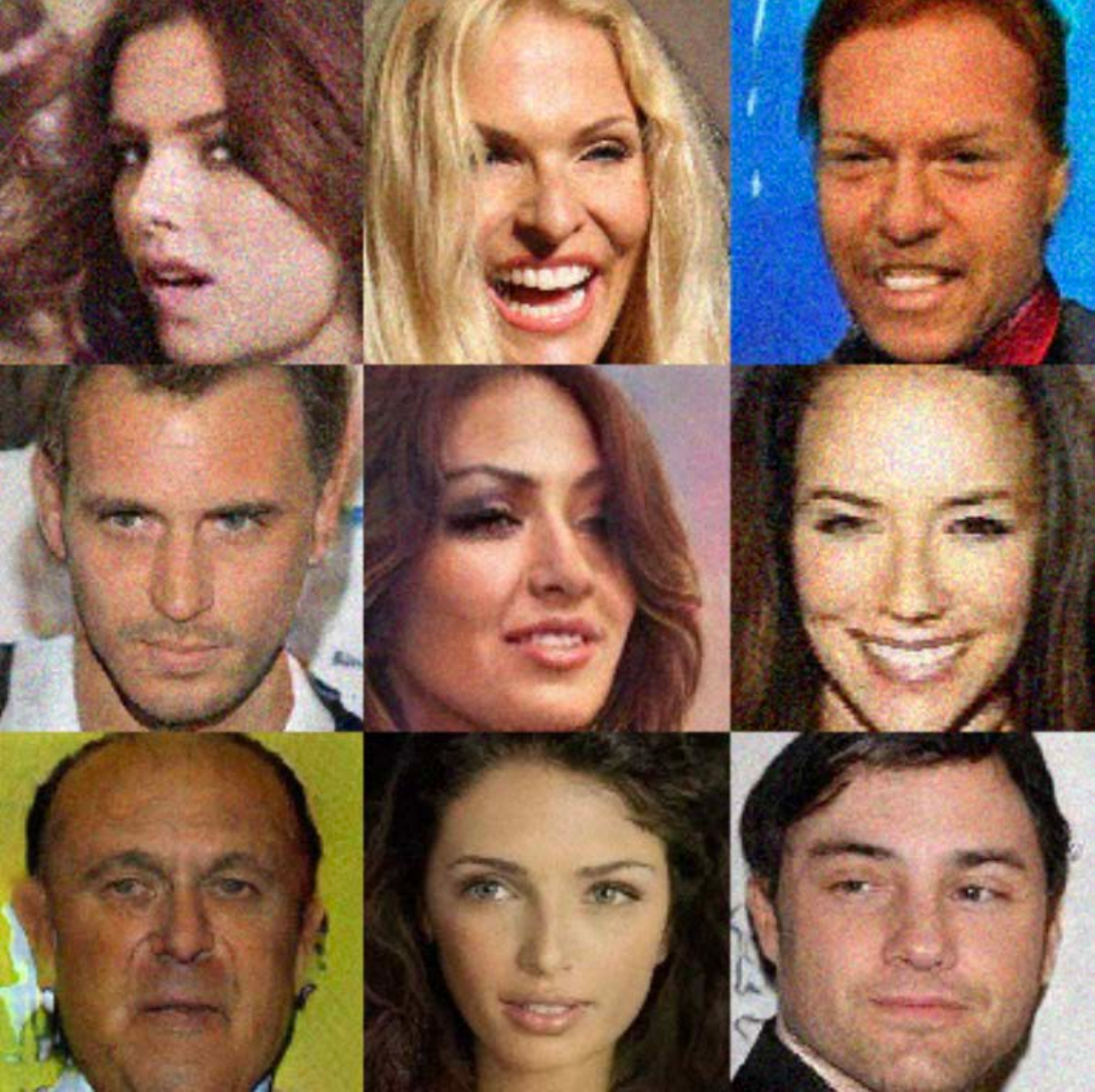}}
\hspace{0.1cm}
\subfigure[Blur]{\includegraphics[height=2.2cm]{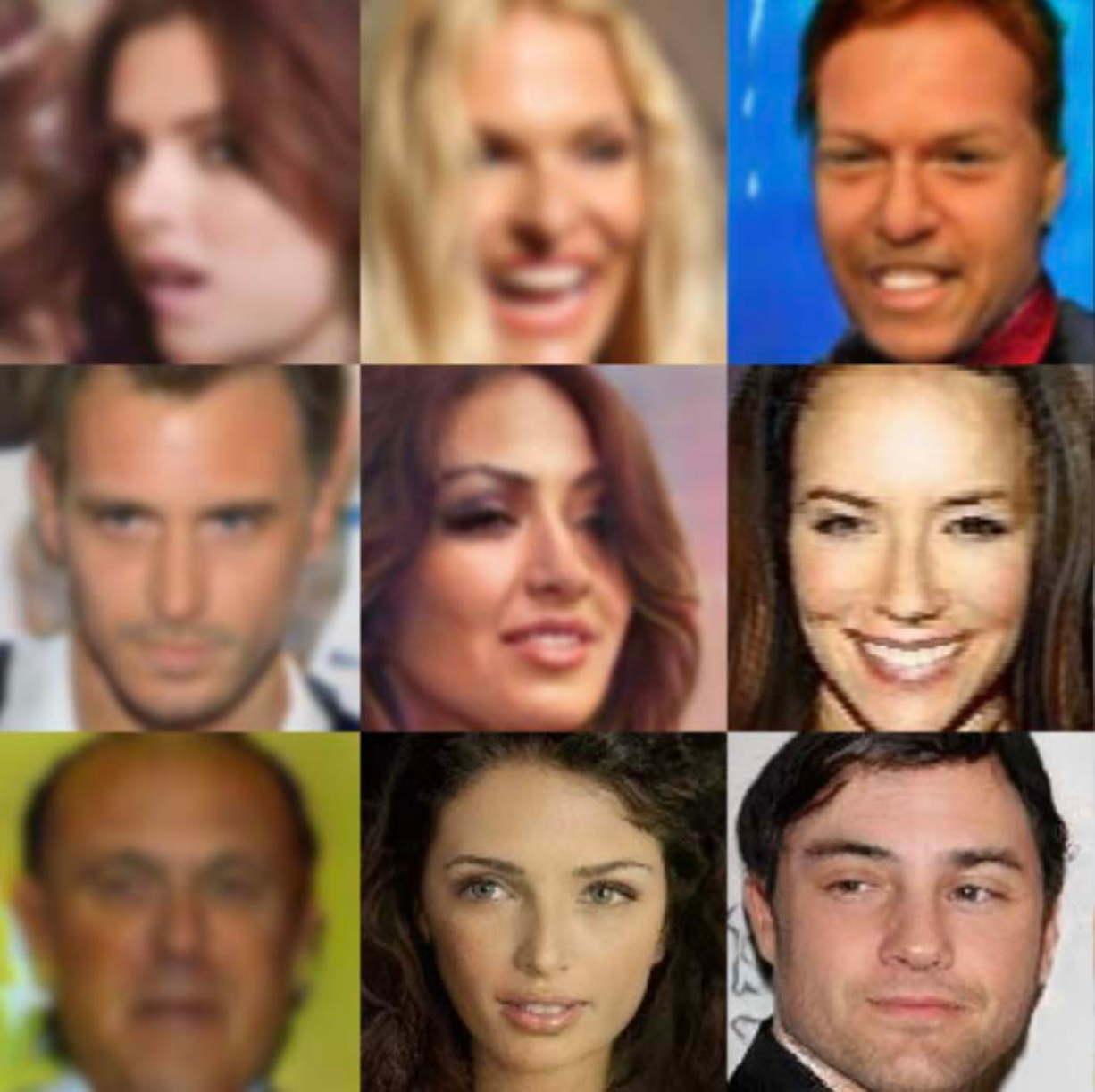}}
\hspace{0.1cm}
\subfigure[Cropping]{\includegraphics[height=2.2cm]{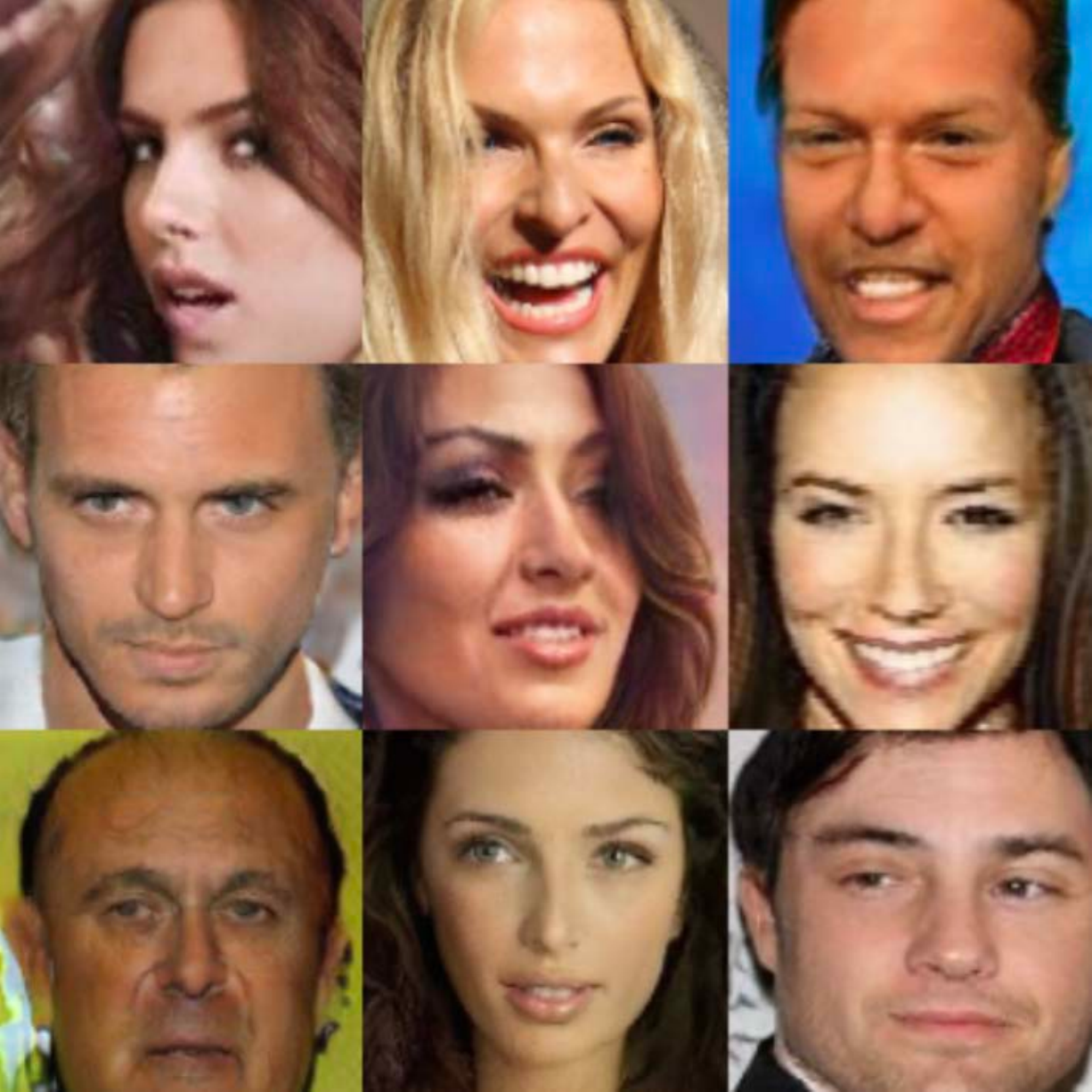}}
\hspace{0.1cm}
\subfigure[Compression]{\includegraphics[height=2.2cm]{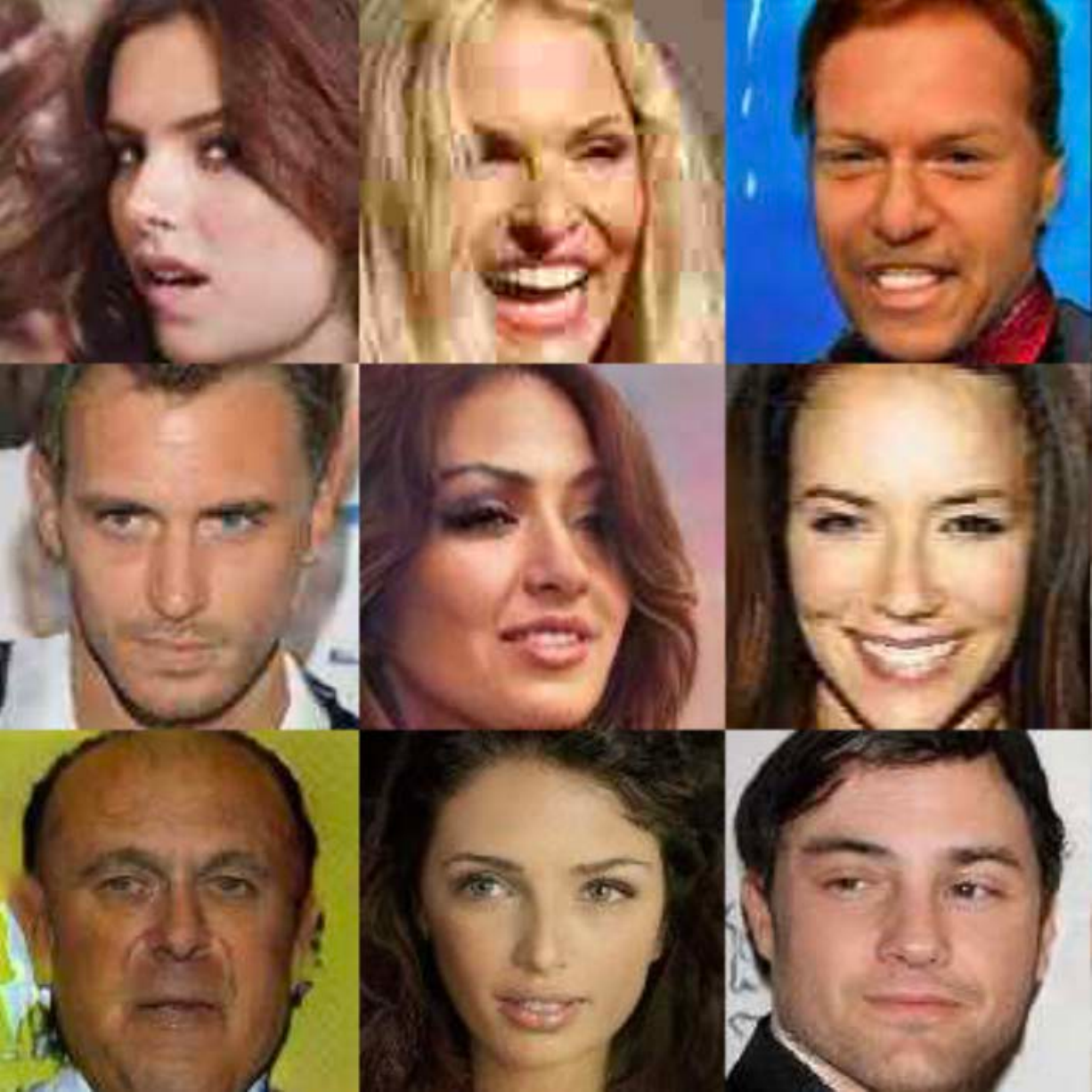}}
\hspace{0.1cm}
\subfigure[Relighting]{\includegraphics[height=2.2cm]{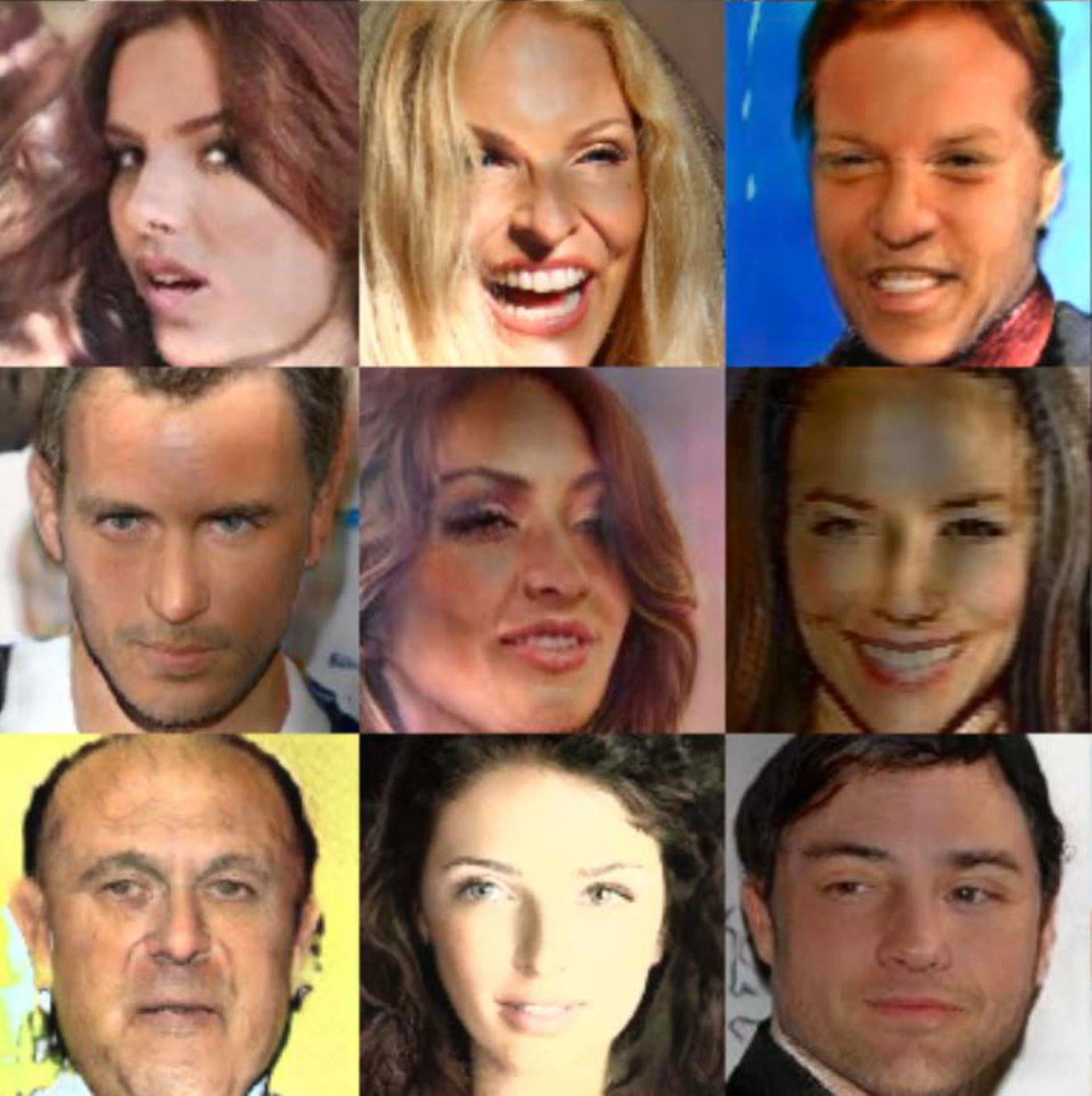}}
\hspace{0.1cm}
\subfigure[Combination]{\includegraphics[height=2.2cm]{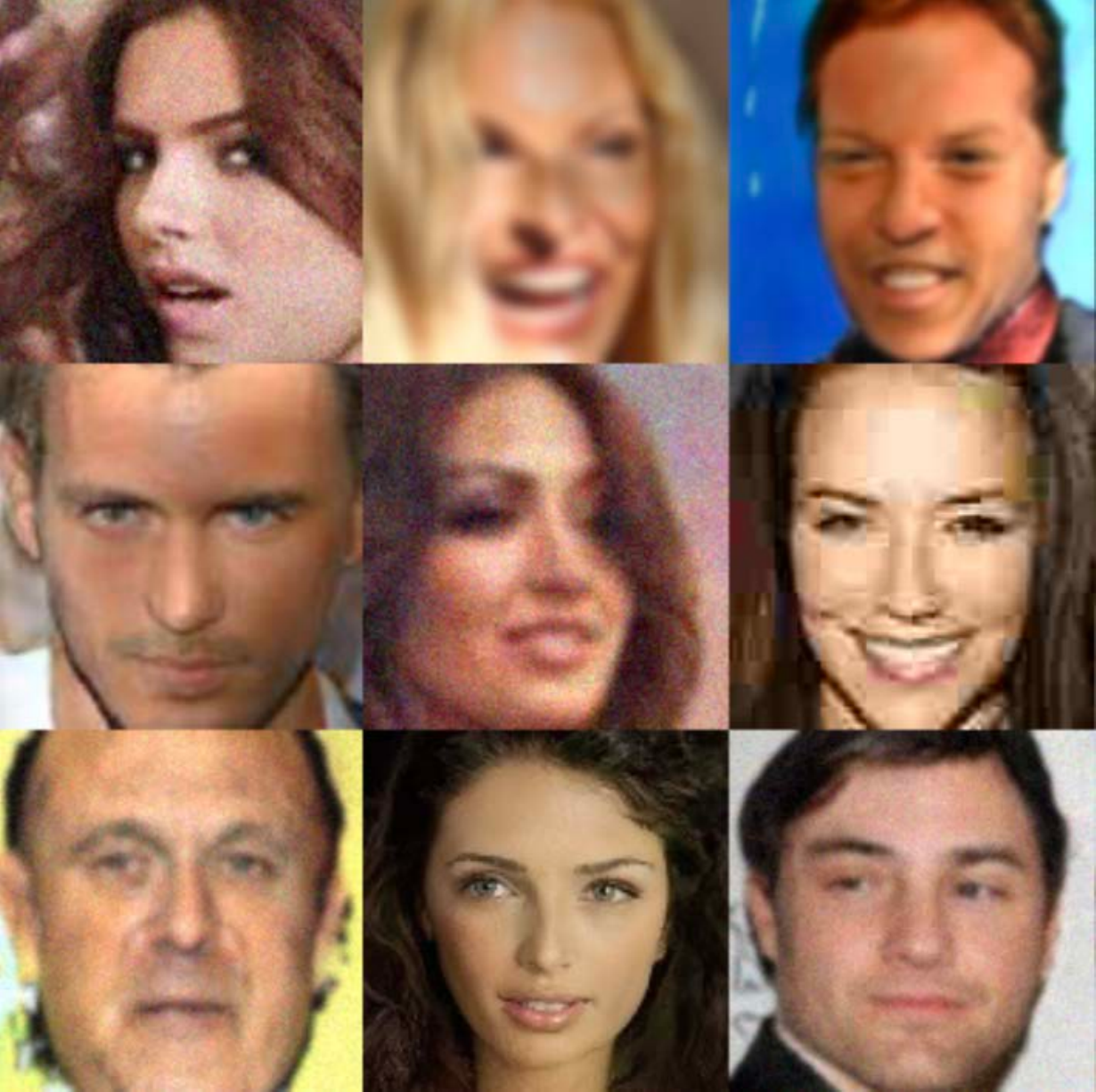}}
\caption{Image samples for the attacks and defenses of our attribution network.}
\label{fig:perturbations}
\vspace{-12pt}
\end{figure*}

\begin{table*}[!t]
\center
\small
\caption{Classification accuracy (\%) of our network w.r.t. different perturbation attacks before or after immunization on CelebA $\{$\textit{real}, \textit{ProGAN\_seed\_v\#i}$\}$. The best performance is highlighted in \textbf{bold}.}
\begin{tabular}{lcccccccccccc}
\toprule
& \multicolumn{12}{c}{CelebA} \tabularnewline
& \multicolumn{2}{c}{\textit{Noise}} & \multicolumn{2}{c}{\textit{Blur}} & \multicolumn{2}{c}{\textit{Cropping}} & \multicolumn{2}{c}{\textit{Compression}} & \multicolumn{2}{c}{\textit{Relighting}} & \multicolumn{2}{c}{\textit{Combination}} \tabularnewline
& Atk & Dfs & Atk & Dfs & Atk & Dfs & Atk & Dfs & Atk & Dfs & Atk & Dfs \tabularnewline
\midrule
PRNU~\cite{marra2019gans} & \textbf{57.88} & 63.82 & 27.37 & 42.43 & 9.84 & 10.68 & \textbf{26.15} & 44.55 & 86.59 & 87.02 & \textbf{19.93} & 21.77 \tabularnewline
Ours & 9.14 & \textbf{93.02} & \textbf{49.64} & \textbf{97.20} & \textbf{46.80} & \textbf{98.28} & 8.77 & \textbf{88.02} & \textbf{94.02} & \textbf{98.66} & 19.31 & \textbf{72.64} \tabularnewline
\bottomrule
\end{tabular}
\label{table:attack_defense_celeba}
\vspace{-6pt}
\end{table*}

\begin{table*}[!t]
\center
\small
\caption{Classification accuracy (\%) of our network w.r.t. different perturbation attacks before or after immunization on LSUN bedroom $\{$\textit{real}, \textit{ProGAN\_seed\_v\#i}$\}$. The best performance is highlighted in \textbf{bold}.}
\begin{tabular}{lcccccccccccc}
\toprule
& \multicolumn{12}{c}{LSUN} \tabularnewline
& \multicolumn{2}{c}{\textit{Noise}} & \multicolumn{2}{c}{\textit{Blur}} & \multicolumn{2}{c}{\textit{Cropping}} & \multicolumn{2}{c}{\textit{Compression}} & \multicolumn{2}{c}{\textit{Relighting}} & \multicolumn{2}{c}{\textit{Combination}} \tabularnewline
& Atk & Dfs & Atk & Dfs & Atk & Dfs & Atk & Dfs & Atk & Dfs & Atk & Dfs \tabularnewline
\midrule
PRNU~\cite{marra2019gans} & \textbf{39.59} & 40.97 & 26.92 &30.79 & 9.30 & 9.42 & 18.27 & 23.66 & 60.86 & 63.31 &  16.54 & 16.89 \tabularnewline
Ours & 11.80 & \textbf{95.30} & \textbf{74.48} & \textbf{96.68} & \textbf{86.20} & \textbf{97.30} & \textbf{24.73} & \textbf{92.40} & \textbf{62.21}  & \textbf{97.36} & \textbf{24.44} & \textbf{83.42} \tabularnewline
\bottomrule
\end{tabular}
\label{table:attack_defense_lsun}
\vspace{-6pt}
\end{table*}

\begin{figure}[!t]
\centering
\includegraphics[width=1\linewidth]{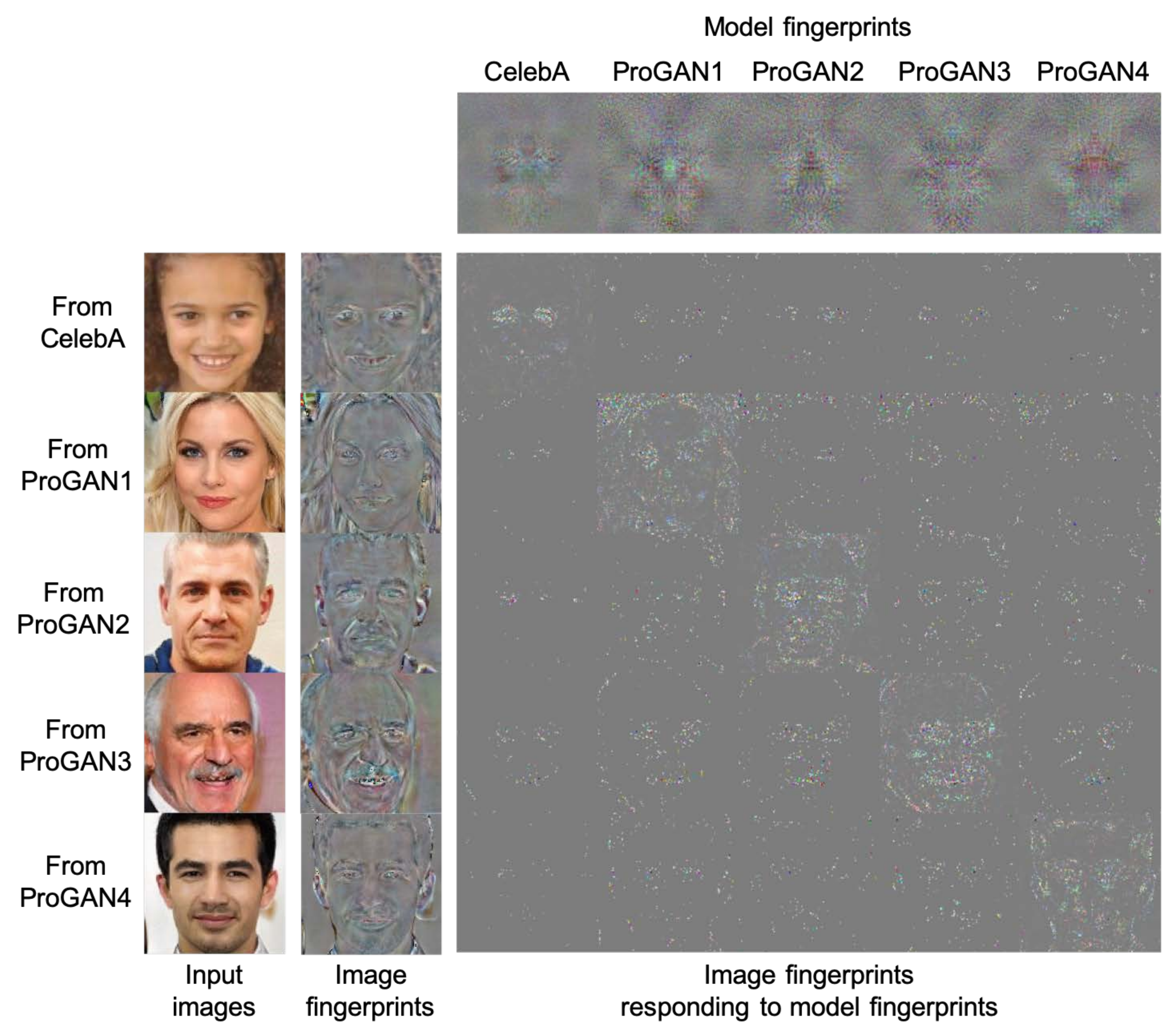}
\caption{Visualization of model and image fingerprint samples. \ning{Their pairwise interactions are shown as the confusion matrix.}}
\label{fig:fingerprints_confusion_matrix}
\vspace{-12pt}
\end{figure}

\subsection{\ning{Immunizability:} how robust is attribution to image perturbation attacks and how effective are the defenses?} \label{sec:attack_defense}
\paragraph{Attacks.} We apply five types of attacks that perturb testing images~\cite{joon18icrl}: \textit{noise}, \textit{blur}, \textit{cropping}, \textit{JPEG compression}, \textit{relighting}, and random combination of them. The intention is to confuse the attribution network by destroying image fingerprints. Examples of the perturbations on face images are shown in Figure~\ref{fig:perturbations}. Examples on bedroom images are shown in the supplementary material.

\textit{Noise} adds i.i.d. Gaussian noise to testing images. The Gaussian variance is randomly sampled from $U[5.0, 20.0]$. \textit{Blur} performs Gaussian filtering on testing images with kernel size randomly picked from $\{1,\; 3,\; 5,\; 7,\; 9\}$. \textit{Cropping} crops testing images with a random offset between $5\%$ and $20\%$ of the image side lengths and then resizes back to the original. \textit{JPEG compression} performs JPEG compression processing with quality factor randomly sampled from $U[10, 75]$. \textit{Relighting} uses SfSNet~\cite{sengupta2018sfsnet} to replace the current image lighting condition with another random one from their lighting dataset. The combination performs each attack with a $50\%$ probability in the order of \textit{relighting}, \textit{cropping}, \textit{blur}, \textit{JPEG compression}, and \textit{noise}.

Given perturbed images and the setup of $\{$\textit{real}, \textit{ProGAN\_seed\_v\#i}$\}$, we show the pre-trained classifier performances in the ``Akt" columns in Table~\ref{table:attack_defense_celeba} and Table~\ref{table:attack_defense_lsun}. All performances decrease due to attacks. In detail, the classifier completely fails to overcome \textit{noise} and \textit{JPEG compression} attacks. It still performs better than random when facing the other four types of attacks. The \textit{relighting} attack is the least effective one because it only perturbs low-frequency image components. The barely unchanged fingerprints in high-frequency components enables reasonable attribution.

\vspace{-16pt}
\paragraph{Defenses.} In order to immunize our classifier against attacks, we finetune the classifier under the assumption that we know the attack category. Given perturbed images and the setup of $\{$\textit{real}, \textit{ProGAN\_seed\_v\#i}$\}$, we show the finetuned classifier performance in the ``Dfs" columns in Table~\ref{table:attack_defense_celeba} and Table~\ref{table:attack_defense_lsun}. It turns out that the immunized classifier completely regains performance over \textit{blur}, \textit{cropping} and \textit{relighting} attacks, and partially regains performance over the others. However, the recovery from \textit{combination} attack is minimal due to its highest complexity. In addition, our method consistently outperforms the method of Marra \etal~\cite{marra2019gans} under each attack after immunization, while theirs does not effectively benefit from such immunization.

\subsection{Fingerprint visualization} \label{sec:fingerprint_exp}
Given the setup of $\{$\textit{real}, \textit{ProGAN\_seed\_v\#i}$\}$, we alternatively apply the fingerprint visualization network (Section~\ref{sec:fingerprint}) to attribute images. We show the attribution performance in the ``Our visNet" row in Table~\ref{table:seed}, which are comparable to that of the attribution model. \ning{Figure~\ref{fig:fingerprints_confusion_matrix} visualizes face fingerprints. Bedroom fingerprints are shown in the supplementary material. It turns out that image fingerprints maximize responses only to their own model fingerprints, which supports effective attribution. To attribute the real-world image, it is sufficient for the fingerprint to focus only on the eyes. To attribute the other images, the fingerprints also consider clues from the background, which, compared to foreground faces, is more variant and harder for GANs to approximate realistically~\cite{medium}.}




\section{Conclusion}
We have presented the first study of learning GAN fingerprints towards image attribution. Our experiments show that even a small difference in GAN training (e.g., the difference in initialization) can leave a distinct fingerprint that commonly exists over all its generated images. That enables fine-grained image attribution and model attribution. Further encouragingly, fingerprints are persistent across different frequencies and different patch sizes, and are not biased by GAN artifacts. Even though fingerprints can be deteriorated by several image perturbation attacks, they are effectively immunizable by simple finetuning. Comparisons also show that, in a variety of conditions, our learned fingerprints are consistently superior to the very recent baseline~\cite{marra2019gans} for attribution, and consistently outperform inception features~\cite{salimans2016improved} for cross-source distinguishability.

\section*{Acknowledgement}
\ning{This project was partially funded by DARPA MediFor program under cooperative agreement FA87501620191.} We acknowledge the Maryland Advanced Research Computing Center for providing computing resources. We thank Hao Zhou for helping with the relighting experiments. We also thank Yaser Yacoob and Abhinav Shrivastava for constructive advice in general.

{\small
\bibliographystyle{ieee_fullname}
\bibliography{main}
}

\clearpage
\renewcommand\thesubsection{\Alph{subsection}}

\section{Supplementary material}

\subsection{Fr\'{e}chet Distance ratio}
As described in Section 4.1 in the main paper, we use the ratio of inter-class and intra-class Fr\'{e}chet Distance ~\cite{dowson1982frechet}, denoted as FD ratio, to evaluate the distinguishability of a feature representation across classes. For inter-class FD calculation, we first measure the FD between two feature distributions from a pair of different classes, and then average over each possible pair. For intra-class FD calculation, we first measure the FD between two feature distributions from two disjoint sets of images in the same class, where we split the class equally, and then average over each class.

Mathematically,
\begin{equation}
\text{FD ratio} = \frac{\text{inter-class FD}}{\text{intra-class FD}}
\end{equation}

inter-class FD $=$
\begin{equation}
\frac{1}{||\{(y, \tilde{y})|y\neq\tilde{y}\}||}\sum_{y\neq\tilde{y}}\text{FD}\Big(\big\{f(I_i)|y_i=y\big\}, \big\{f(I_j)|y_j=\tilde{y}\big\}\Big)
\end{equation}

intra-class FD $=$
\begin{equation}
\frac{1}{||\mathbb{Y}||}\sum_{y\in\mathbb{Y}, \{i\}\cap\{j\}=\emptyset}\text{FD}\Big(\big\{f(I_i)|y_i=y\big\}, \big\{f(I_j)|y_j=y\big\}\Big)
\end{equation}
where $\mathbb{Y}$ is the class set for image sources and $f(\cdot)$ is a feature representation mapping from image domain to a feature domain.

Then in all the tables in the main paper, we compare FD ratio between the inception feature~\cite{salimans2016improved} as a baseline and our learned features. The larger the ratio, the more distinguishable the feature representation across sources. We also show in Figure 1 in the main paper the t-sne visualization~\cite{maaten2008visualizing} of the two features.

\subsection{Face samples}
\begin{figure*}[!t]
\centering
\includegraphics[width=1\linewidth]{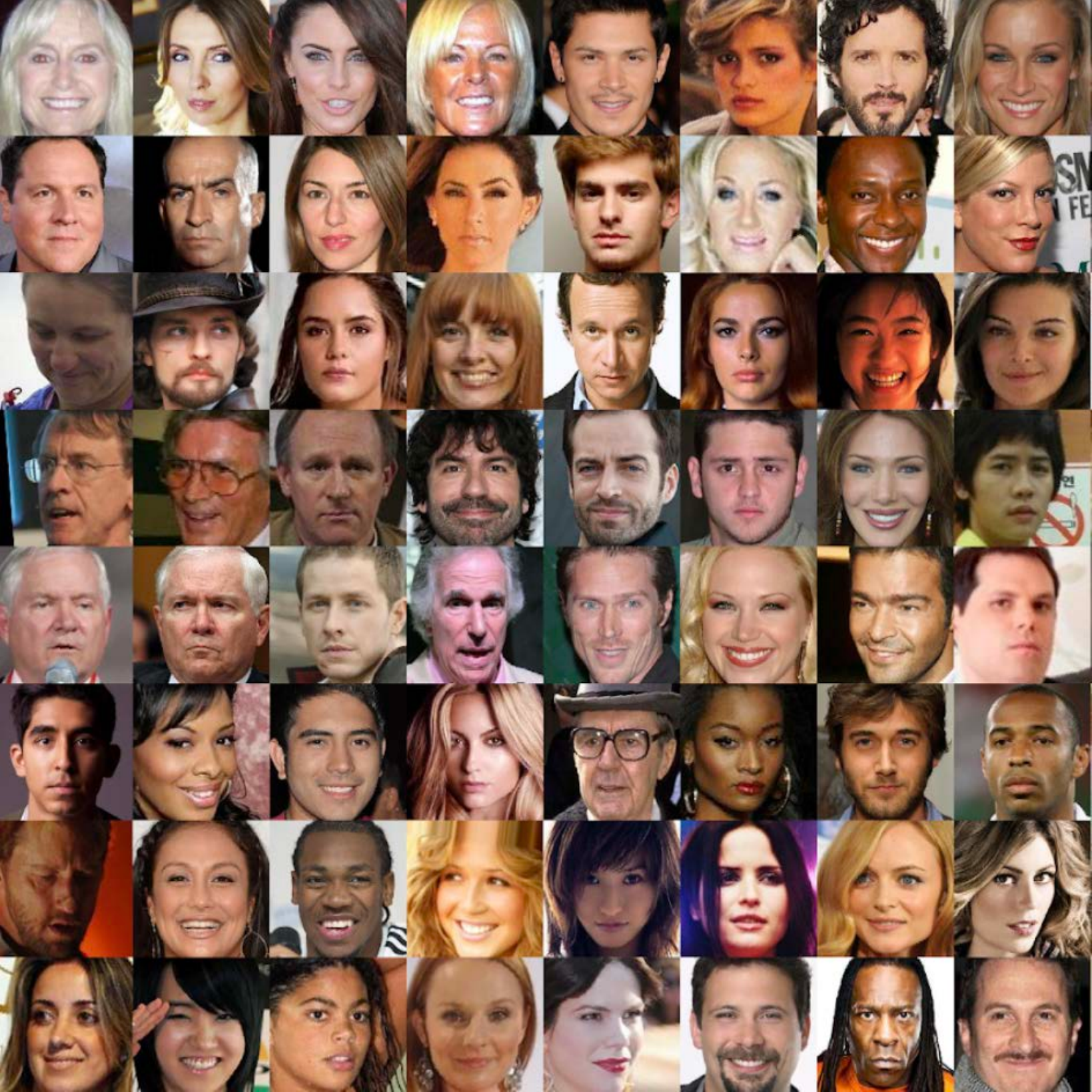}
\caption{Face samples from CelebA real dataset~\cite{liu2015faceattributes}}
\label{fig:celeba_large}
\end{figure*}

\begin{figure*}[!t]
\centering
\includegraphics[width=1\linewidth]{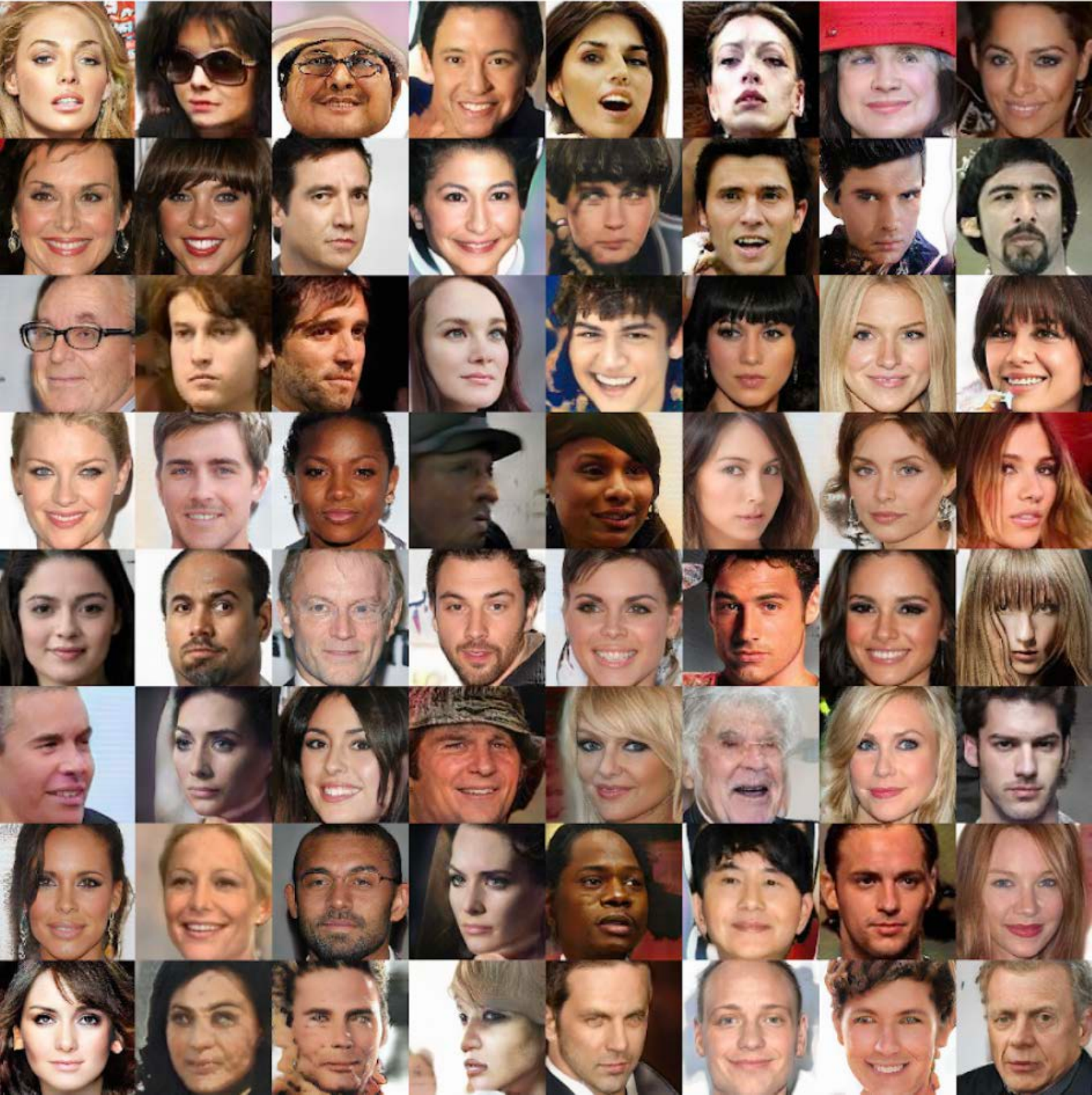}
\caption{Face samples from ProGAN~\cite{karras2018progressive}}
\label{fig:progan_large}
\end{figure*}

\begin{figure*}[!t]
\centering
\includegraphics[width=1\linewidth]{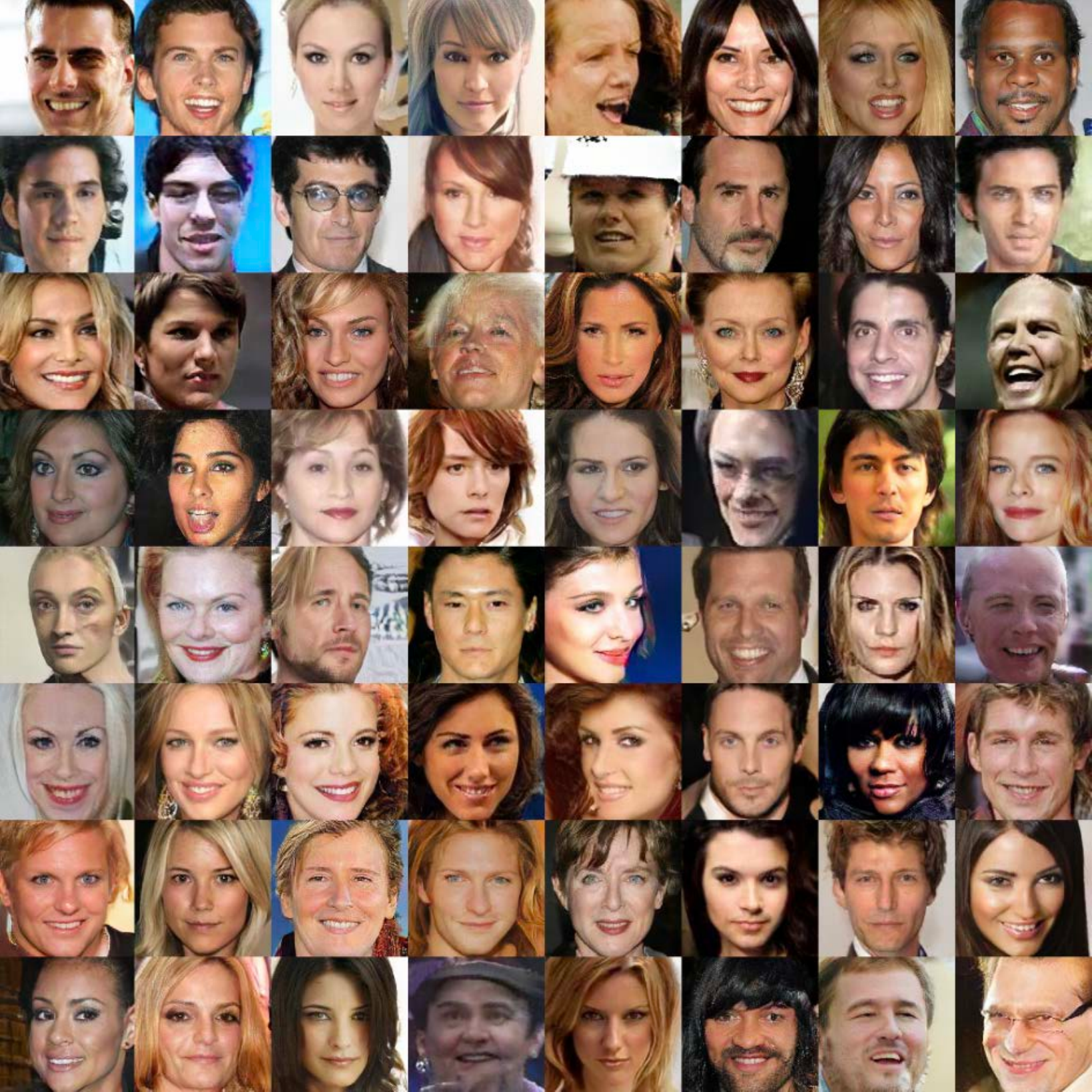}
\caption{Face samples from SNGAN~\cite{miyato2018spectral}}
\label{fig:sngan_large}
\end{figure*}

\begin{figure*}[!t]
\centering
\includegraphics[width=1\linewidth]{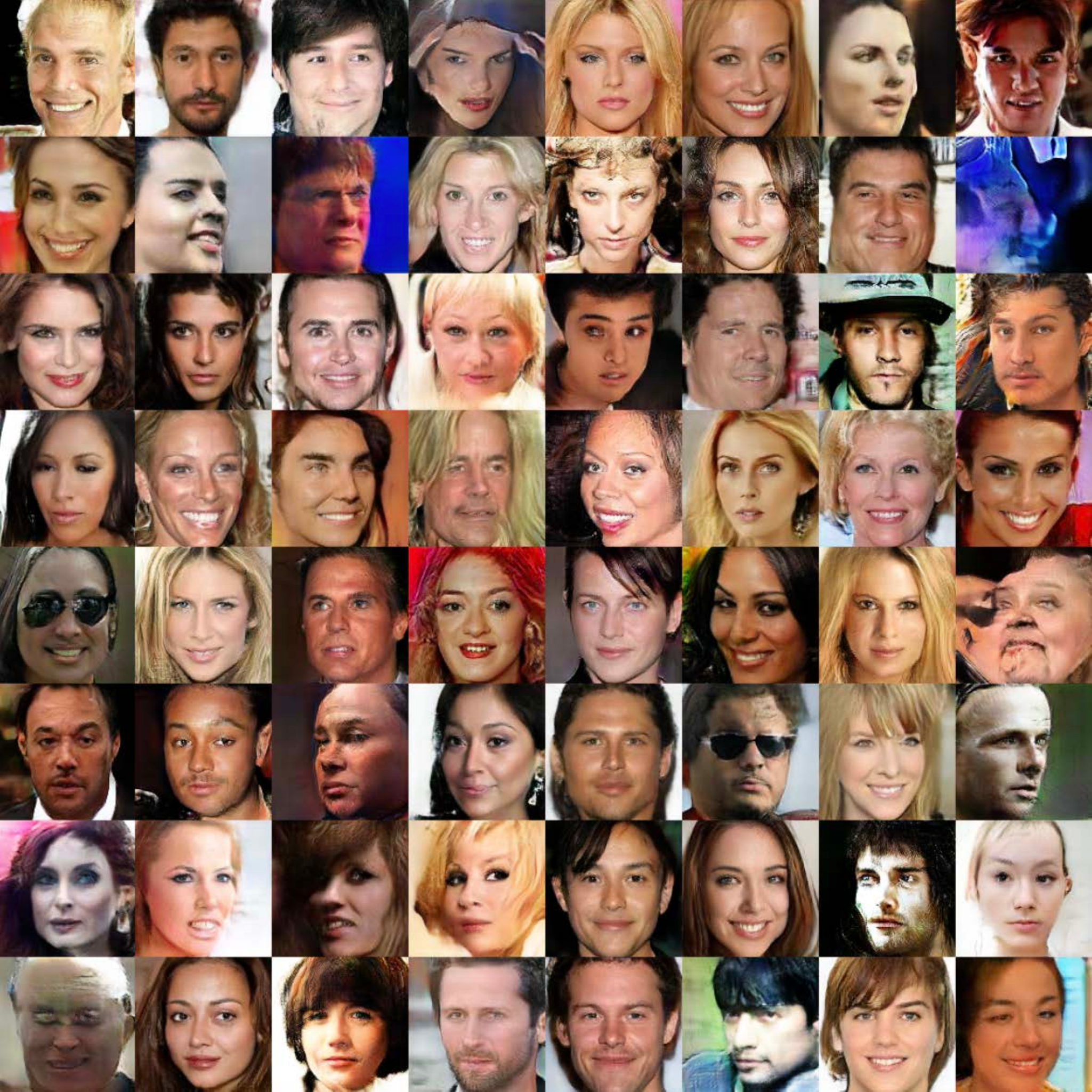}
\caption{Face samples from CramerGAN~\cite{bellemare2017cramer}}
\label{fig:cramergan_large}
\end{figure*}

\begin{figure*}[!t]
\centering
\includegraphics[width=1\linewidth]{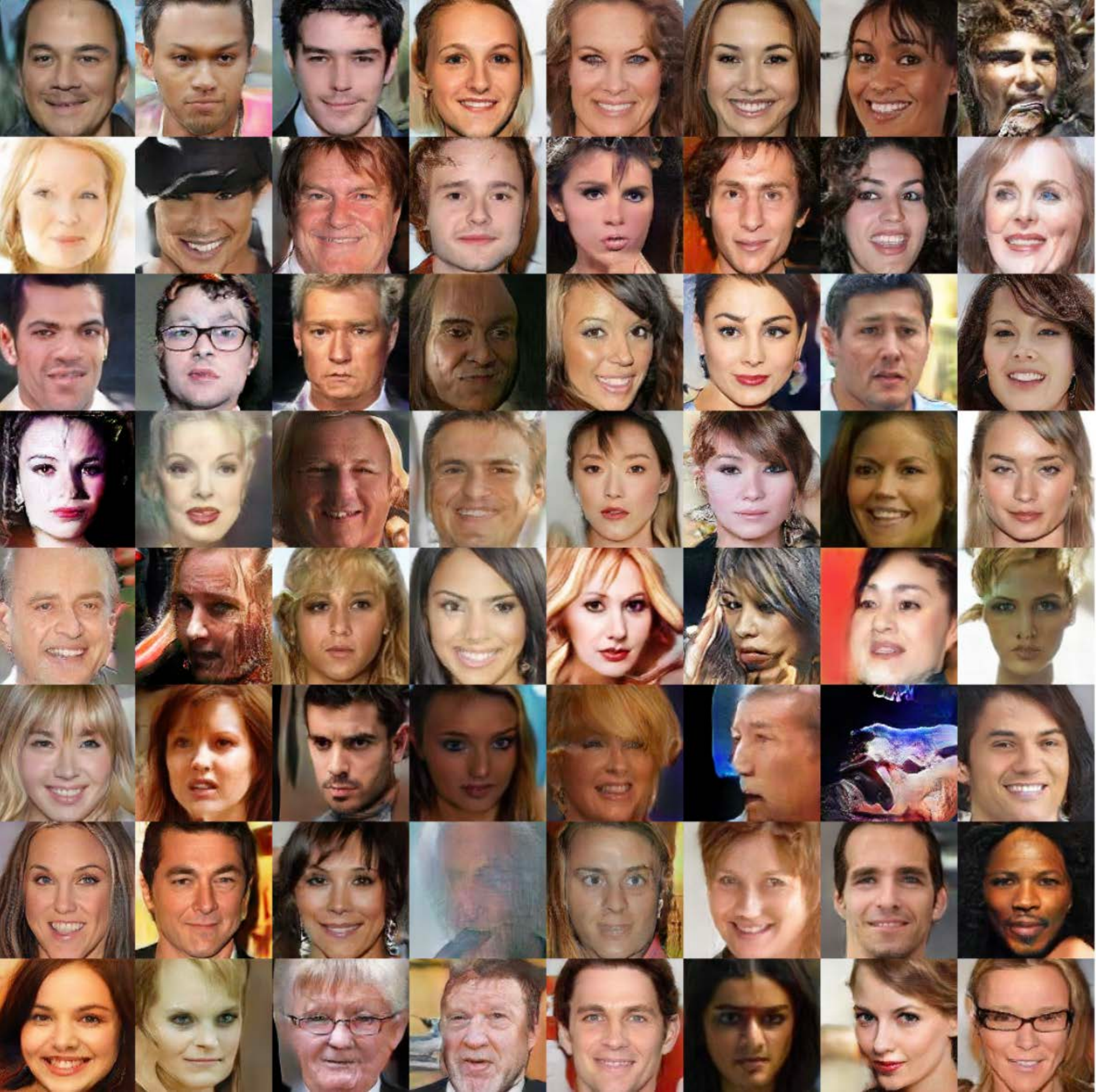}
\caption{Face samples from MMDGAN~\cite{binkowski2018demystifying}}
\label{fig:mmdgan_large}
\end{figure*}

\begin{figure*}[!t]
\centering
\includegraphics[width=1\linewidth]{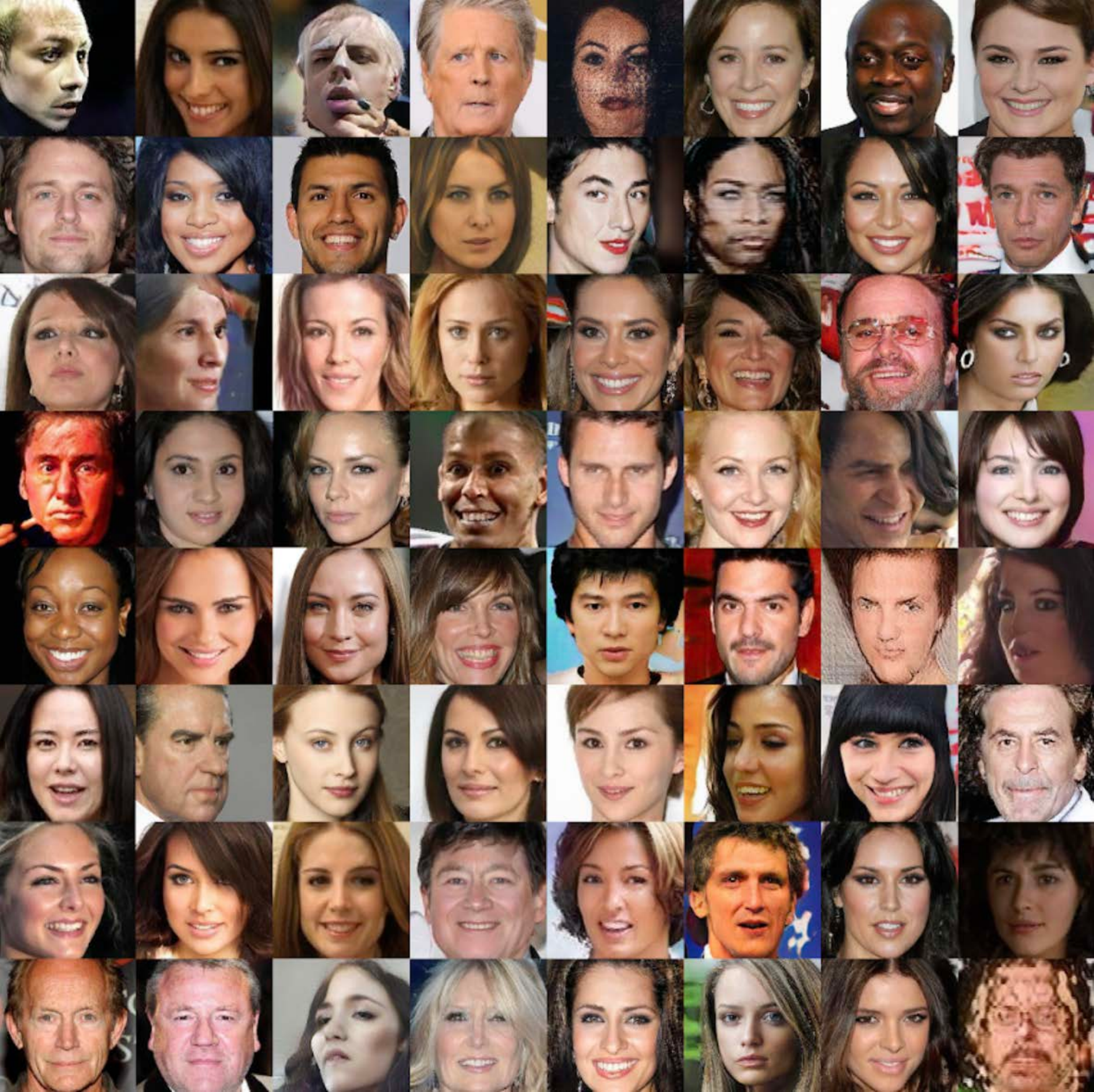}
\caption{Arbitrary face samples from the setup of $\{$\textit{real}, \textit{ProGAN\_seed\_v\#i}$\}$ where $i\in \{1$, ..., $10\}$.}
\label{fig:random_samples_large}
\end{figure*}

\begin{figure*}[!t]
\centering
\includegraphics[width=1\linewidth]{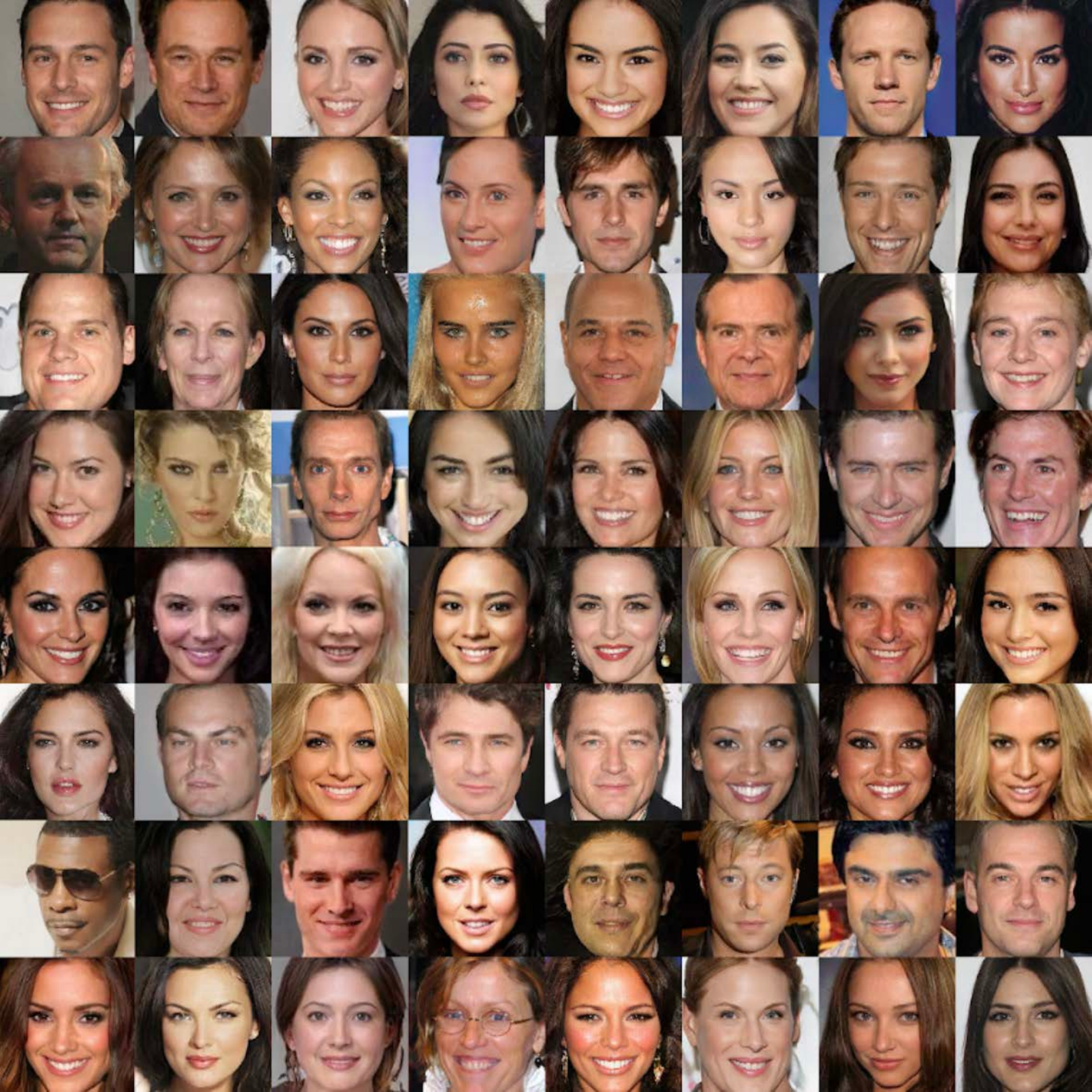}
\caption{Filtered face samples from the setup of $\{$\textit{real}, \textit{ProGAN\_seed\_v\#i}$\}$ with the top 10\% largest Perceptual Similarity~\cite{zhang2018perceptual} to real dataset distribution.}
\label{fig:filtered_samples_large}
\end{figure*}

\begin{figure*}[!t]
\centering
\includegraphics[width=1\linewidth]{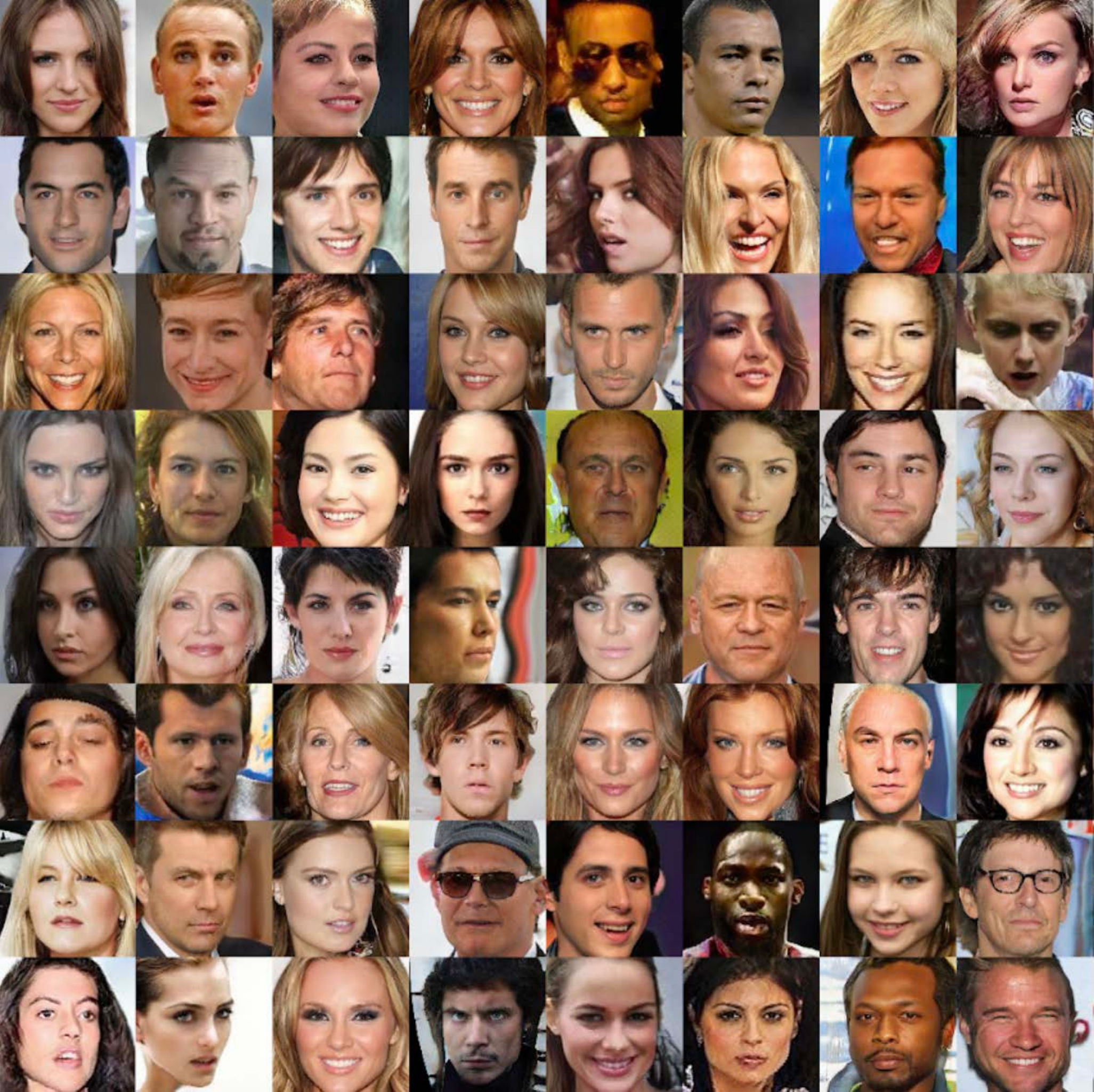}
\caption{Arbitrary face samples without attack from the setup of $\{$\textit{real}, \textit{ProGAN\_seed\_v\#i}$\}$.}
\label{fig:plain_large}
\end{figure*}

\begin{figure*}[!t]
\centering
\includegraphics[width=1\linewidth]{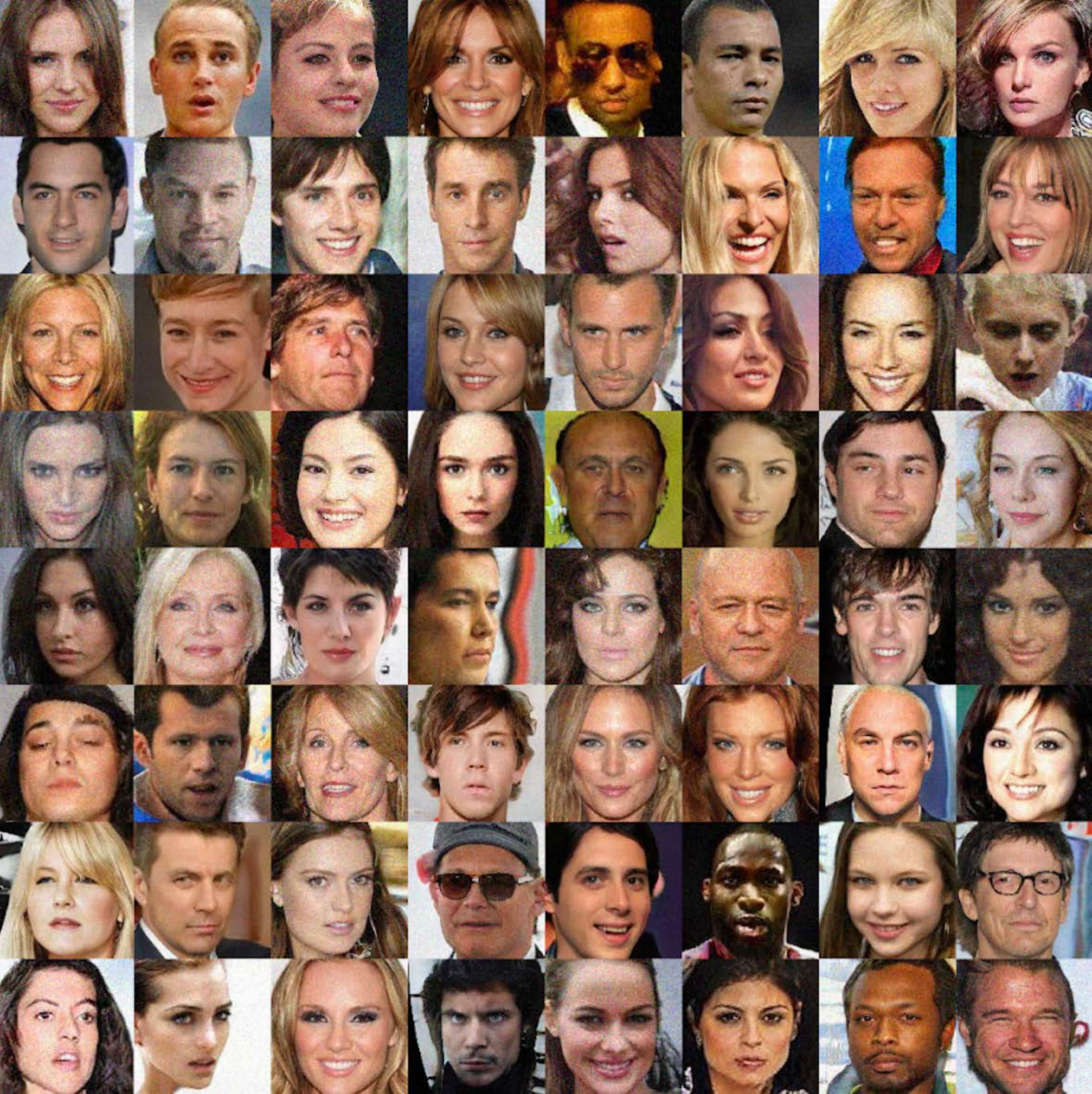}
\caption{Arbitrary face samples with \textit{noise} attack from the setup of $\{$\textit{real}, \textit{ProGAN\_seed\_v\#i}$\}$.}
\label{fig:noise_large}
\end{figure*}

\begin{figure*}[!t]
\centering
\includegraphics[width=1\linewidth]{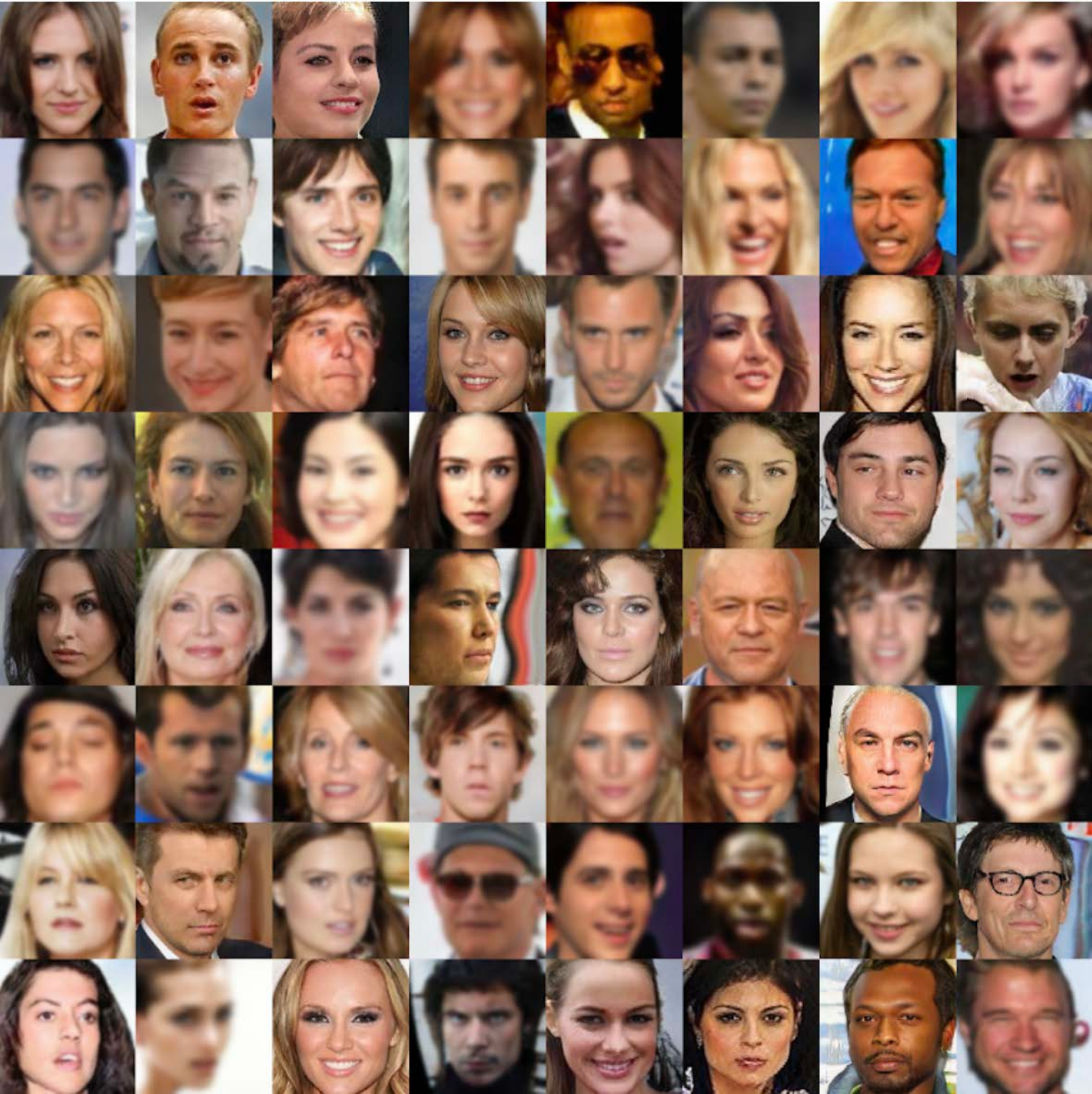}
\caption{Arbitrary face samples with \textit{blur} attack from the setup of $\{$\textit{real}, \textit{ProGAN\_seed\_v\#i}$\}$.}
\label{fig:blur_large}
\end{figure*}

\begin{figure*}[!t]
\centering
\includegraphics[width=1\linewidth]{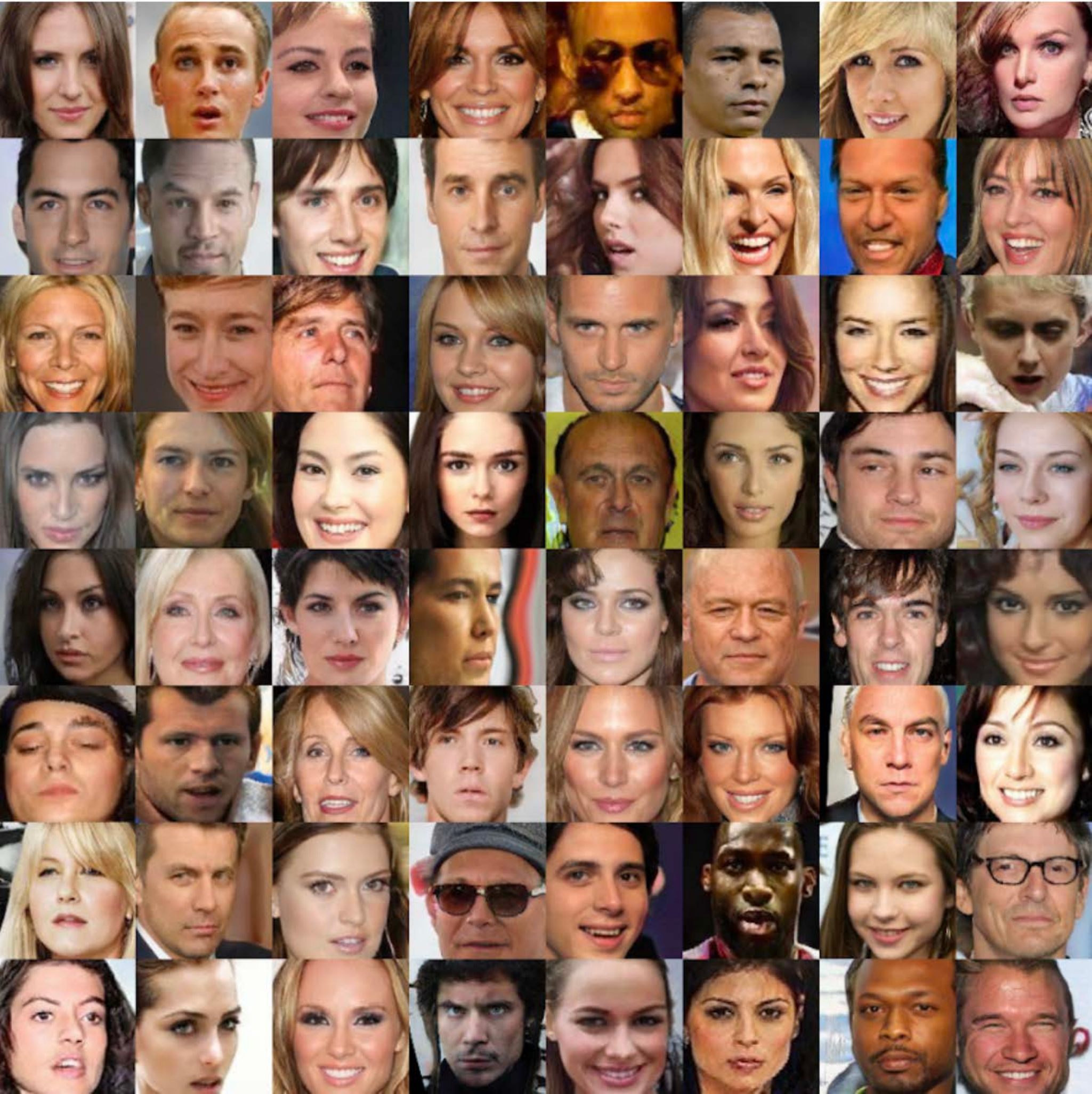}
\caption{Arbitrary face samples with \textit{cropping} attack from the setup of $\{$\textit{real}, \textit{ProGAN\_seed\_v\#i}$\}$.}
\label{fig:crop_large}
\end{figure*}

\begin{figure*}[!t]
\centering
\includegraphics[width=1\linewidth]{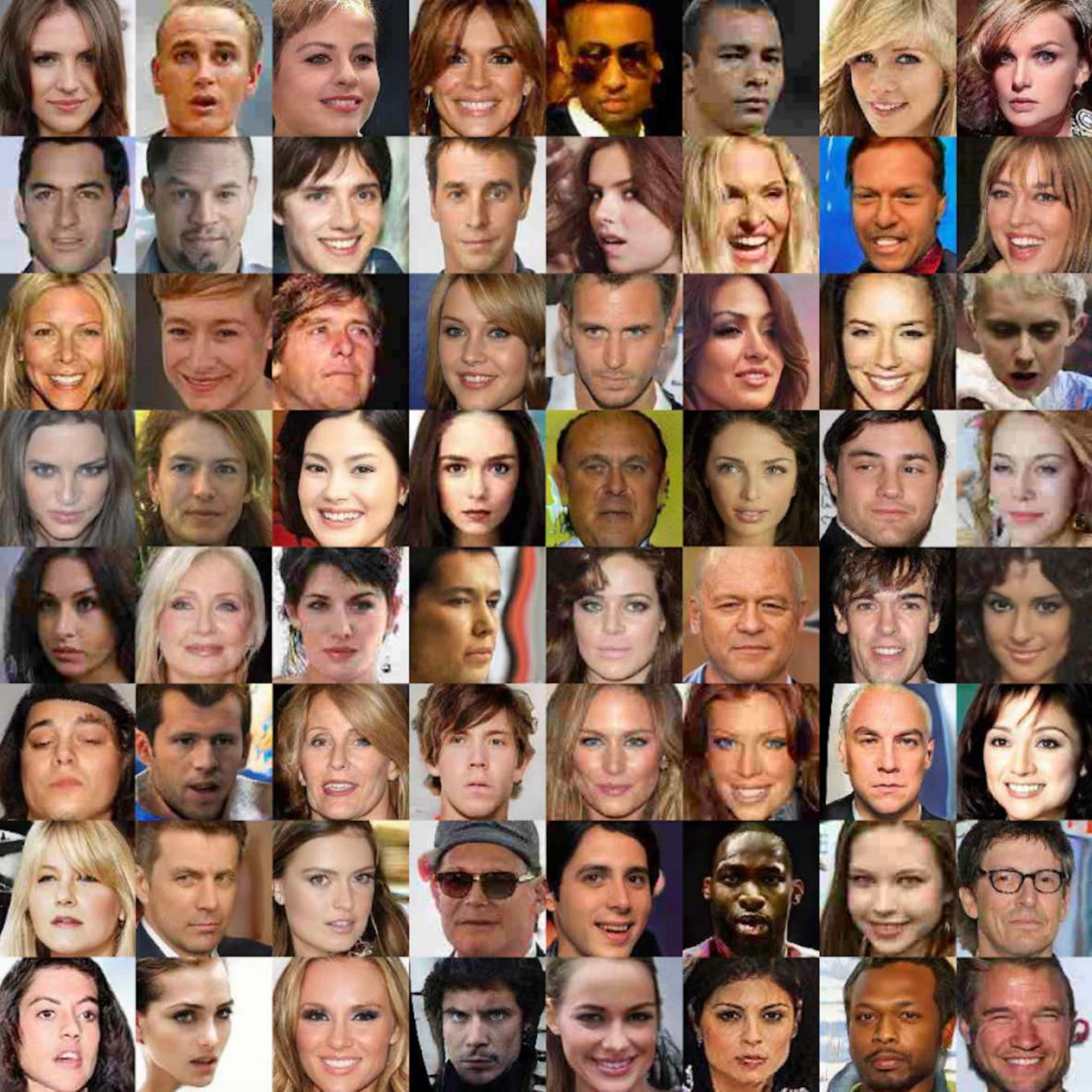}
\caption{Arbitrary face samples with \textit{JPEG compression} attack from the setup of $\{$\textit{real}, \textit{ProGAN\_seed\_v\#i}$\}$.}
\label{fig:compression_large}
\end{figure*}

\begin{figure*}[!t]
\centering
\includegraphics[width=1\linewidth]{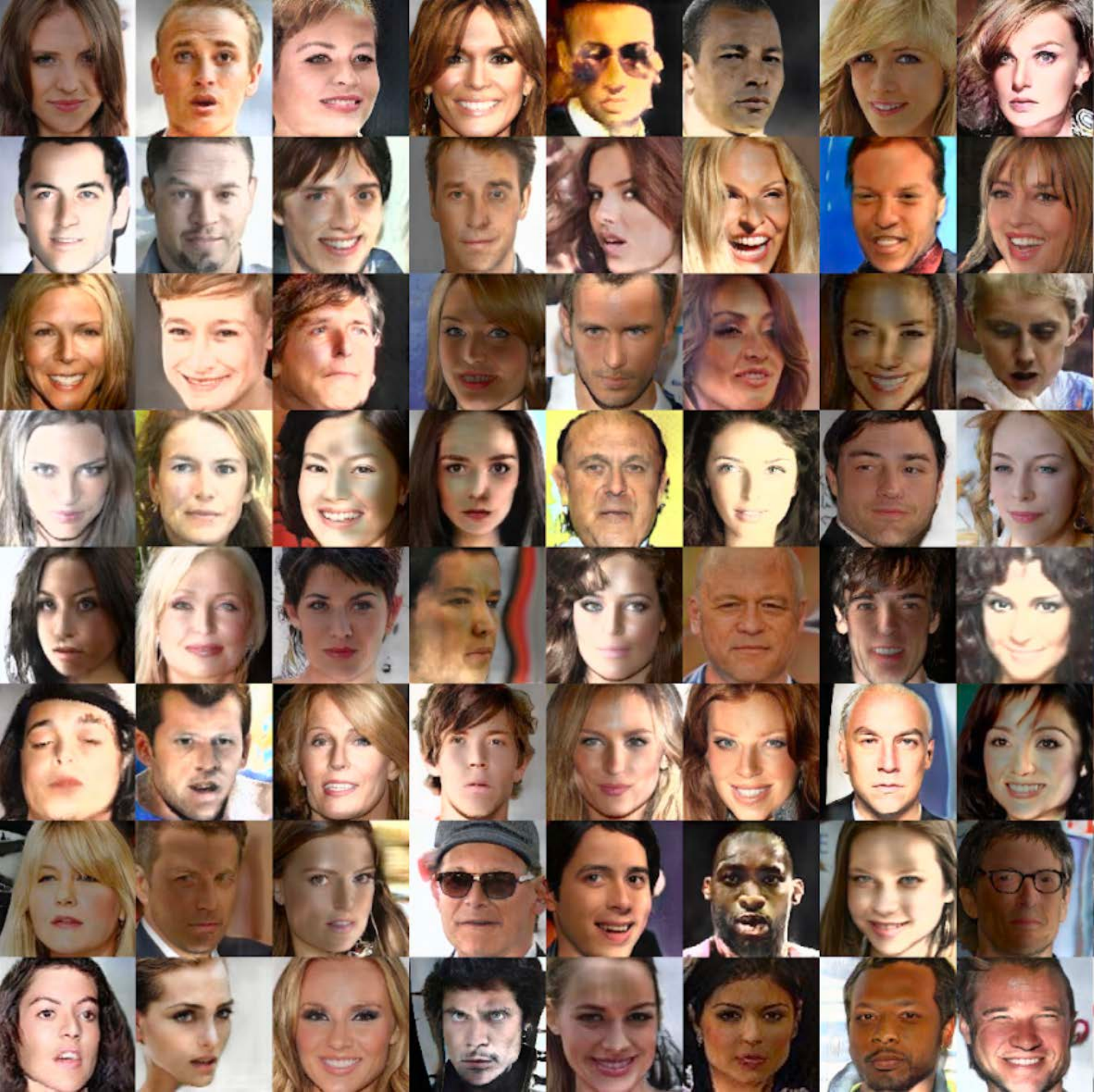}
\caption{Arbitrary face samples with \textit{relighting} attack from the setup of $\{$\textit{real}, \textit{ProGAN\_seed\_v\#i}$\}$.}
\label{fig:relighting_large}
\end{figure*}

\begin{figure*}[!t]
\centering
\includegraphics[width=1\linewidth]{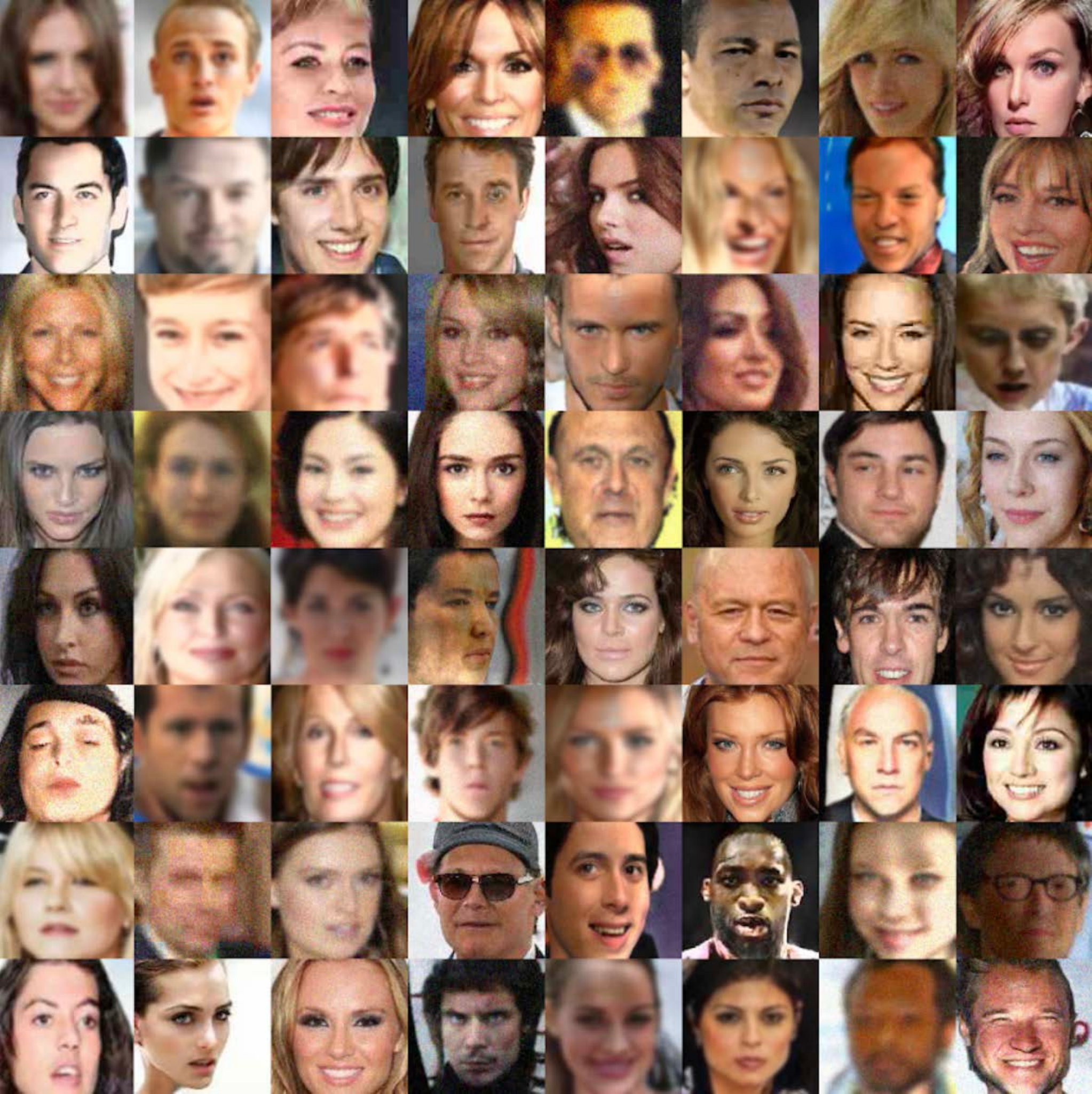}
\caption{Arbitrary face samples with the combination attack from the setup of $\{$\textit{real}, \textit{ProGAN\_seed\_v\#i}$\}$.}
\label{fig:combo_large}
\end{figure*}

We show more face samples corresponding to the experiments in the main paper. See Figure~\ref{fig:celeba_large}~to~\ref{fig:combo_large}.

\subsection{Bedroom samples}
\begin{figure*}[!t]
\centering
\includegraphics[width=1\linewidth]{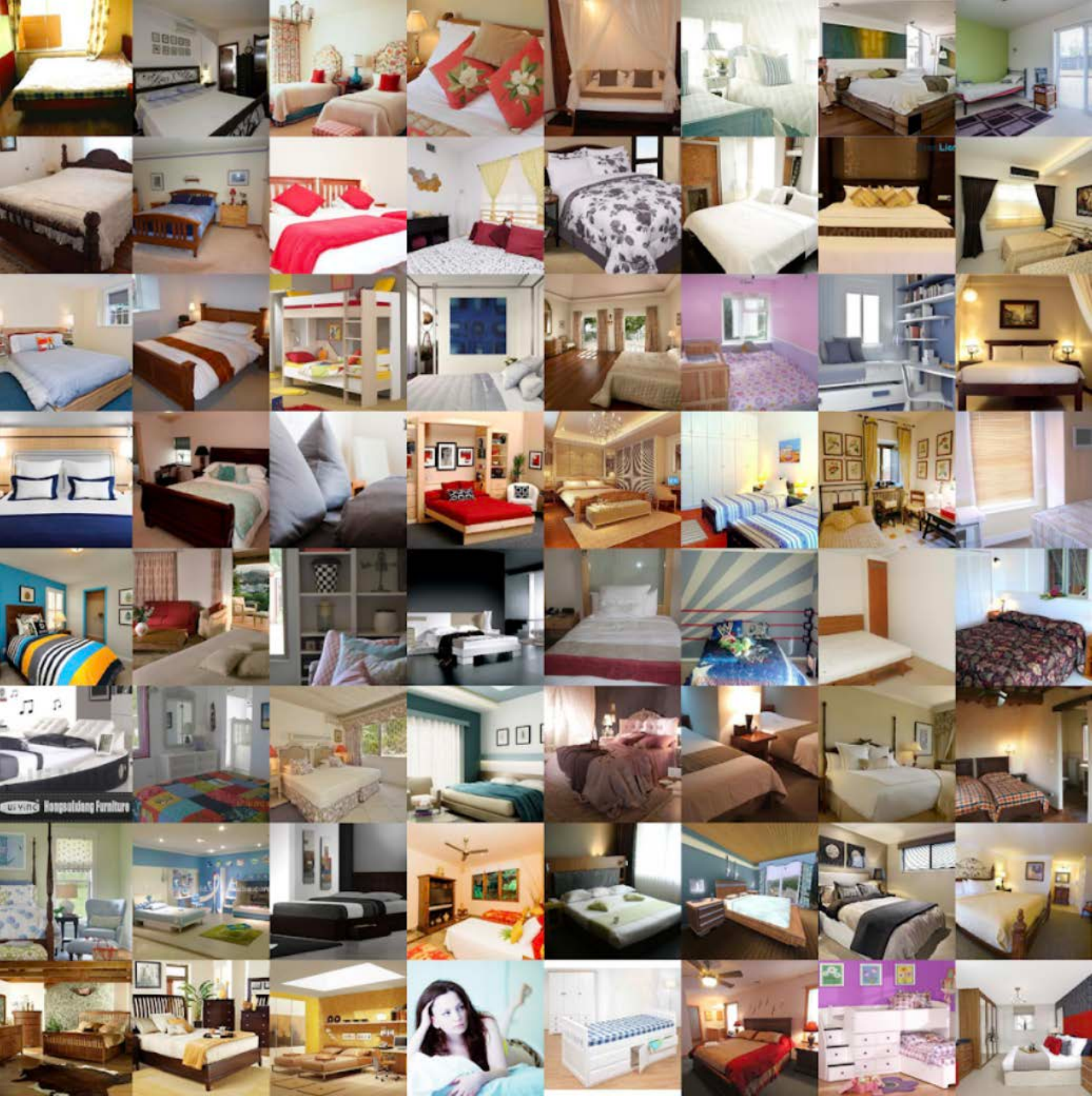}
\caption{Bedroom samples from LSUN real dataset~\cite{yu2015lsun}}
\label{fig:lsun_large}
\end{figure*}

\begin{figure*}[!t]
\centering
\includegraphics[width=1\linewidth]{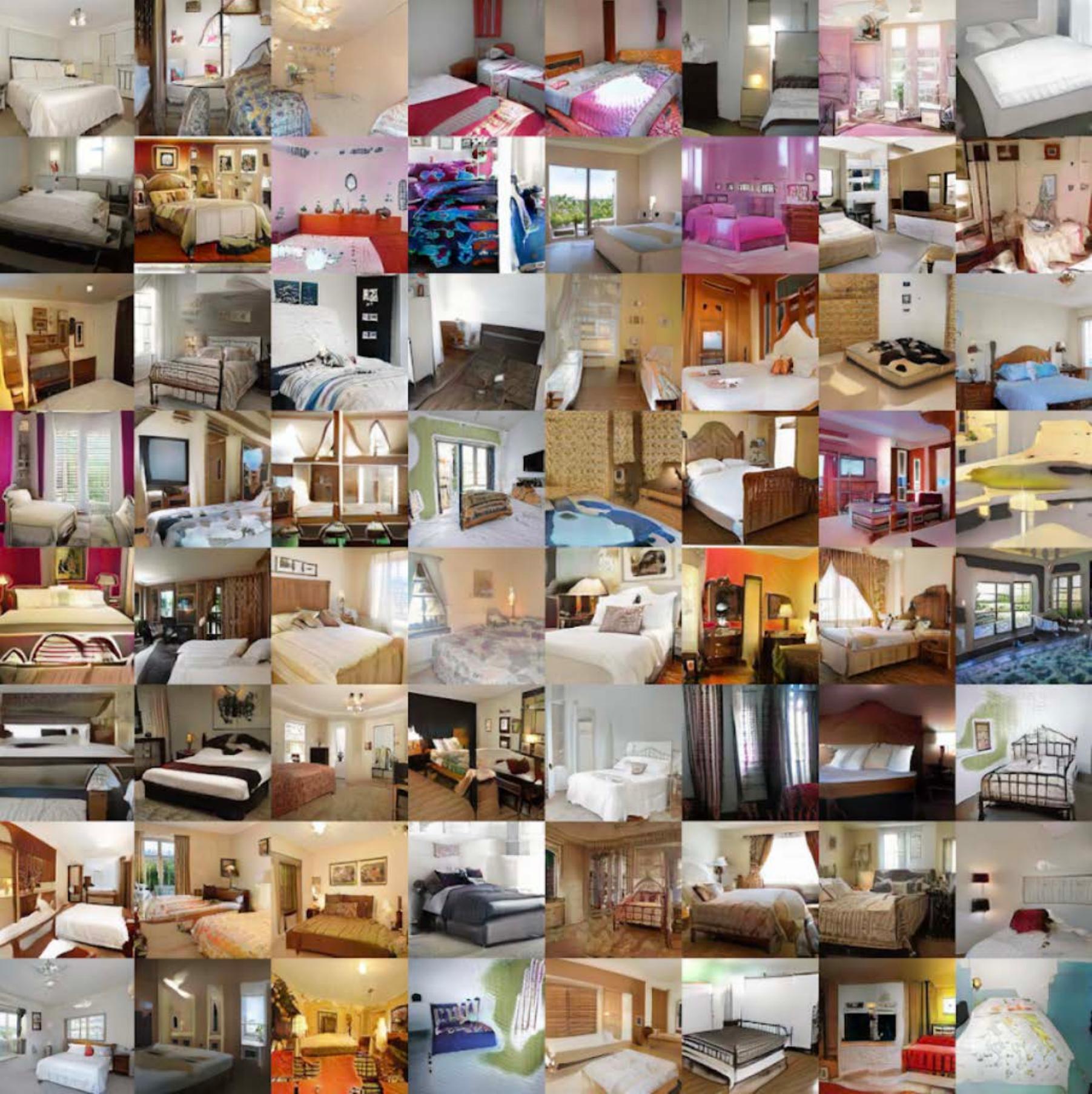}
\caption{Bedroom samples from ProGAN~\cite{karras2018progressive}}
\label{fig:progan_lsun_large}
\end{figure*}

\begin{figure*}[!t]
\centering
\includegraphics[width=1\linewidth]{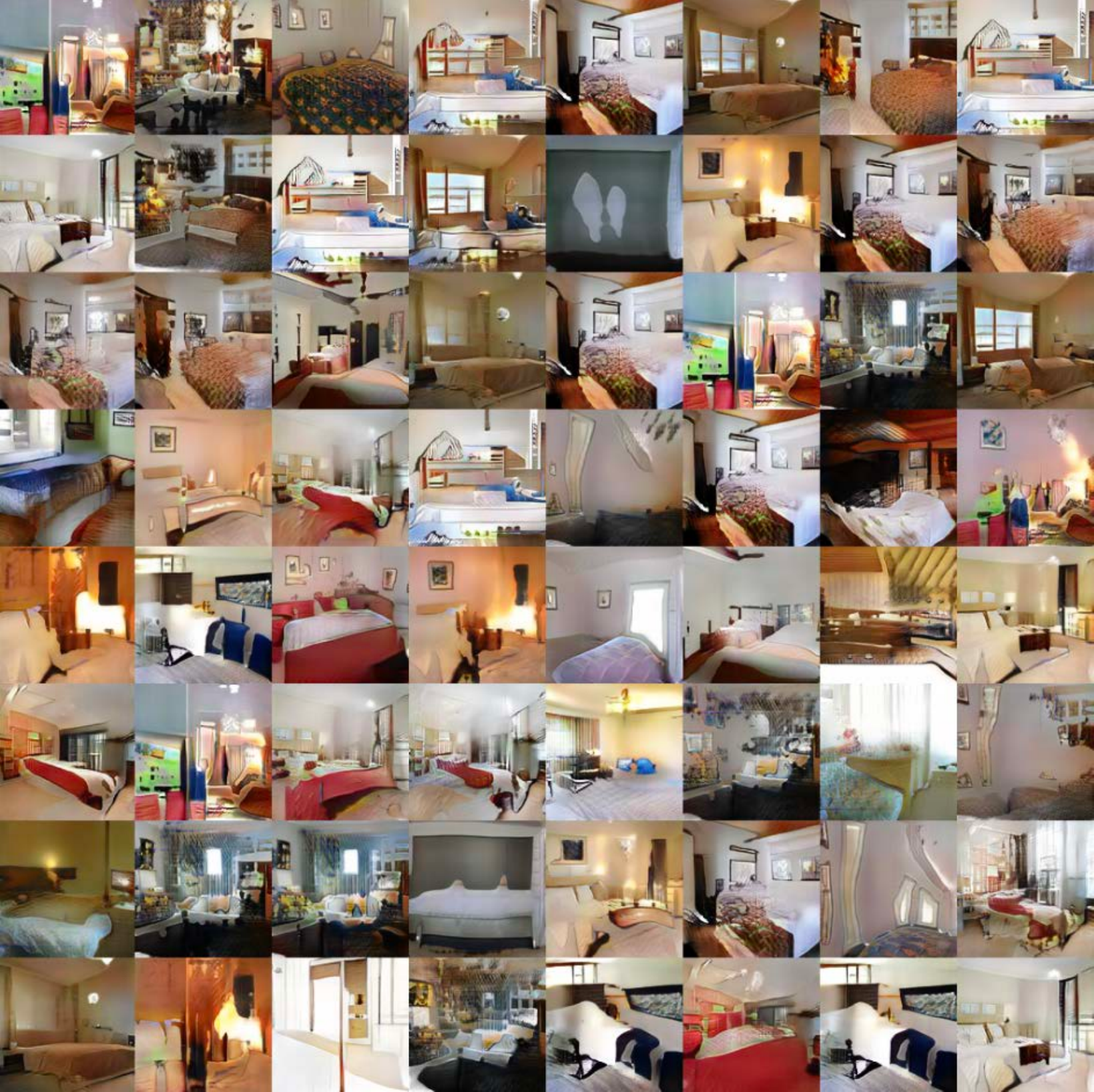}
\caption{Bedroom samples from SNGAN~\cite{miyato2018spectral}}
\label{fig:sngan_lsun_large}
\end{figure*}

\begin{figure*}[!t]
\centering
\includegraphics[width=1\linewidth]{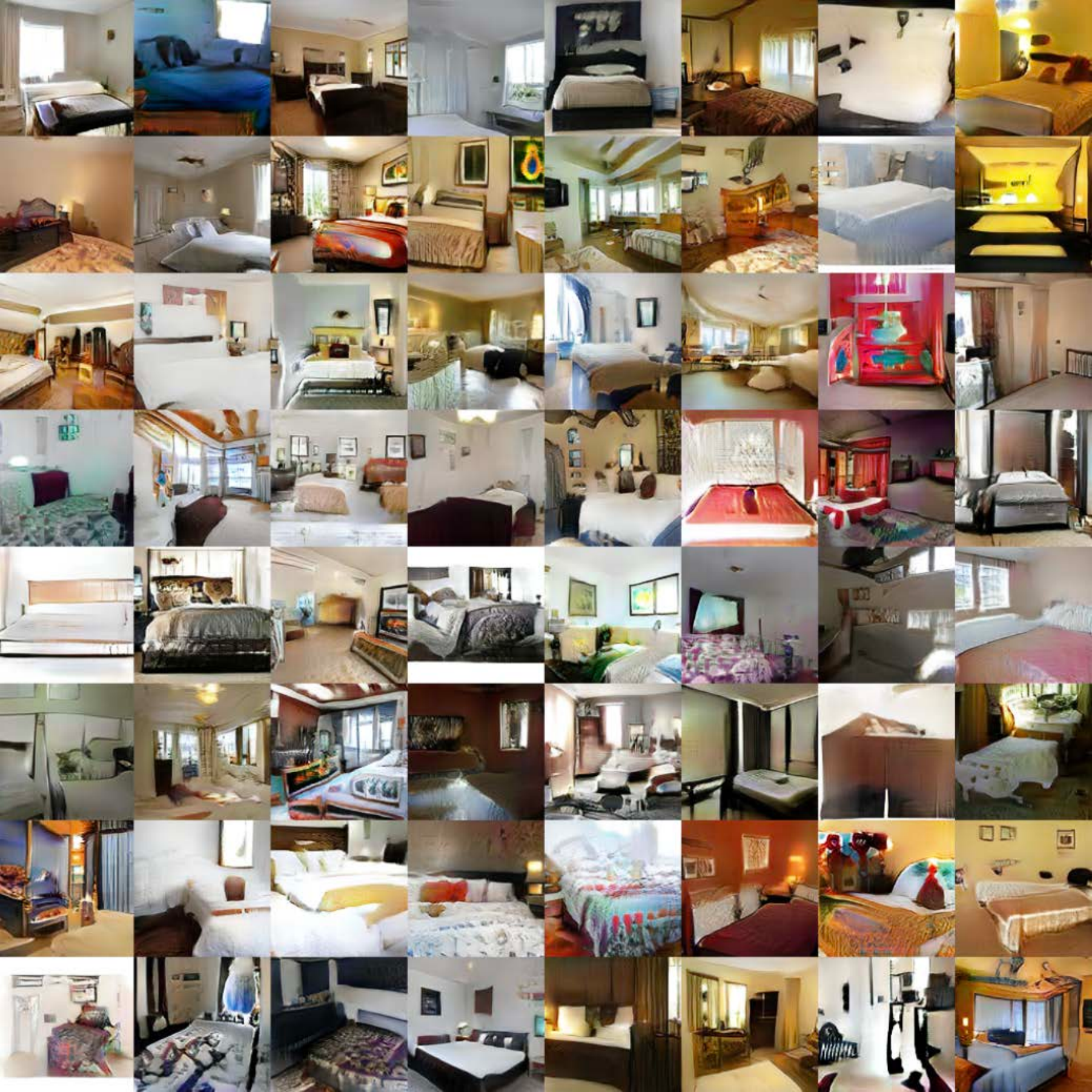}
\caption{Bedroom samples from CramerGAN~\cite{bellemare2017cramer}}
\label{fig:cramergan_lsun_large}
\end{figure*}

\begin{figure*}[!t]
\centering
\includegraphics[width=1\linewidth]{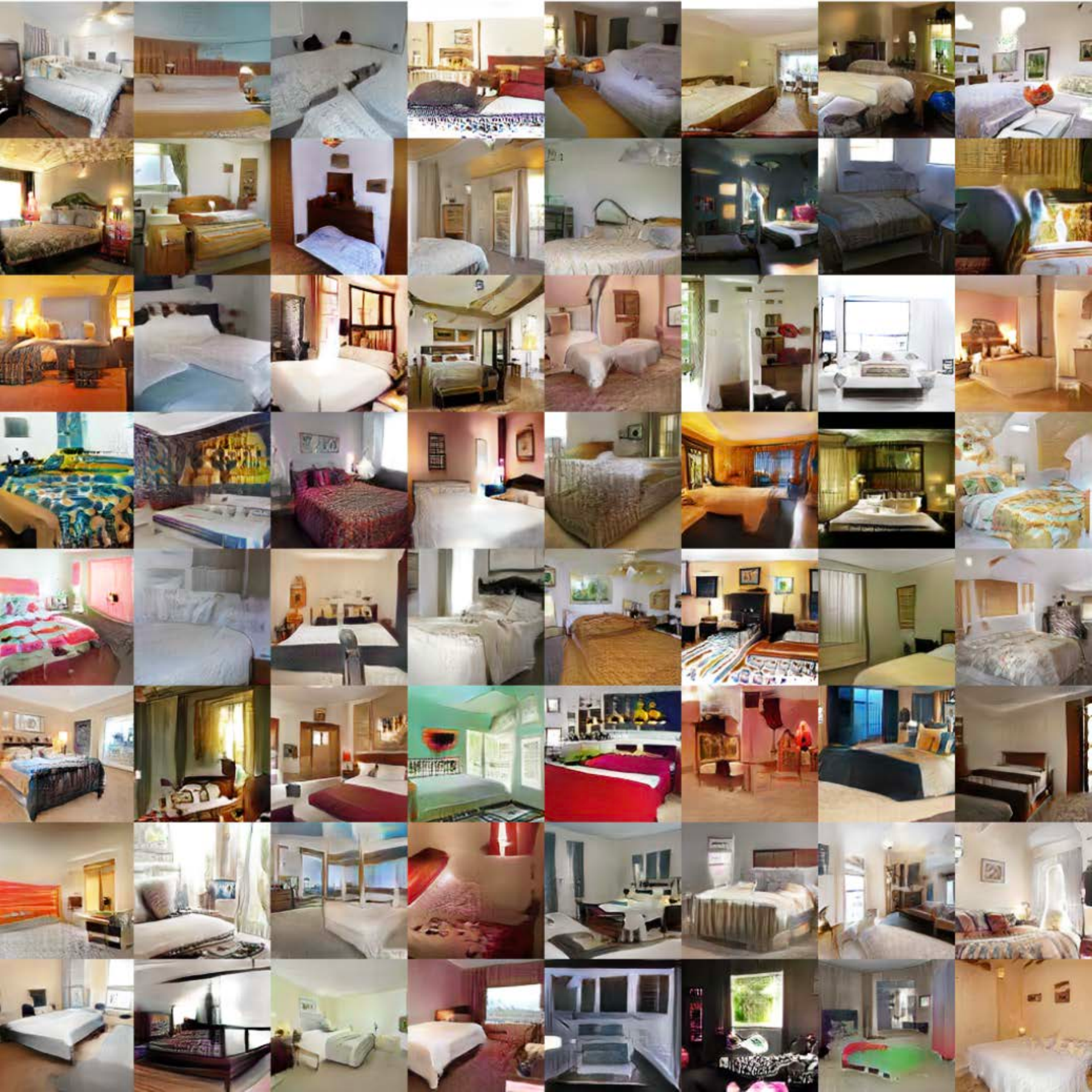}
\caption{Bedroom samples from MMDGAN~\cite{binkowski2018demystifying}}
\label{fig:mmdgan_lsun_large}
\end{figure*}

\begin{figure*}[!t]
\centering
\includegraphics[width=1\linewidth]{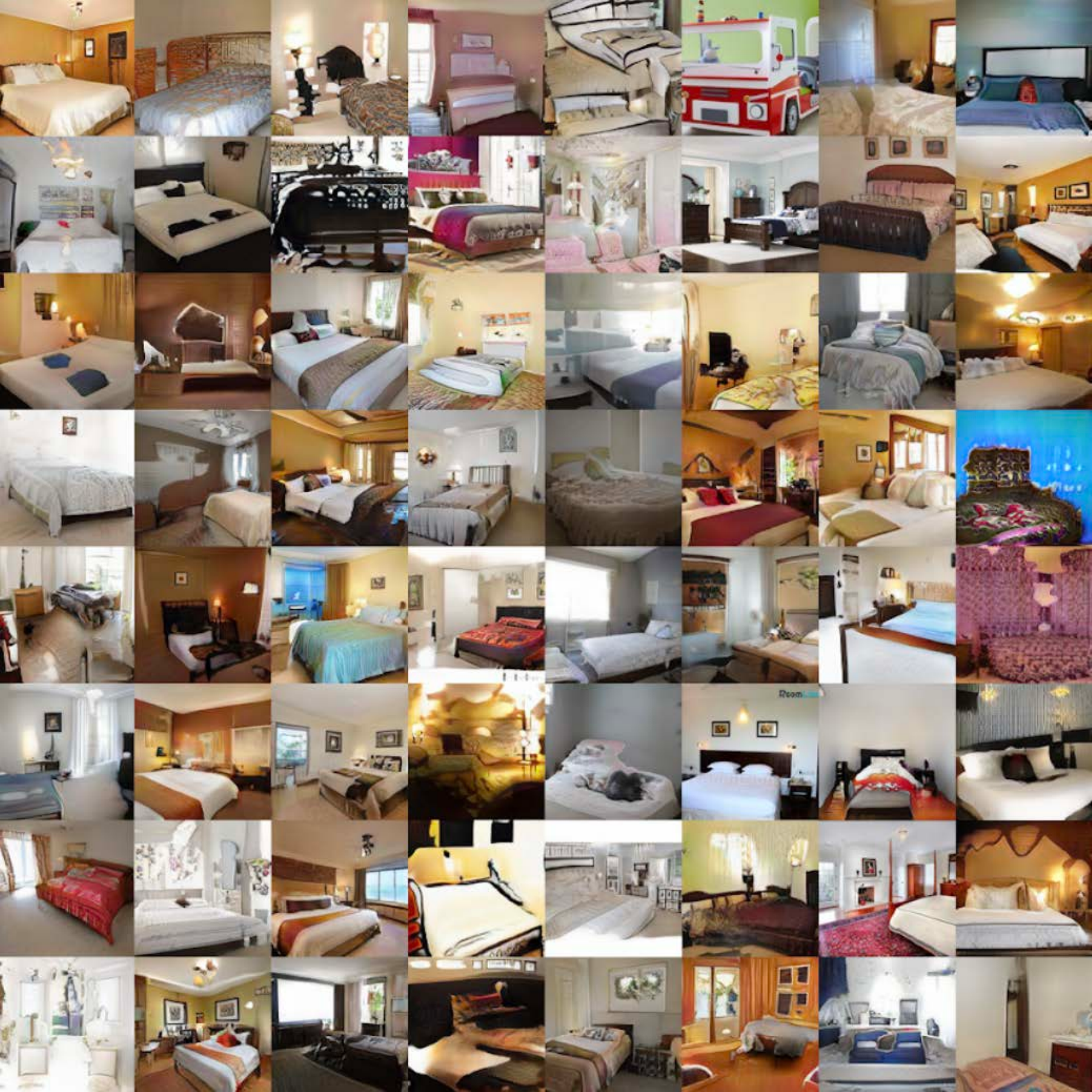}
\caption{Arbitrary bedroom samples from the setup of $\{$\textit{real}, \textit{ProGAN\_seed\_v\#i}$\}$ where $i\in \{1$, ..., $10\}$.}
\label{fig:random_samples_lsun_large}
\end{figure*}

\begin{figure*}[!t]
\centering
\includegraphics[width=1\linewidth]{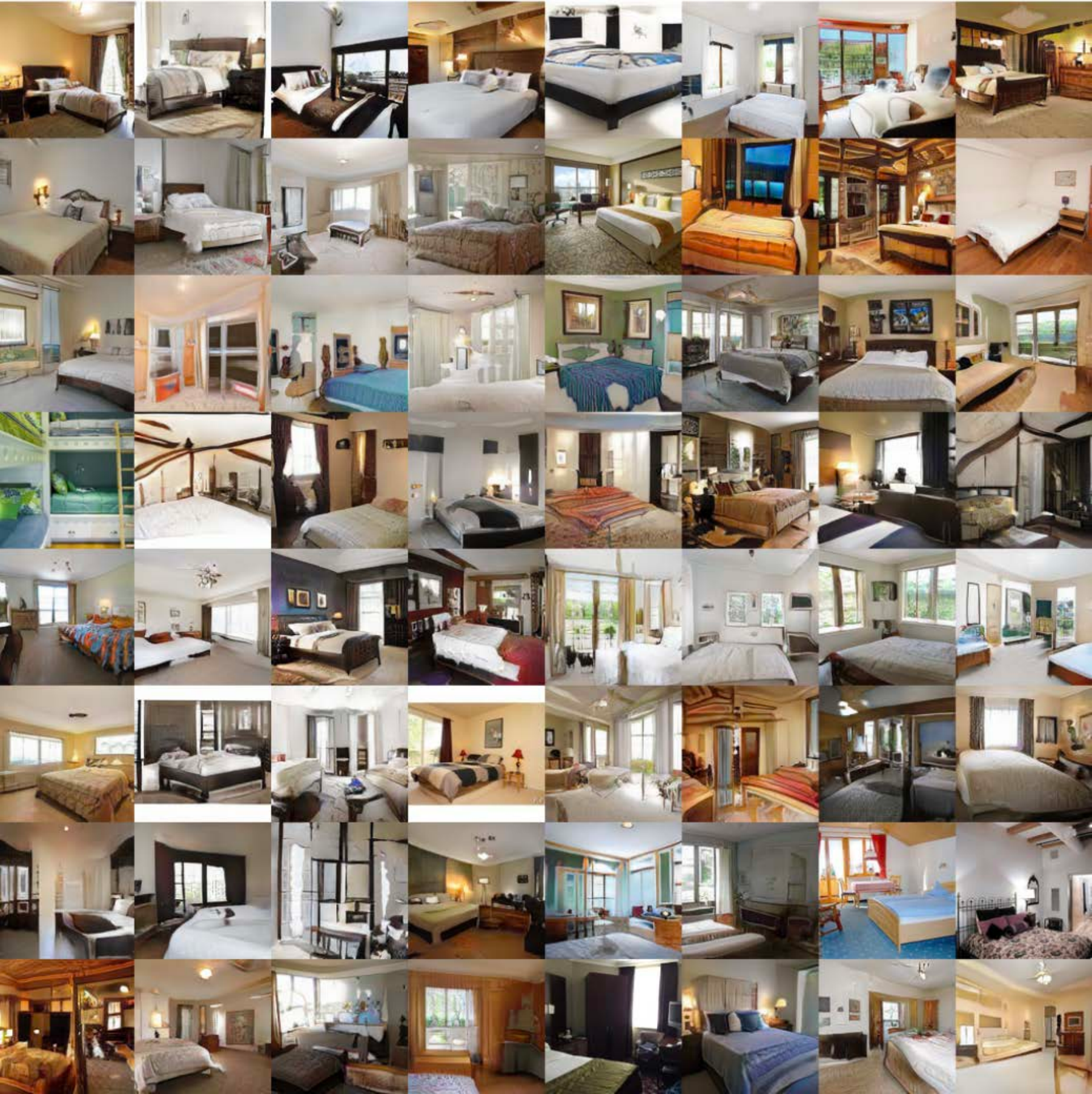}
\caption{Filtered bedroom samples from the setup of $\{$\textit{real}, \textit{ProGAN\_seed\_v\#i}$\}$ with the top 10\% largest Perceptual Similarity~\cite{zhang2018perceptual} to real dataset distribution.}
\label{fig:filtered_samples_lsun_large}
\end{figure*}

\begin{figure*}[!t]
\centering
\includegraphics[width=1\linewidth]{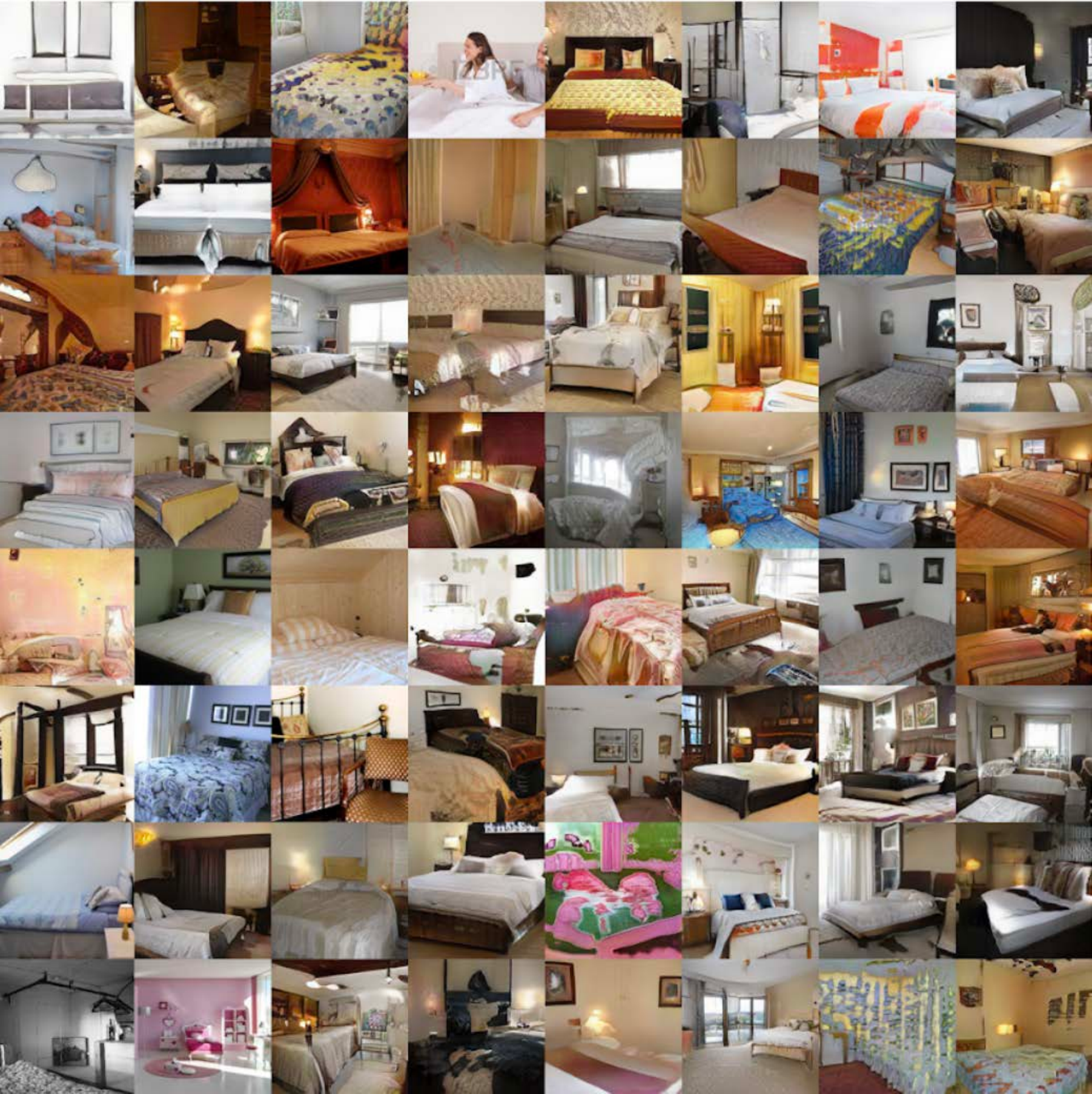}
\caption{Arbitrary bedroom samples without attack from the setup of $\{$\textit{real}, \textit{ProGAN\_seed\_v\#i}$\}$.}
\label{fig:plain_lsun_large}
\end{figure*}

\begin{figure*}[!t]
\centering
\includegraphics[width=1\linewidth]{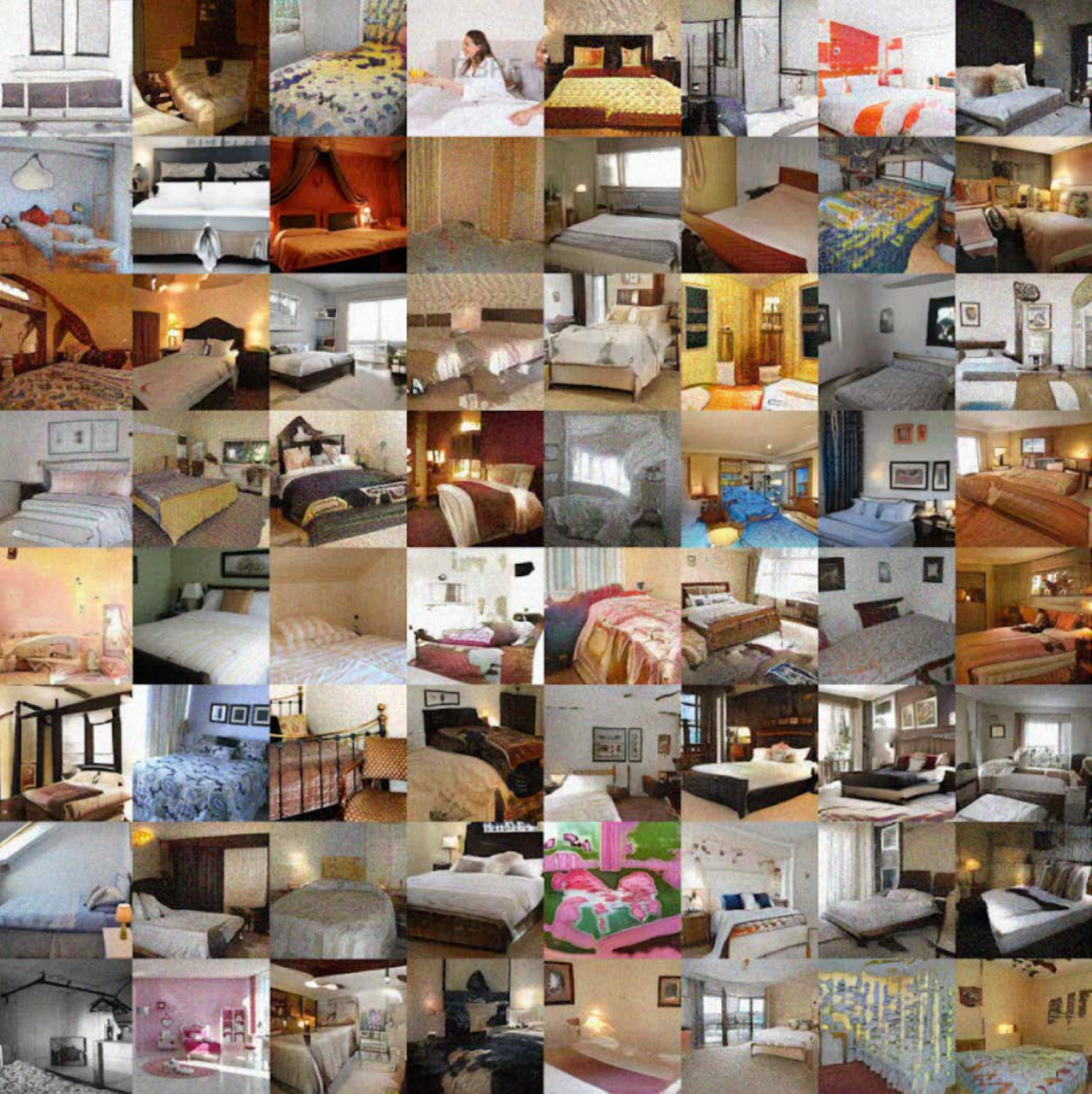}
\caption{Arbitrary bedroom samples with \textit{noise} attack from the setup of $\{$\textit{real}, \textit{ProGAN\_seed\_v\#i}$\}$.}
\label{fig:noise_lsun_large}
\end{figure*}

\begin{figure*}[!t]
\centering
\includegraphics[width=1\linewidth]{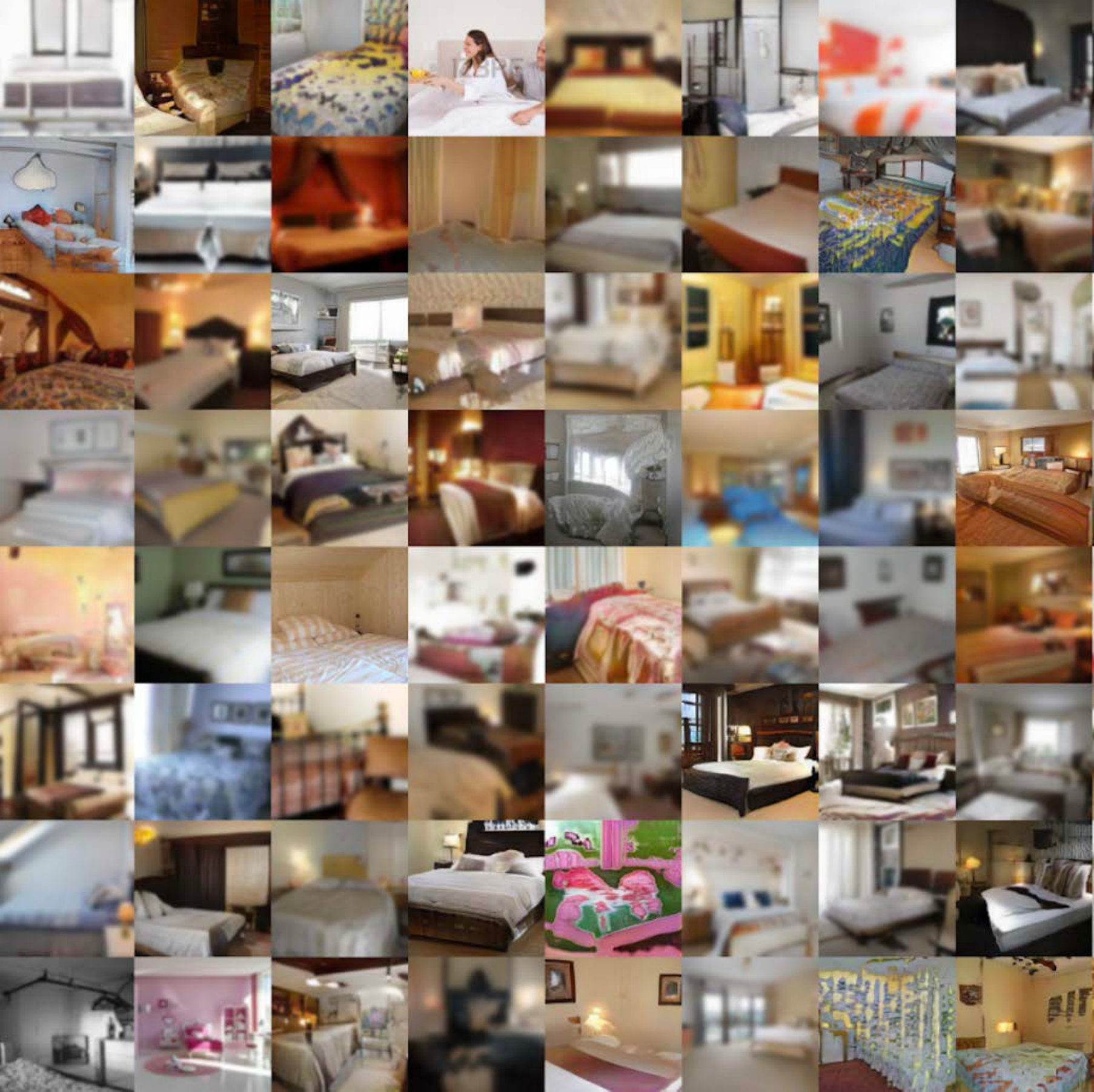}
\caption{Arbitrary bedroom samples with \textit{blur} attack from the setup of $\{$\textit{real}, \textit{ProGAN\_seed\_v\#i}$\}$.}
\label{fig:blur_lsun_large}
\end{figure*}

\begin{figure*}[!t]
\centering
\includegraphics[width=1\linewidth]{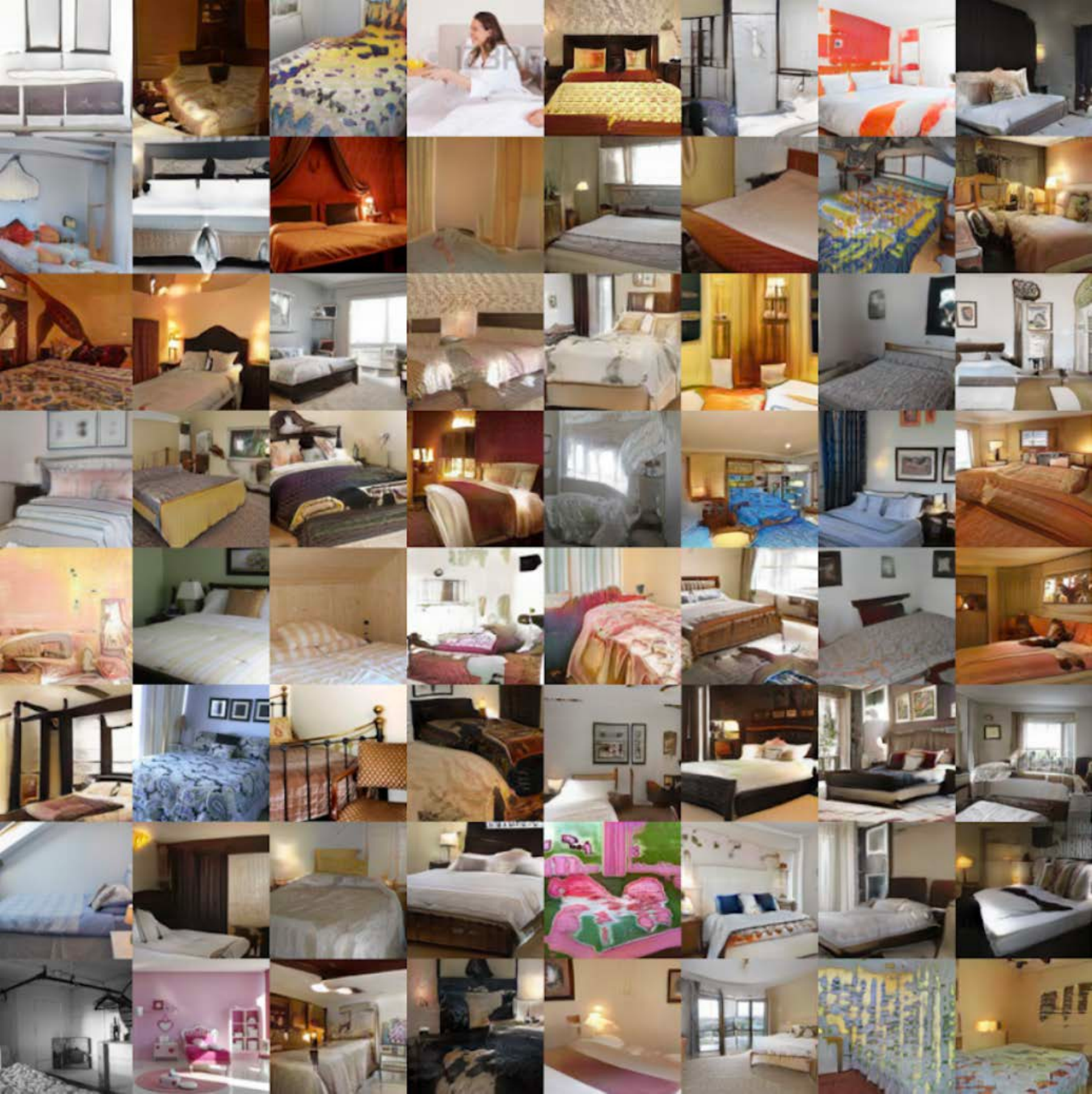}
\caption{Arbitrary bedroom samples with \textit{cropping} attack from the setup of $\{$\textit{real}, \textit{ProGAN\_seed\_v\#i}$\}$.}
\label{fig:crop_lsun_large}
\end{figure*}

\begin{figure*}[!t]
\centering
\includegraphics[width=1\linewidth]{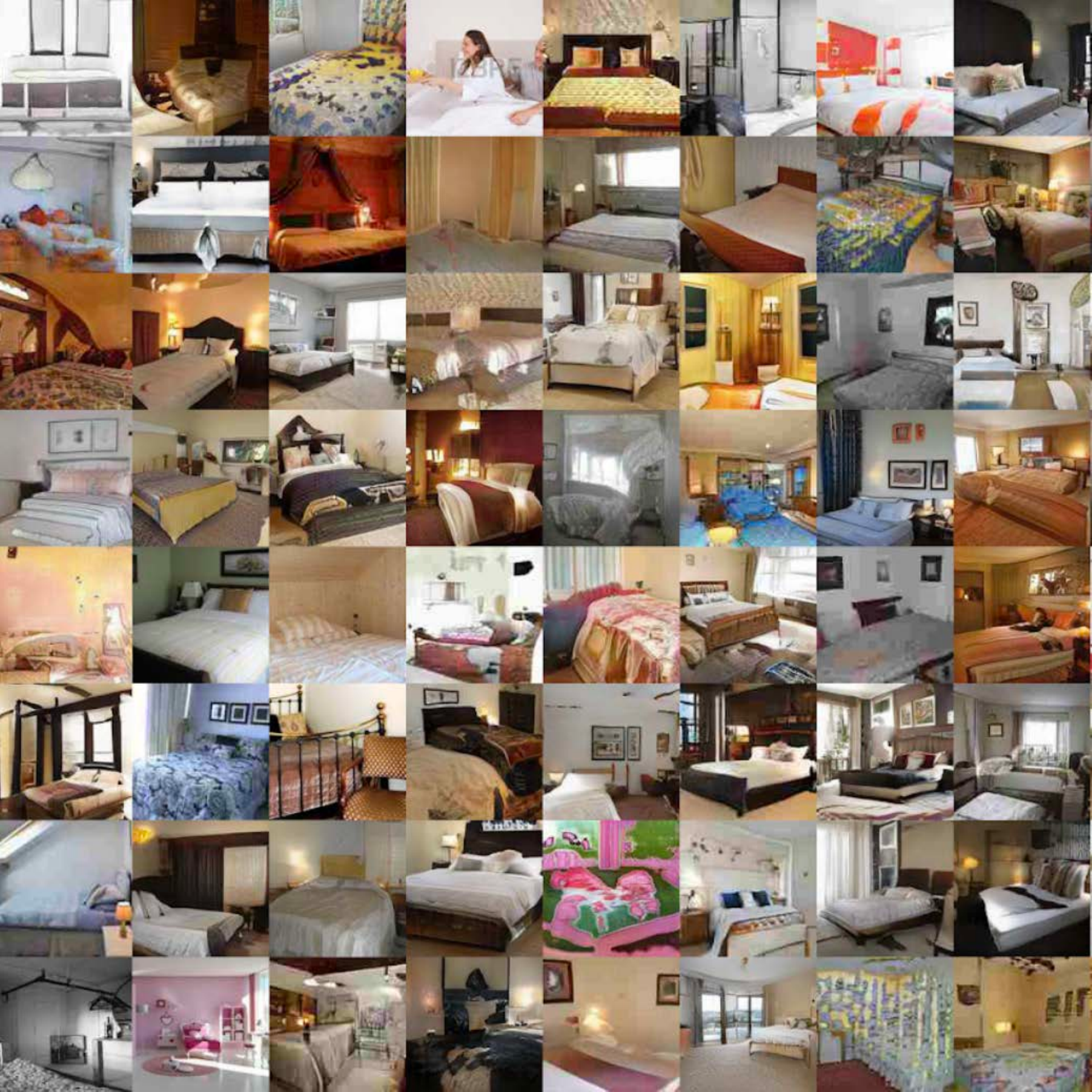}
\caption{Arbitrary bedroom samples with \textit{JPEG compression} attack from the setup of $\{$\textit{real}, \textit{ProGAN\_seed\_v\#i}$\}$.}
\label{fig:compression_lsun_large}
\end{figure*}

\begin{figure*}[!t]
\centering
\includegraphics[width=1\linewidth]{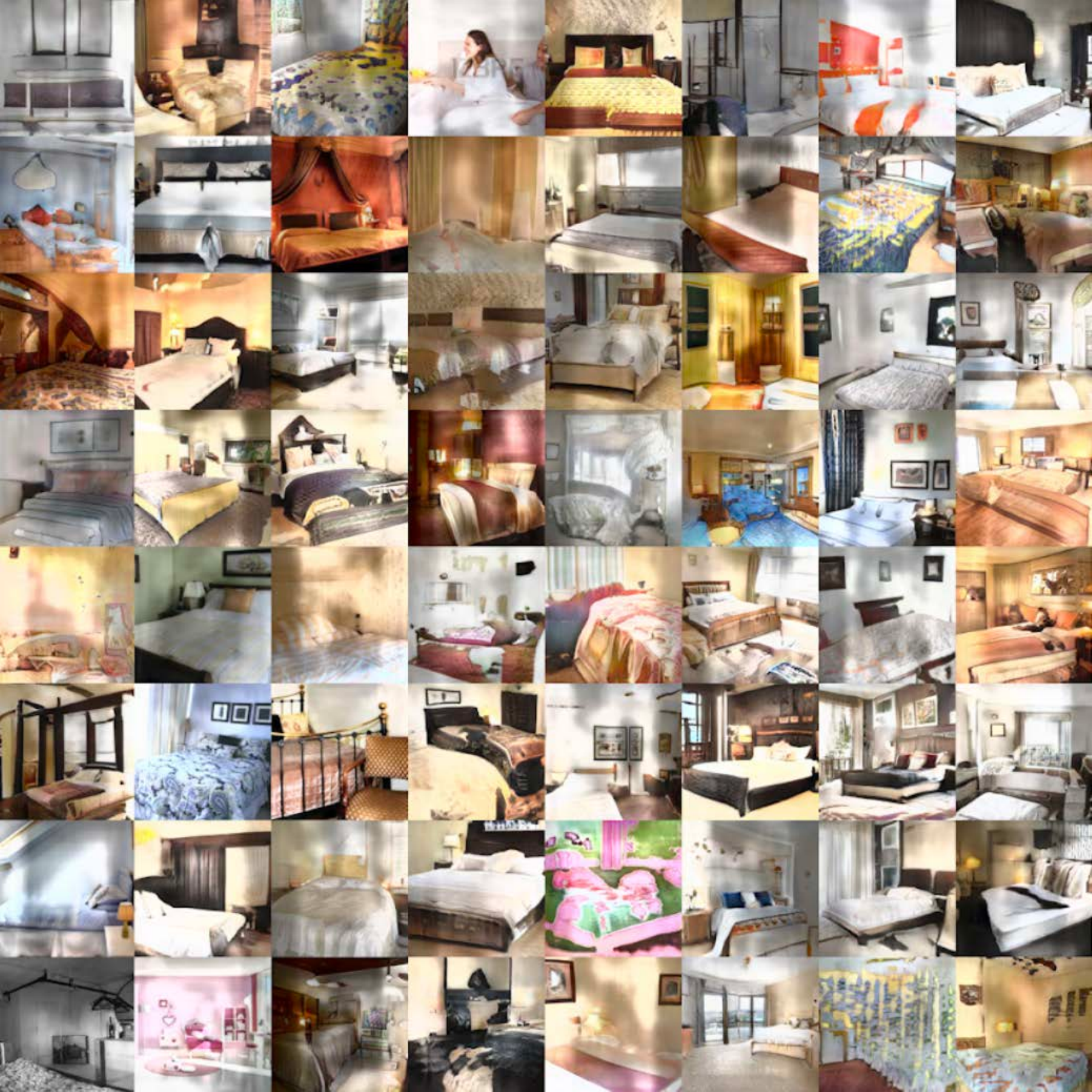}
\caption{Arbitrary bedroom samples with \textit{relighting} attack from the setup of $\{$\textit{real}, \textit{ProGAN\_seed\_v\#i}$\}$.}
\label{fig:relighting_lsun_large}
\end{figure*}

\begin{figure*}[!t]
\centering
\includegraphics[width=1\linewidth]{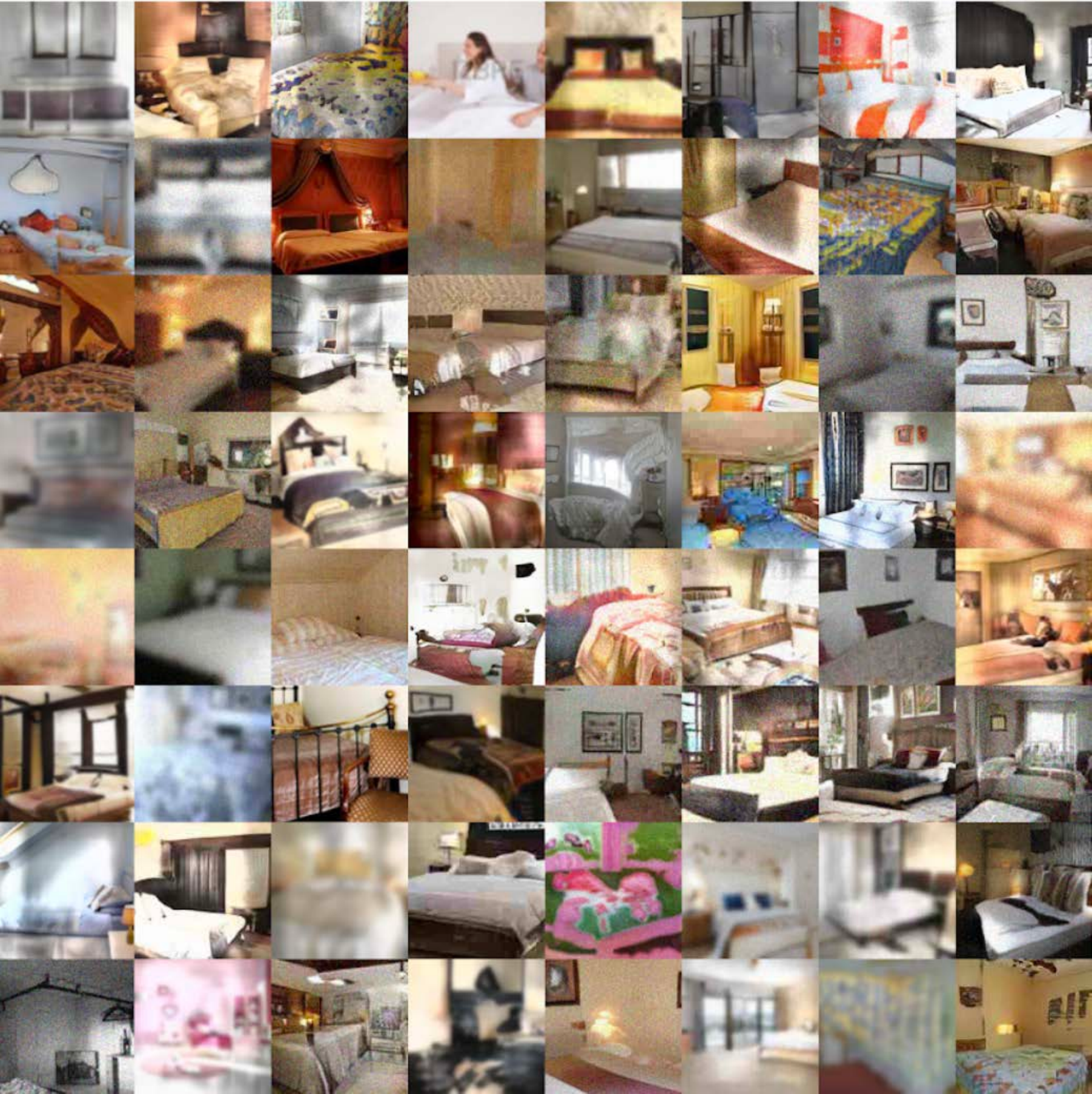}
\caption{Arbitrary bedroom samples with the combination attack from the setup of $\{$\textit{real}, \textit{ProGAN\_seed\_v\#i}$\}$.}
\label{fig:combo_lsun_large}
\end{figure*}

\begin{figure*}[!t]
\centering
\includegraphics[width=1\linewidth]{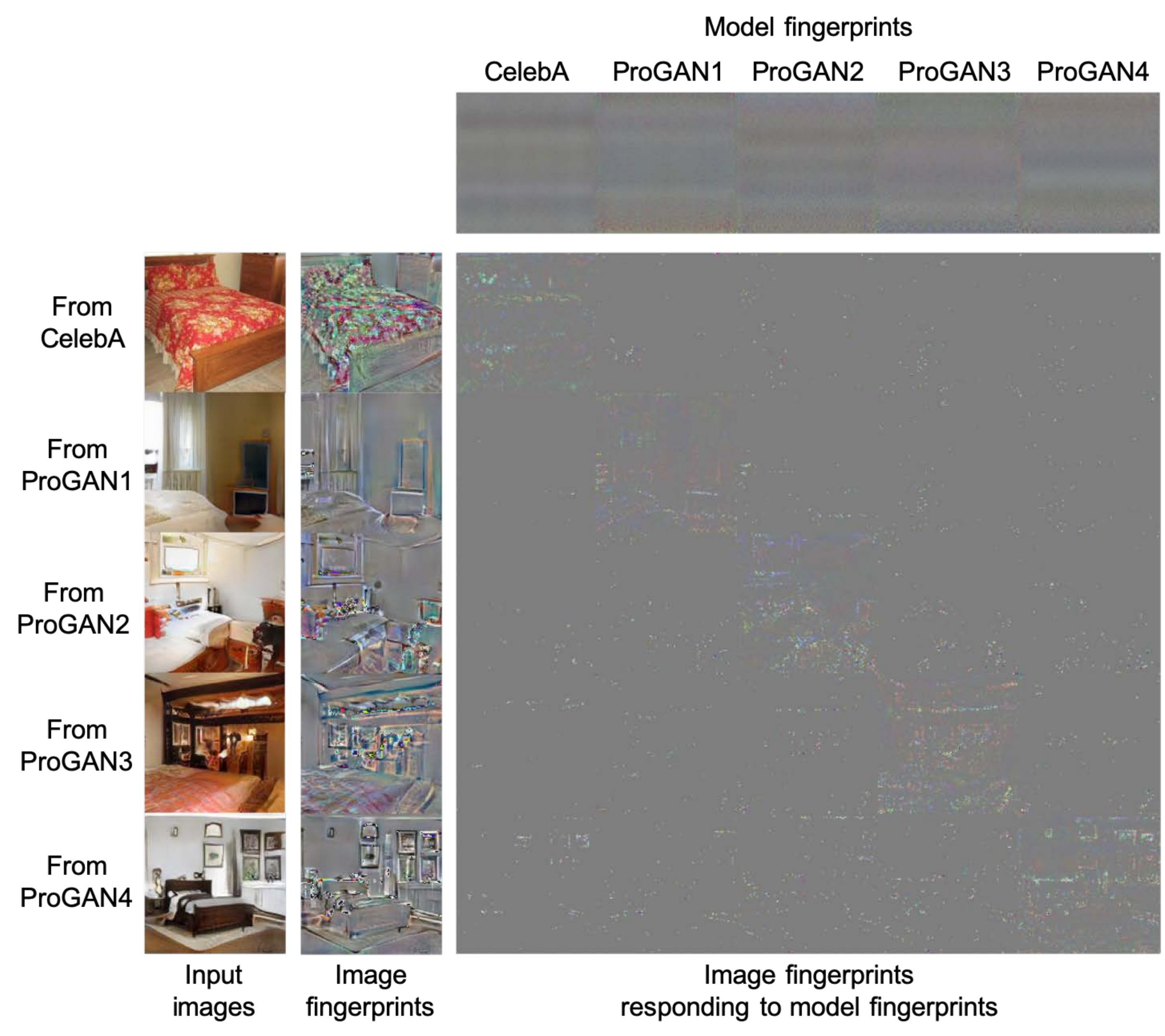}
\caption{Visualization of bedroom model and image fingerprint samples. Their pairwise interactions are shown as the confusion matrix. It turns out that image fingerprints maximize responses only to their own model fingerprints, which supports effective attribution.}
\label{fig:fingerprints_lsun}
\end{figure*}

We show bedroom samples corresponding to the experiments in the main paper. See Figure~\ref{fig:lsun_large}~to~\ref{fig:fingerprints_lsun}. In general, LSUN bedroom dataset is more challenging to a GAN model because of lack of image alignment. However, ProGAN~\cite{karras2018progressive} still performs equally well on this dataset and does not affect our conclusions in the main paper.

\end{document}